%% file: main.tex
\RequirePackage{fix-cm}
\documentclass{article}
\usepackage[final,sglblindworkshop]{neurips_2026}
\workshoptitle{Foundations of LLM Post-Training in Changing Environments}
\usepackage[utf8]{inputenc}
\usepackage[T1]{fontenc}
\usepackage{longtable}
\usepackage{hyperref,url,booktabs,amsfonts,nicefrac,microtype,xcolor,multirow,amsmath,graphicx,pgf}
\usepackage{tikz}
\AtBeginDocument{\parfillskip=0pt plus 0.7\linewidth}

\makeatletter
\renewcommand{\@noticestring}{%
  \@trackname\\
  Released checkpoint (\textsc{RL-Base}): \url{https://huggingface.co/gnani/gnani-evon-v3.3-30B-A3B}%
}
\makeatother

\title{Outside the Purview of the Reward:\\Capability Drift in Multilingual Post-Training}
\author{\textbf{Sutanshu Saksena Raj \quad Purusharth Malik \quad Pasunuti Prasanjith}\\
\textbf{Daksh Maan \quad Parijat Chatterjee \quad Avinash Benki}\\
Gnani.AI\\
\texttt{\{avinash, sutanshu.raj\} @ gnani.ai}}

\begin{document}
\maketitle

\begin{abstract}
Post-training can improve rewarded behavior while weakening capabilities users still need. We study two GRPO branches from a shared supervised checkpoint in a multilingual mixture-of-experts model. The released branch gains 36.07 points on MATH but loses 19.32 F1 points on XQuAD-Indic, while XQuAD-English improves under the same scoring rule. An independently trained sibling adds Indic instruction following and language-aware shaping. It recovers the regressing Indic task scores and preserves MATH, but does not restore script compliance and trades off competition mathematics and code. In a later round, judge training is followed by a sharp extraction decline from the Indic branch; reward-free distillation from its earlier checkpoint restores most of the lost F1. A MATH score decomposition associates most of the measured gain with increased boxed-answer emission; accuracy among boxed answers changes little. These results show why benchmark scores should be read alongside the response forms they reward.
\end{abstract}

\section{Introduction}
Post-training is often judged by benchmark gains. Yet a higher score can coexist with weaker performance on behaviors users still need, such as answering in their language or following a familiar output convention. Aggregate scores can hide this trade-off.

We study a 30B-total, 3B-active multilingual mixture-of-experts model~\citep{nvidia2025nemotron} covering English and eight Indic languages. Two GRPO branches start from the same supervised checkpoint. The released branch trains on mathematics, structured outputs, workplace tool use, and multiple-choice question answering; the sibling adds Indic instruction following and language-aware rewards.

We map the released branch's regressions and assess the role of reward coverage. We then follow the Indic-focused branch through judge training and reward-free distillation, using MATH and tool-use diagnostics to interpret score changes.

\section{Related work}
Multilingual instruction tuning can transfer task behavior across languages~\citep{muennighoff-etal-2023-crosslingual,ustun-etal-2024-aya}, although multilingual mixtures do not always outperform monolingual ones~\citep{chen-etal-2024-monolingual}. Indic datasets and benchmarks also highlight the importance of language-specific resources and evaluation~\citep{khan-etal-2024-indicllmsuite,singh-etal-2024-indicgenbench}. In our results, translated and grounded splits of the same benchmark family move in opposite directions.

Adaptation can weaken inherited capabilities, including under parameter-efficient fine-tuning~\citep{aggarwal-etal-2024-maple,toukmaji-flanigan-2025-lowresource,hu2021lora}. We study retention across the reinforcement-learning stage, where both the task distribution and the objective change. Unlike parameter-efficient adaptation, our runs update all parameters, including expert routing.

The closest work concerns language mixing under RL. \citet{li-etal-2025-impact} find that mixing languages during bilingual reasoning can help performance, with monolingual decoding costing 5.6 points on MATH500. We examine a different location: the answer the user reads. Our sibling branch rewards the requested language only after the final reasoning block, leaving internal reasoning unrestricted. This distinguishes useful flexibility in a reasoning trace from unwanted language drift in the final answer.

\section{Setup}
\label{sec:setup}
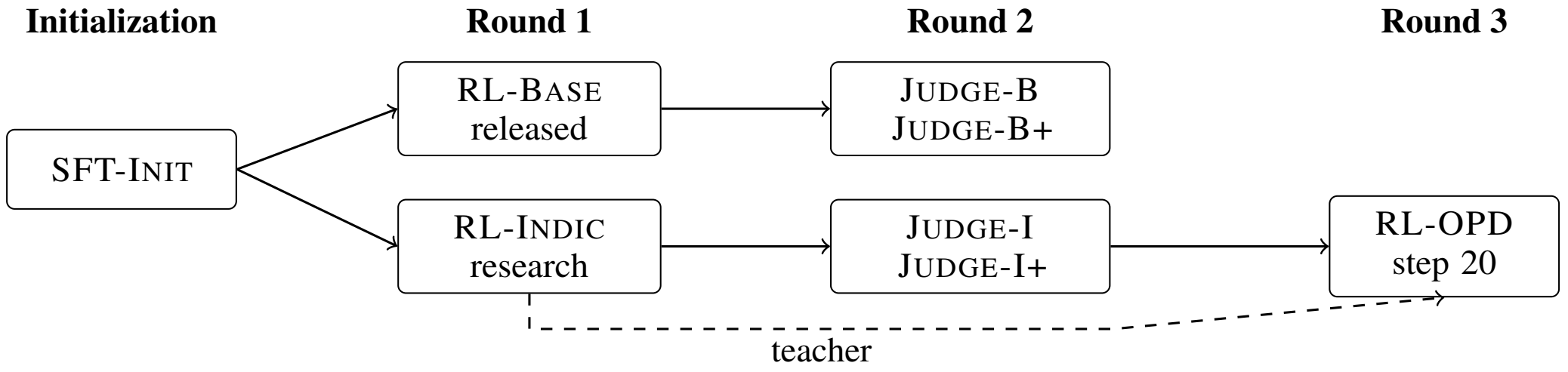
\begin{figure}[!htbp]
\centering
\resizebox{0.98\linewidth}{!}{\input{figures/checkpoint_lineage.tex}}
\caption{Checkpoint lineage across post-training rounds. \textsc{Judge-B} and \textsc{Judge-I} each have separate base and \texttt{+} configurations; \texttt{+} marks the auxiliary reward bundle. The dashed arrow marks \textsc{RL-Indic} as the distillation teacher.}
\label{fig:checkpoint-lineage}
\end{figure}

\paragraph{Checkpoints.} Figure~\ref{fig:checkpoint-lineage} summarizes the checkpoint lineage. Both first-round runs start from \textsc{SFT-Init}. \textsc{RL-Base} is the released branch trained on the original task mixture; \textsc{RL-Indic} is an independent research branch with added Indic instruction following and language-aware shaping. Judge training starts separately from each branch. \textsc{RL-OPD} starts from \textsc{Judge-I+} at step 20 and distills from \textsc{RL-Indic} without a task reward. All checkpoints share an architecture, vocabulary, and tokenizer.

\paragraph{Initialization.} The supervised corpus contains 64.0\% English and 36.0\% Indic rows, with Indic text in native and romanized scripts. The checkpoint was exported partway through the final curriculum pass. It already follows the requested script reliably, so script retention is a meaningful post-training test. Table~\ref{tab:sft-config} gives the training record, and Appendix~\ref{app:sft} describes the corpus and pre-RL conventions.

\paragraph{Environments and rewards.} \textsc{RL-Base} uses judged final-answer mathematics, JSON schema conformance, an iterative tool-call workplace assistant, and multiple-choice QA. Each environment returns a task-specific reward, with no language, script, length, grounding, or repetition term. \textsc{RL-Indic} adds an Indic instruction-following environment with a language term over the final answer and a repetition penalty over the whole generation. Tables~\ref{tab:rl-base-config} and~\ref{tab:rl-indic-config} give the reward coefficients and input mixture.

\paragraph{Optimization.} \textsc{RL-Base} samples 16 completions per prompt, and \textsc{RL-Indic} samples eight. Both use a leave-one-out baseline and normalize by the group's Bessel-corrected standard deviation when it is nonzero. The sibling branch also floors the divisor to avoid amplifying nearly constant rewards. Appendix~\ref{app:advantage} gives the exact estimators and the frequency with which the floor matters.

\subsection{Constructing the Indic reward}
\label{sec:indic-reward}
The added instruction-following environment scores both task completion and the form of the answer. Let $x$ be a training prompt, $y$ its completion, and $y_{\mathrm{ans}}$ the answer after the final reasoning block. For valid completions, the shaped reward is
\begin{equation}
R(x,y)=R_{\mathrm{task}}(x,y)
       +0.3R_{\mathrm{lang}}(x,y_{\mathrm{ans}})
       -0.3R_{\mathrm{rep}}(y).
\label{eq:indic-reward}
\end{equation}
Malformed or abandoned answers receive $-0.25$. The configured length coefficient is zero, so length contributes no penalty. Task reward is the fraction of the prompt's instruction constraints that are satisfied. Language reward is the fraction of answer words in the training row's target script. For $K$ constraints and a nonempty answer-word set $W$, these can be written as
\begin{equation}
R_{\mathrm{task}}=\frac{1}{K}\sum_{j=1}^{K}\mathbf{1}[c_j\text{ satisfied}],
\qquad
R_{\mathrm{lang}}=\frac{1}{|W|}\sum_{w\in W}\mathbf{1}[w\text{ in target script}].
\label{eq:reward-components}
\end{equation}
The language-identification gate distinguishes languages sharing a script, and unknown labels skip the language term. Repetition detection covers repeated $n$-grams and character runs across the whole completion. This scope leaves English internal reasoning unpenalized, while discouraging repetition in either the reasoning or the answer. It also keeps language compliance distinct from semantic correctness: an answer in the right script can still be wrong.

Rewards become training signals through the group's relative advantages. Writing $\mu_{-i}$ for the mean reward of the other completions, the nonzero-spread case is
\begin{equation}
A_i=\frac{r_i-\mu_{-i}}{D},\qquad
D=\begin{cases}
\sigma+10^{-6} & \text{for \textsc{RL-Base}},\\
\max(\sigma,0.05)+10^{-6} & \text{for \textsc{RL-Indic}}.
\end{cases}
\label{eq:advantages}
\end{equation}
Zero-spread groups are left undivided and discarded by dynamic sampling. The floor changes update magnitudes, not the ordering of completions within a group. In the Indic instruction-following environment, 44\% of all groups fall below its threshold. Among groups with nonzero spread, 39\% do. The floor therefore affects a substantial part of this environment's training signal, rather than only rare numerical edge cases.

\subsection{A learned judge and reward-free distillation}
\label{sec:later-rounds}

\paragraph{The judge.} The second round replaces the verifiable graders with a generative reward model trained for the purpose. Given a conversation and two candidate responses, it reasons before emitting an absolute score for each response and a relative preference, so its two outputs can be checked against each other for consistency. It is a mixture of experts with 26B total and 4B active parameters, trained with GRPO on preference pairs drawn from our Indic data and from HelpSteer3, each pair presented in both orders. Its own reward combines side agreement, a calibration term on the predicted scores, and a self-consistency penalty, with unparseable verdicts placed strictly below the clip floor. A batch-level rank-balance term keeps the predicted ranking from collapsing onto a few modes. Pairwise accuracy saturates early while calibration continues to move, so the exported checkpoint is chosen on calibration rather than on accuracy. Appendix~\ref{app:judge} gives the objective and the validation record.

\paragraph{Judge benchmarks.} Under greedy decoding with both presentation orders averaged, the judge scores above a 235B-parameter reference on JudgeBench and RM-Bench. Its largest margin is on the style-adversarial RM-Bench slice. Because it is pairwise, these scores are not comparable to those of scalar reward models (Appendix~\ref{app:judge}).

\paragraph{Round two.} We optimize against the frozen judge with the same GRPO objective and a KL penalty toward the initializing policy. Scoring every pair in a group is quadratic, so the judge is instead called on a circular ring in which each rollout is judged twice, once in each position, which cancels the judge's own positional bias. Both configurations use group-relative length control and a shortest-answer bonus. Both also penalize truncation. The \texttt{+} configuration additionally applies a think-length penalty to a pseudo-random fifth of groups exceeding a 5,000-token reasoning budget. It also adds repetition and script-consistency terms. Neither configuration rewards extraction from a supplied passage.

\paragraph{Round three.} \textsc{RL-OPD} starts from \textsc{Judge-I+} at step 20 and distills \textsc{RL-Indic} back into it, matching the student's next-token distribution to the teacher's over the teacher's top-$k$ set. The loss weights the forward and reverse truncated divergences equally, because the regressions it targets are collapse-shaped and the forward term is the one that penalizes a student for abandoning mass the teacher still places. No reward model, verifier, or frozen reference policy enters the objective. The teacher is the only anchor, and teacher and student share a vocabulary and a byte-identical tokenizer.

\subsection{Sampling and evaluation}
\paragraph{Dynamic sampling changes the mixture.} Both runs discard prompt groups whose completions all receive the same reward. An environment's share of retained groups is therefore
\begin{equation}
s_{\text{effective},e} \;\propto\; s_{\text{nominal},e}\; p_{\text{mixed},e},
\label{eq:shares}
\end{equation}
where $p_{\text{mixed},e}$ is the fraction of groups in environment $e$ with more than one reward value. Groups survive only at intermediate competence, because a task the model always fails is as uninformative as one it always solves, so an environment's realized share can rise and then fall while its input share stays fixed. The specified mixture thus need not match the realized training distribution. We cannot reconstruct the latter because the retained logs do not identify discarded groups.

\paragraph{Two measured comparisons.} Each checkpoint was scored in its own single-model campaign, and the comparisons we report pair those campaigns. Comparison~A sets \textsc{SFT-Init} against \textsc{RL-Base}, and comparison~B sets \textsc{RL-Base} against \textsc{RL-Indic}. The campaigns share a benchmark catalog, BF16 weights, reasoning mode, and decoding configuration, but they ran weeks apart, and harness versions and evaluated sample identifiers were not recovered. The \textsc{RL-Base} column is one scorecard read in both pairings, so its agreement across them is an identity rather than evidence of comparability. Only the reconstructed \textsc{RL-Indic} versus \textsc{SFT-Init} comparison rests on that assumption. Appendix~\ref{app:head} summarizes the retained evaluation record.

\section{RQ1: where the regressions concentrate}
\label{sec:rq1}
\input{matrix}

The regressions are concentrated in Indic benchmarks, and the three largest all supply a passage: XQuAD-Indic, IndicQA, and IndicIFEval-ground. IFEval-Hindi, which supplies none, moves down by about two points. HumanEval+ is nearly flat, while BBQ's ambiguous-context accuracy also falls. We treat BBQ separately as a bias diagnostic rather than a capability measure. Table~\ref{tab:full} reports the complete metric set.

Table~\ref{tab:matrix} helps distinguish possible explanations. Language alone is not enough, because other Indic tasks improve. Nor is task family, because both question answering and instruction following appear among the gains and losses. Simply supplying a source text does not explain the split either. FLORES-IN translation and CrossSum-IN summarization both improve despite requiring a source.

The most informative comparison is within XQuAD. English improves by 11.71 F1 points, while Indic falls by 19.32 under the same extractive requirement and scoring rule. Extraction therefore cannot explain the regressions on its own. The pattern suggests an interaction between language and task similarity: English gains extend beyond the rewarded tasks, while Indic extraction from a supplied passage deteriorates.

The instruction-following regressions are reported separately, not classified as extraction failures. IndicIFEval-ground's construction remains unverified, as does the presence of an in-context passage in XorQA-Indic. If XorQA supplies such a passage, its improvement would counter the proposed extractive pattern.

\section{RQ2: can reward coverage explain the pattern?}
\label{sec:rq2}
We propose that reward coverage is asymmetric across languages. The English-dominant environments may support broad transfer in English, even to extraction tasks they do not directly reward. Indic tasks may benefit mainly when they resemble the training tasks. Answering from a supplied Indic passage lies further from that mixture. This is an explanation to test, not a mechanism established by the benchmark deltas alone.

\paragraph{What the graders accept.} Structured-output and workplace rewards can be satisfied with JSON fields, English API names, or other machine-readable content. They do not require an Indic user-facing answer. This creates room for language drift without an explicit reward penalty.

\paragraph{What the rollout rewards show.} Within retained mixed-script prompt groups, Latin-script answers receive 0.168 more reward on average. Structured outputs and workplace assistance contribute most of this difference. Table~\ref{tab:reward-differential} gives the environment breakdown and bootstrap interval. The estimate uses sufficiently long answers and comes mainly from early training, when both scripts still occur within a prompt group.

Pairing fixes the prompt, but not the completion's correctness, length, formatting, or content. The differential is therefore an association, not the effect of changing script alone. Reward scales also differ by environment, and GRPO uses normalized advantages rather than raw rewards. Appendix~\ref{app:differential} reports the estimator, sampling limitations, and length-filter sensitivity.

\paragraph{A reward that can see the behavior.} The judge reads the whole answer, so an Indic extractive response is visible to it. Yet training against the judge sharply reduces extraction from \textsc{RL-Indic}. One possible explanation is that helpfulness comparisons favor elaborated answers over short verbatim spans, but we have not measured that preference directly. The result shows that visibility alone does not ensure that a reward protects a behavior.

\paragraph{The loss is not limited to a weak start.} The second round also starts from \textsc{RL-Base}, whose XQuAD-Indic score is much lower than \textsc{RL-Indic}'s. The sharp decline from \textsc{RL-Indic} shows that the regression does not depend on starting from a weak checkpoint.

\paragraph{Language and task shape come apart.} The same round separates the two explanations that Section~\ref{sec:rq1} could not. Grounded Indic instruction following gains about five points of strict accuracy under the judge reward while grounded Indic extraction loses two thirds of its F1, in one run and one language (Appendix~\ref{app:judge}). Both benchmarks supply a passage and both are Indic, yet only one produces the kind of prose a helpfulness judge scores well. The regressions are not a language effect in disguise.

\section{RQ3: what a bundled intervention recovers, and at what cost}
\label{sec:rq3}
\begin{table}[t]
\centering\footnotesize
\caption{Indic recovery and its trade-offs. $\Delta_{\mathrm{B}}$ is the directly measured \textsc{RL-Indic} minus \textsc{RL-Base} difference from comparison~B. \textsc{SFT-Init} scores come from comparison~A, so comparisons with that column assume campaign comparability. Scores are percentages, and deltas are percentage points. Table~\ref{tab:threeway} gives the complete comparison.}
\label{tab:rq3}
\setlength{\tabcolsep}{5pt}
\begin{tabular}{@{}lrrrr@{}}\toprule
Benchmark & \textsc{SFT-Init} & \textsc{RL-Base} & \textsc{RL-Indic} & $\Delta_{\mathrm{B}}$ \\\midrule
\multicolumn{5}{@{}l}{\textit{Indic regressions}} \\
XQuAD-Indic (EM) & 18.84 & \textbf{12.86} & 32.00 & $+19.14$ \\
IndicIFEval-ground & 40.52 & \textbf{33.05} & 60.00 & $+26.95$ \\
IndicQA (EM) & 12.77 & \textbf{12.22} & 23.46 & $+11.24$ \\
IFEval-Hindi & 52.71 & \textbf{50.59} & 72.29 & $+21.70$ \\
\addlinespace[3pt]
\multicolumn{5}{@{}l}{\textit{Other reasoning and instruction-following tasks}} \\
MATH & 48.69 & 84.76 & 86.76 & $+2.00$ \\
IFEval (prompt strict) & 66.36 & 82.81 & 86.88 & $+4.07$ \\
GPQA-Diamond & 38.19 & 65.03 & 60.04 & $-4.99$ \\
\addlinespace[3pt]
\multicolumn{5}{@{}l}{\textit{The cost}} \\
HMMT February 2025 & 59.58 & 60.83 & 50.42 & $-10.41$ \\
AIME 2025 & 72.92 & 74.38 & 65.42 & $-8.96$ \\
LiveCodeBench & 51.35 & 61.27 & 48.21 & $-13.06$ \\
MMLU-Pro & 66.67 & 77.26 & 63.34 & $-13.92$ \\\bottomrule
\end{tabular}
\end{table}

In the direct comparison, \textsc{RL-Indic} improves all four regressing Indic benchmarks by 11.24 to 26.95 points relative to \textsc{RL-Base}, while also improving MATH. Table~\ref{tab:rq3} shows these gains alongside the costs. Its scores also exceed the supervised baseline on the regression set, conditional on the campaign-comparability assumption in \S\ref{sec:setup}.

\paragraph{A different recipe, not an isolated reward effect.} The sibling branch changes the environment set, reward terms, English guardrail pool, group size, and advantage floor together. It shows that the same initialization can reach a different balance, but cannot identify which change produces the recovery. Separating added Indic task data from language and repetition shaping is the most useful next ablation.

\paragraph{Keeping English in the mixture.} An Indic-only dataset with an Indic-only reward produced an unwanted policy: English prompts also received Indic-script answers. The final mixture therefore retains a 9.9\% English guardrail pool. The goal is to follow the requested language rather than always emit Indic. Explicit language names also prove more effective cues than implicit or referential wording (Table~\ref{tab:language-cues}).

\paragraph{Recovery has a cost.} Against the same \textsc{RL-Base} baseline, MMLU-Pro falls by 13.92 points and LiveCodeBench by 13.06. Competition mathematics also declines (Table~\ref{tab:rq3}). The branches have nearly identical pooled means, but this hides opposing movements across tasks. Appendix~\ref{app:head} reports the aggregate and its sensitivity to metric choice. Preserving MATH does not imply preserving reasoning or code performance more broadly.

\paragraph{Task recovery is not script recovery.} On translated IndicIFEval, in-script output falls from 95.1\% to 84.6\% under \textsc{RL-Indic}, even as task accuracy improves. Native-Indic script retention also falls slightly (Figure~\ref{fig:script-diagnostics}B). The added language reward improves task performance without reliably controlling the answer's script.

\paragraph{The auxiliary bundle under the judge reward.} The \texttt{+} configuration produces a higher share of Indic-script answers than the configuration without it (Table~\ref{tab:span}). Since this comparison changes the whole auxiliary bundle, it does not isolate the effect of script shaping.

\section{What benchmark accuracy can hide}
\label{sec:identify}
The large MATH gain raises a measurement question: did the model solve more problems, or did it become better at presenting answers in the expected format? Boxed-answer emission rises from 55.2\% to 91.6\%, while accuracy among boxed answers changes little. Table~\ref{tab:math-conventions} shows that nearly all of the pooled score increase is associated with the emission-rate change.

\begin{table}[t]
\centering\small
\caption{MATH performance and answer conventions in comparison~A. Rates are percentages. The gain decomposition pools all 5,000 items, unlike the scorecard's average over seven subtasks. Emission-rate change is weighted by the initial conditional accuracy. Conditional-accuracy change is weighted by the released branch's emission rate. The pooled gain is 34.80 points, compared with 36.07~on~the~scorecard.}
\label{tab:math-conventions}
\begin{tabular}{@{}lrr@{}}\toprule
Measurement & \textsc{SFT-Init} & \textsc{RL-Base} \\\midrule
Boxed-answer emission & 55.2 & 91.6 \\
Accuracy given a boxed answer & 91.1 & 92.6 \\
Accuracy given an unboxed answer & 1.5 & 11.4 \\\midrule
Component of pooled gain & Points & Share \\\midrule
Boxed-answer emission rate & 33.13 & 95.2\% \\
Accuracy given a boxed answer & 1.39 & 4.0\% \\
Credit for unboxed answers & 0.28 & 0.8\% \\\bottomrule
\end{tabular}
\end{table}

This decomposition does not establish that the unboxed answers were correct. The boxed subset may itself change during training. Boxing is also measured after stripping the reasoning block, as in the scorer. Using the raw response changes the supervised checkpoint's conditional accuracy by nearly five points. Robust extraction or semantic regrading of all responses is needed to assess underlying MATH capability, and we have not run that test.

XQuAD-Indic points to a different failure. The share of answers written in English rises from 8.7\% to 91.7\%, while a prose-tolerant regrade barely changes F1 (Figure~\ref{fig:script-diagnostics}C). More permissive extraction does not explain away this regression. The language shift remains a central part of the observed failure.

\begin{table}[t]
\centering\small
\caption{Answer form and score on XQuAD-Indic, measured in one harness with reasoning enabled. Span is the median length in characters of the text after the final reasoning block, since the harness recorded no token counts. Script is the share of answers that are at least half Indic characters, and was not measured for \textsc{RL-Base}. The \textsc{RL-Base} row is an independent scoring of the checkpoint in Table~\ref{tab:full}, so it differs slightly from that column. A harness change made exact match more permissive after August 2026, which separates the first two rows from the rest; F1 is unaffected, and we read every comparison across that boundary on F1.}
\label{tab:span}
\begin{tabular}{@{}lrrrr@{}}\toprule
Checkpoint & Span & Script & EM & F1 \\\midrule
\textsc{RL-Base}        & 14  & --     & 12.65 & 17.17 \\
\textsc{RL-Indic}       & 18  & 80.0   & 32.00 & 50.44 \\\midrule
\textsc{Judge-B}, step 50   & 153 & 43.7 & 3.54  & 15.10 \\
\textsc{Judge-B+}, step 40  & 210 & 11.2 & 1.87  & 7.67  \\
\textsc{Judge-B+}, step 50  & 83  & 53.1 & 8.49  & 22.79 \\\midrule
\textsc{RL-OPD}, step 20    & 26  & 83.2 & 33.52 & 47.13 \\\bottomrule
\end{tabular}
\end{table}

The second round also changes an unscored output property: median answer length grows substantially under the judge reward, while distillation brings it close to its earlier value (Table~\ref{tab:span}). Because exact match is sensitive to answer length, this shift may contribute to the score changes, but it does not by itself explain them.

Script compliance moves the other way in the same runs. The auxiliary bundle raises the share of Indic-script answers on XQuAD-Indic and IndicQA. On IndicQA, F1 rises while exact match falls (Table~\ref{tab:span}); because the whole bundle changes, the score difference cannot be attributed to script shaping alone. These metrics can also respond differently to the same output change.

The measured effect also depends on the starting checkpoint. From \textsc{RL-Base}, one judge-trained checkpoint gains F1 while losing exact match, consistent with a change in answer form. From \textsc{RL-Indic}, F1 falls sharply. The contrast cautions against treating one score as a complete account of the change.

\section{Repeated adaptation and recovery without a task reward}
\label{sec:rounds}

\begin{table}[t]
\centering\small
\caption{XQuAD-Indic across three post-training rounds on the \textsc{RL-Indic} branch, each run for twenty steps. \textsc{RL-OPD} has been scored on this benchmark, not on the full matched suite. Rounds two and three use a more permissive exact match than round one, so we read the recovery on F1.}
\label{tab:rounds}
\begin{tabular}{@{}llrr@{}}\toprule
Round & Checkpoint & EM & F1 \\\midrule
1, verifiable rewards & \textsc{RL-Indic} & 32.00 & 50.44 \\
2, judge reward & \textsc{Judge-I+}, step 20 & 1.91 & 17.67 \\
3, no reward & \textsc{RL-OPD}, step 20 & 33.52 & 47.13 \\\bottomrule
\end{tabular}
\end{table}

Following \textsc{RL-Indic} through three rounds shows that later post-training can erase an earlier gain in extraction, and that distillation from the earlier checkpoint can restore most of it (Table~\ref{tab:rounds}). The distillation uses no task reward, but it does use the teacher's next-token distributions as its training signal.

The recovery is not a restoration. F1 lands a few points below the teacher's and the student's answers stay longer, so what has come back is most of the behavior rather than the teacher's own answers. Exact match reads slightly higher than the teacher's, but under a more permissive grader, so we do not count it as a gain. The student also inherits the teacher's limits by construction, and we have scored it on this benchmark alone.

\section{Interaction policy is also shaped by the environment}
\label{sec:policy}
\begin{figure}[!htbp]
\centering
\resizebox{\linewidth}{!}{\input{figures/tool_policy.tex}}
\caption{Tool-use trade-offs under different comparison baselines. (A) Direct \textsc{RL-Indic} minus \textsc{RL-Base} changes in percentage points, with branch scores beneath each benchmark name. (B) BFCL category accuracies relative to \textsc{SFT-Init}. Parallel calls deteriorate while sequential interaction improves. Given a second turn, the sibling supplies the missing call in 95 of 96 examined cases.}
\label{fig:tool-policy}
\end{figure}
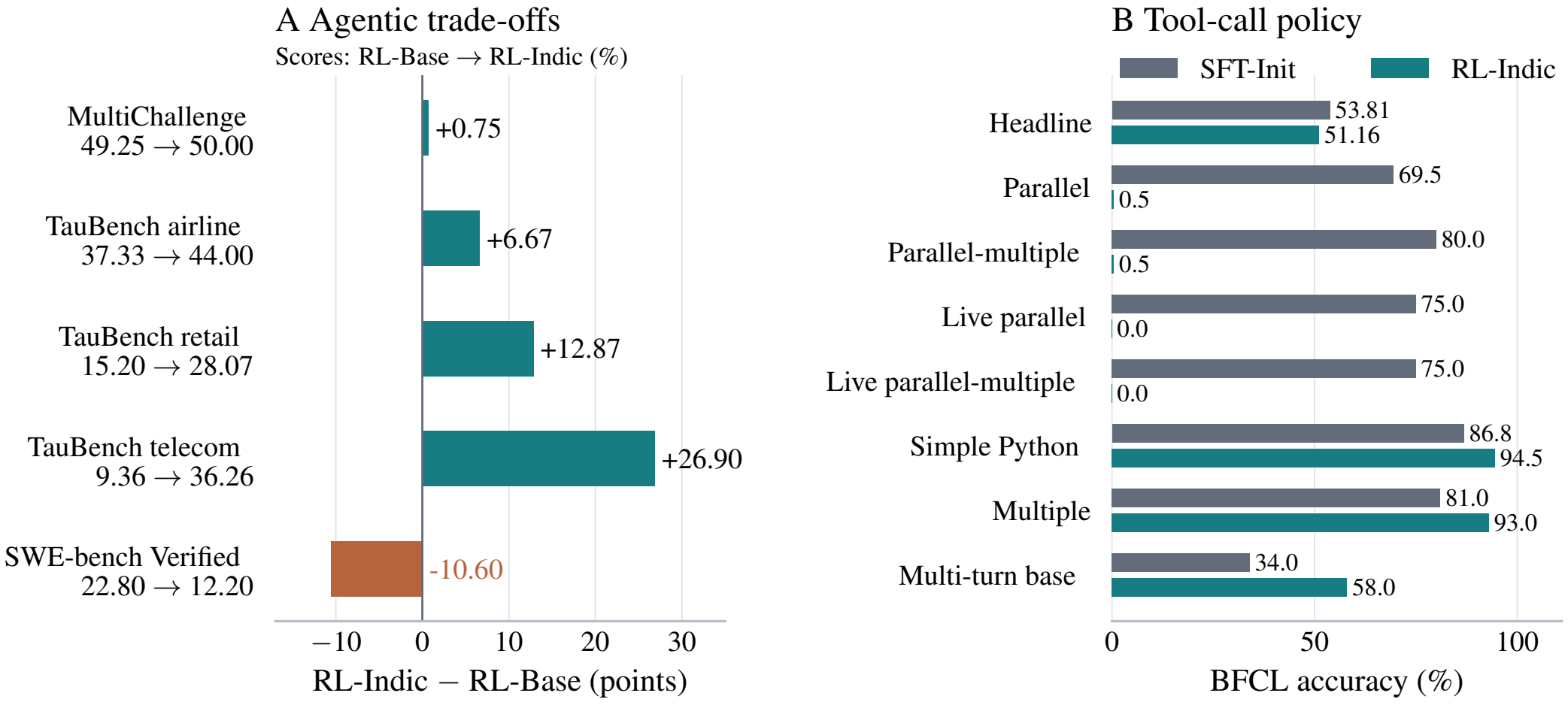

Training also shapes how the model uses tools. On BFCL v4, \textsc{RL-Indic} usually emits one call when the benchmark expects parallel calls in a single message. Parallel categories fall close to zero, while multi-turn base accuracy rises from 34.0 to 58.0. Given another turn, the model usually supplies the missing call (Figure~\ref{fig:tool-policy}B). This follows the training environment's pattern of alternating calls~and~observations.

The policy has mixed consequences. Relative to \textsc{RL-Base}, the sibling gains 26.90 points on TauBench telecom but loses 10.60 on SWE-bench Verified. It also improves the other measured interactive multi-turn tasks (Figure~\ref{fig:tool-policy}A). The contrast suggests that training-time turn structure benefits interactive use, but can hinder autonomous workflows that batch independent actions.

\section{Threats to validity}
\label{sec:threats}
\paragraph{Evaluation comparability.} Identical shared-checkpoint scores support, but do not prove, comparability between campaigns, and unrecorded differences could affect the reconstructed comparison with \textsc{SFT-Init} even if \textsc{RL-Base} stayed unchanged. The directly measured comparisons do not rest on that assumption (Appendix~\ref{app:head}). Script-adjacent measurements vary across recorded campaigns (Figure~\ref{fig:script-diagnostics}A); we report the release protocol's reading, and the others are shown as a sensitivity check and enter no delta in this paper.

\paragraph{Mechanism and measurement.} The bundled intervention does not isolate a reward term, and the script differential is sensitive to length filtering, reward scale, and rollout selection. IndicIFEval-ground and XorQA-Indic need construction checks before they can test the proposed taxonomy. Robust MATH regrading would separate answer conventions from task capability, as semantically matched rescoring would for the script differential.

\paragraph{Statistical and adaptation scope.} The main branch comparisons use one seed, and benchmark deltas have no per-item uncertainty intervals. Only the reward differential carries a bootstrap interval. The three-round trace in Section~\ref{sec:rounds} follows one branch and one benchmark, so it shows that these effects can compound and reverse without establishing how far either behavior generalizes. The truncated logs also limit within-run analysis.

\section{Conclusion}
The Indic-focused recipe recovers several regressing task scores but does not restore script compliance, and later judge training sharply reduces extraction from that branch. Distillation from the earlier checkpoint restores most of the lost F1 without a task reward. These results support tracking extraction alongside benchmark scores across post-training rounds.

\paragraph{Model availability.}
\label{sec:availability}
\textsc{RL-Base} is released as \texttt{gnani-evon-v3.3-30B-A3B}, and its model card documents intended uses, limitations, and licensing. The remaining checkpoints in this study are research artifacts that we do not release.

\typeout{MAIN-END-PAGE=\thepage}
\clearpage
\input{references}
\clearpage
\appendix
\input{main_extended}
\end{document}

%% file: figures/checkpoint_lineage.tex
\begin{tikzpicture}[
  checkpoint/.style={draw, rounded corners=2pt, align=center,
    minimum height=0.72cm, inner sep=4pt, font=\small},
  flow/.style={->, semithick},
  teacher/.style={->, semithick, dashed}
]
\node[font=\small\bfseries] at (1.5,0.45) {Initialization};
\node[font=\small\bfseries] at (5.2,0.45) {Round 1};
\node[font=\small\bfseries] at (9.2,0.45) {Round 2};
\node[font=\small\bfseries] at (13.5,0.45) {Round 3};

\node[checkpoint, text width=1.8cm] (sft) at (1.5,-0.9) {\textsc{SFT-Init}};
\node[checkpoint, text width=2.1cm] (base) at (5.2,-0.35) {\textsc{RL-Base}\\released};
\node[checkpoint, text width=2.1cm] (indic) at (5.2,-1.6) {\textsc{RL-Indic}\\research};
\node[checkpoint, text width=2.25cm] (judgeb) at (9.2,-0.35) {\textsc{Judge-B}\\\textsc{Judge-B+}};
\node[checkpoint, text width=2.25cm] (judgei) at (9.2,-1.6) {\textsc{Judge-I}\\\textsc{Judge-I+}};
\node[checkpoint, text width=1.8cm] (opd) at (13.5,-1.6) {\textsc{RL-OPD}\\step 20};

\draw[flow] (sft.east) -- (base.west);
\draw[flow] (sft.east) -- (indic.west);
\draw[flow] (base.east) -- (judgeb.west);
\draw[flow] (indic.east) -- (judgei.west);
\draw[flow] (judgei.east) -- (opd.west);
\draw[teacher] (indic.south) -- (5.2,-2.35) -- (10.5,-2.35)
  node[midway, below=1pt, font=\small, fill=white, inner sep=1pt] {teacher}
  -- (opd.south);
\end{tikzpicture}

%% file: matrix.tex
\begin{table}[t]
\centering\footnotesize
\caption{Benchmark-property matrix for comparison~A. Extractive rows report F1. The matrix
distinguishes source text supplied in a prompt from answers graded against a span of that text.
XQuAD-English and XQuAD-Indic share these properties but move in opposite directions.
Question marks identify unverified benchmark properties.}
\label{tab:matrix}
\setlength{\tabcolsep}{4pt}
\begin{tabular}{@{}llcccl r@{}}\toprule
Benchmark & Input & Passage & Recovered & Output script & Grader & $\Delta_{\text{Base}}$ \\
 & language & supplied & from it & constrained & & \\\midrule
MATH & En & no & no & no & judge, final answer & $+36.07$ \\
GPQA-Diamond & En & no & no & no & exact match, option & $+26.84$ \\
IFBench & En & no & no & no & constraint checker & $+22.66$ \\
IFEval & En & no & no & no & constraint checker & $+16.45$ \\
MMLU-Pro & En & no & no & no & exact match, option & $+10.59$ \\
\textbf{XQuAD-English} & \textbf{En} & \textbf{yes} & \textbf{yes} & no & EM / F1 vs.\ span & $\mathbf{+11.71}$ \\
\addlinespace[3pt]
IndicMMLU-Pro & Indic & no & no & no (option letter) & exact match & $+21.81$ \\
MILU & Indic & no & no & no (option letter) & exact match & $+5.17$ \\
BhashaBench-Multi & Indic & no & no & no (option letter) & exact match & $+7.16$ \\
XorQA-Indic & Indic & no$^{?}$ & no$^{?}$ & yes & EM / F1 vs.\ answer & $+27.93$ \\
IndicIFEval-translated & Indic & no & no & yes & constraint checker & $+12.36$ \\
FLORES-IN & En$\rightarrow$Indic & yes & no & yes & chrF++ vs.\ reference & $+21.78$ \\
CrossSum-IN & cross-ling. & yes & no & yes & chrF vs.\ reference & $+5.81$ \\
\addlinespace[3pt]
\textbf{XQuAD-Indic} & Indic & \textbf{yes} & \textbf{yes} & yes & EM / F1 vs.\ span & $\mathbf{-19.32}$ \\
\textbf{IndicQA} & Indic & \textbf{yes} & \textbf{yes} & yes & EM / F1 vs.\ span & $\mathbf{-8.64}$ \\
\textbf{IndicIFEval-ground} & Indic & ? & ? & yes & constraints$^{?}$ & $\mathbf{-7.47}$ \\
\addlinespace[3pt]
IFEval-Hindi & Hindi & no & no & yes & constraint checker & $-2.12$ \\\bottomrule
\end{tabular}
\end{table}

%% file: figures/tool_policy.tex
\begingroup%
\makeatletter%
\begin{pgfpicture}%
\pgfpathrectangle{\pgfpointorigin}{\pgfqpoint{7.300000in}{3.800000in}}%
\pgfusepath{use as bounding box, clip}%
\begin{pgfscope}%
\pgfsetbuttcap%
\pgfsetmiterjoin%
\definecolor{currentfill}{rgb}{1.000000,1.000000,1.000000}%
\pgfsetfillcolor{currentfill}%
\pgfsetlinewidth{0.000000pt}%
\definecolor{currentstroke}{rgb}{1.000000,1.000000,1.000000}%
\pgfsetstrokecolor{currentstroke}%
\pgfsetdash{}{0pt}%
\pgfpathmoveto{\pgfqpoint{0.000000in}{0.000000in}}%
\pgfpathlineto{\pgfqpoint{7.300000in}{0.000000in}}%
\pgfpathlineto{\pgfqpoint{7.300000in}{3.800000in}}%
\pgfpathlineto{\pgfqpoint{0.000000in}{3.800000in}}%
\pgfpathlineto{\pgfqpoint{0.000000in}{0.000000in}}%
\pgfpathclose%
\pgfusepath{fill}%
\end{pgfscope}%
\begin{pgfscope}%
\pgfsetbuttcap%
\pgfsetmiterjoin%
\definecolor{currentfill}{rgb}{1.000000,1.000000,1.000000}%
\pgfsetfillcolor{currentfill}%
\pgfsetlinewidth{0.000000pt}%
\definecolor{currentstroke}{rgb}{0.000000,0.000000,0.000000}%
\pgfsetstrokecolor{currentstroke}%
\pgfsetstrokeopacity{0.000000}%
\pgfsetdash{}{0pt}%
\pgfpathmoveto{\pgfqpoint{1.241000in}{0.722000in}}%
\pgfpathlineto{\pgfqpoint{3.308483in}{0.722000in}}%
\pgfpathlineto{\pgfqpoint{3.308483in}{3.230000in}}%
\pgfpathlineto{\pgfqpoint{1.241000in}{3.230000in}}%
\pgfpathlineto{\pgfqpoint{1.241000in}{0.722000in}}%
\pgfpathclose%
\pgfusepath{fill}%
\end{pgfscope}%
\begin{pgfscope}%
\pgfpathrectangle{\pgfqpoint{1.241000in}{0.722000in}}{\pgfqpoint{2.067483in}{2.508000in}}%
\pgfusepath{clip}%
\pgfsetrectcap%
\pgfsetroundjoin%
\pgfsetlinewidth{0.501875pt}%
\definecolor{currentstroke}{rgb}{0.886275,0.898039,0.909804}%
\pgfsetstrokecolor{currentstroke}%
\pgfsetdash{}{0pt}%
\pgfpathmoveto{\pgfqpoint{1.519315in}{0.722000in}}%
\pgfpathlineto{\pgfqpoint{1.519315in}{3.230000in}}%
\pgfusepath{stroke}%
\end{pgfscope}%
\begin{pgfscope}%
\definecolor{textcolor}{rgb}{0.000000,0.000000,0.000000}%
\pgfsetstrokecolor{textcolor}%
\pgfsetfillcolor{textcolor}%
\pgftext[x=1.519315in,y=0.673389in,,top]{\color{textcolor}{\fontfamily{\familydefault}\fontsize{9.500000}{11.400000}\selectfont\catcode`\^=\active\def^{\ifmmode\sp\else\^{}\fi}\catcode`\%=\active\def
\end{pgfscope}%
\begin{pgfscope}%
\pgfpathrectangle{\pgfqpoint{1.241000in}{0.722000in}}{\pgfqpoint{2.067483in}{2.508000in}}%
\pgfusepath{clip}%
\pgfsetrectcap%
\pgfsetroundjoin%
\pgfsetlinewidth{0.501875pt}%
\definecolor{currentstroke}{rgb}{0.886275,0.898039,0.909804}%
\pgfsetstrokecolor{currentstroke}%
\pgfsetdash{}{0pt}%
\pgfpathmoveto{\pgfqpoint{1.916908in}{0.722000in}}%
\pgfpathlineto{\pgfqpoint{1.916908in}{3.230000in}}%
\pgfusepath{stroke}%
\end{pgfscope}%
\begin{pgfscope}%
\definecolor{textcolor}{rgb}{0.000000,0.000000,0.000000}%
\pgfsetstrokecolor{textcolor}%
\pgfsetfillcolor{textcolor}%
\pgftext[x=1.916908in,y=0.673389in,,top]{\color{textcolor}{\fontfamily{\familydefault}\fontsize{9.500000}{11.400000}\selectfont\catcode`\^=\active\def^{\ifmmode\sp\else\^{}\fi}\catcode`\%=\active\def
\end{pgfscope}%
\begin{pgfscope}%
\pgfpathrectangle{\pgfqpoint{1.241000in}{0.722000in}}{\pgfqpoint{2.067483in}{2.508000in}}%
\pgfusepath{clip}%
\pgfsetrectcap%
\pgfsetroundjoin%
\pgfsetlinewidth{0.501875pt}%
\definecolor{currentstroke}{rgb}{0.886275,0.898039,0.909804}%
\pgfsetstrokecolor{currentstroke}%
\pgfsetdash{}{0pt}%
\pgfpathmoveto{\pgfqpoint{2.314501in}{0.722000in}}%
\pgfpathlineto{\pgfqpoint{2.314501in}{3.230000in}}%
\pgfusepath{stroke}%
\end{pgfscope}%
\begin{pgfscope}%
\definecolor{textcolor}{rgb}{0.000000,0.000000,0.000000}%
\pgfsetstrokecolor{textcolor}%
\pgfsetfillcolor{textcolor}%
\pgftext[x=2.314501in,y=0.673389in,,top]{\color{textcolor}{\fontfamily{\familydefault}\fontsize{9.500000}{11.400000}\selectfont\catcode`\^=\active\def^{\ifmmode\sp\else\^{}\fi}\catcode`\%=\active\def
\end{pgfscope}%
\begin{pgfscope}%
\pgfpathrectangle{\pgfqpoint{1.241000in}{0.722000in}}{\pgfqpoint{2.067483in}{2.508000in}}%
\pgfusepath{clip}%
\pgfsetrectcap%
\pgfsetroundjoin%
\pgfsetlinewidth{0.501875pt}%
\definecolor{currentstroke}{rgb}{0.886275,0.898039,0.909804}%
\pgfsetstrokecolor{currentstroke}%
\pgfsetdash{}{0pt}%
\pgfpathmoveto{\pgfqpoint{2.712093in}{0.722000in}}%
\pgfpathlineto{\pgfqpoint{2.712093in}{3.230000in}}%
\pgfusepath{stroke}%
\end{pgfscope}%
\begin{pgfscope}%
\definecolor{textcolor}{rgb}{0.000000,0.000000,0.000000}%
\pgfsetstrokecolor{textcolor}%
\pgfsetfillcolor{textcolor}%
\pgftext[x=2.712093in,y=0.673389in,,top]{\color{textcolor}{\fontfamily{\familydefault}\fontsize{9.500000}{11.400000}\selectfont\catcode`\^=\active\def^{\ifmmode\sp\else\^{}\fi}\catcode`\%=\active\def
\end{pgfscope}%
\begin{pgfscope}%
\pgfpathrectangle{\pgfqpoint{1.241000in}{0.722000in}}{\pgfqpoint{2.067483in}{2.508000in}}%
\pgfusepath{clip}%
\pgfsetrectcap%
\pgfsetroundjoin%
\pgfsetlinewidth{0.501875pt}%
\definecolor{currentstroke}{rgb}{0.886275,0.898039,0.909804}%
\pgfsetstrokecolor{currentstroke}%
\pgfsetdash{}{0pt}%
\pgfpathmoveto{\pgfqpoint{3.109686in}{0.722000in}}%
\pgfpathlineto{\pgfqpoint{3.109686in}{3.230000in}}%
\pgfusepath{stroke}%
\end{pgfscope}%
\begin{pgfscope}%
\definecolor{textcolor}{rgb}{0.000000,0.000000,0.000000}%
\pgfsetstrokecolor{textcolor}%
\pgfsetfillcolor{textcolor}%
\pgftext[x=3.109686in,y=0.673389in,,top]{\color{textcolor}{\fontfamily{\familydefault}\fontsize{9.500000}{11.400000}\selectfont\catcode`\^=\active\def^{\ifmmode\sp\else\^{}\fi}\catcode`\%=\active\def
\end{pgfscope}%
\begin{pgfscope}%
\definecolor{textcolor}{rgb}{0.000000,0.000000,0.000000}%
\pgfsetstrokecolor{textcolor}%
\pgfsetfillcolor{textcolor}%
\pgftext[x=2.274741in,y=0.501311in,,top]{\color{textcolor}{\fontfamily{\familydefault}\fontsize{10.500000}{12.600000}\selectfont\catcode`\^=\active\def^{\ifmmode\sp\else\^{}\fi}\catcode`\%=\active\def
\end{pgfscope}%
\begin{pgfscope}%
\definecolor{textcolor}{rgb}{0.000000,0.000000,0.000000}%
\pgfsetstrokecolor{textcolor}%
\pgfsetfillcolor{textcolor}%
\pgftext[x=0.284699in, y=3.002158in, left, base]{\color{textcolor}{\fontfamily{\familydefault}\fontsize{9.500000}{11.400000}\selectfont\catcode`\^=\active\def^{\ifmmode\sp\else\^{}\fi}\catcode`\%=\active\def
\end{pgfscope}%
\begin{pgfscope}%
\definecolor{textcolor}{rgb}{0.000000,0.000000,0.000000}%
\pgfsetstrokecolor{textcolor}%
\pgfsetfillcolor{textcolor}%
\pgftext[x=0.348770in, y=2.867461in, left, base]{\color{textcolor}{\fontfamily{\familydefault}\fontsize{9.500000}{11.400000}\selectfont\catcode`\^=\active\def^{\ifmmode\sp\else\^{}\fi}\catcode`\%=\active\def
\end{pgfscope}%
\begin{pgfscope}%
\definecolor{textcolor}{rgb}{0.000000,0.000000,0.000000}%
\pgfsetstrokecolor{textcolor}%
\pgfsetfillcolor{textcolor}%
\pgftext[x=0.184585in, y=2.495491in, left, base]{\color{textcolor}{\fontfamily{\familydefault}\fontsize{9.500000}{11.400000}\selectfont\catcode`\^=\active\def^{\ifmmode\sp\else\^{}\fi}\catcode`\%=\active\def
\end{pgfscope}%
\begin{pgfscope}%
\definecolor{textcolor}{rgb}{0.000000,0.000000,0.000000}%
\pgfsetstrokecolor{textcolor}%
\pgfsetfillcolor{textcolor}%
\pgftext[x=0.348770in, y=2.360794in, left, base]{\color{textcolor}{\fontfamily{\familydefault}\fontsize{9.500000}{11.400000}\selectfont\catcode`\^=\active\def^{\ifmmode\sp\else\^{}\fi}\catcode`\%=\active\def
\end{pgfscope}%
\begin{pgfscope}%
\definecolor{textcolor}{rgb}{0.000000,0.000000,0.000000}%
\pgfsetstrokecolor{textcolor}%
\pgfsetfillcolor{textcolor}%
\pgftext[x=0.244841in, y=1.988825in, left, base]{\color{textcolor}{\fontfamily{\familydefault}\fontsize{9.500000}{11.400000}\selectfont\catcode`\^=\active\def^{\ifmmode\sp\else\^{}\fi}\catcode`\%=\active\def
\end{pgfscope}%
\begin{pgfscope}%
\definecolor{textcolor}{rgb}{0.000000,0.000000,0.000000}%
\pgfsetstrokecolor{textcolor}%
\pgfsetfillcolor{textcolor}%
\pgftext[x=0.348770in, y=1.854128in, left, base]{\color{textcolor}{\fontfamily{\familydefault}\fontsize{9.500000}{11.400000}\selectfont\catcode`\^=\active\def^{\ifmmode\sp\else\^{}\fi}\catcode`\%=\active\def
\end{pgfscope}%
\begin{pgfscope}%
\definecolor{textcolor}{rgb}{0.000000,0.000000,0.000000}%
\pgfsetstrokecolor{textcolor}%
\pgfsetfillcolor{textcolor}%
\pgftext[x=0.101987in, y=1.482158in, left, base]{\color{textcolor}{\fontfamily{\familydefault}\fontsize{9.500000}{11.400000}\selectfont\catcode`\^=\active\def^{\ifmmode\sp\else\^{}\fi}\catcode`\%=\active\def
\end{pgfscope}%
\begin{pgfscope}%
\definecolor{textcolor}{rgb}{0.000000,0.000000,0.000000}%
\pgfsetstrokecolor{textcolor}%
\pgfsetfillcolor{textcolor}%
\pgftext[x=0.416558in, y=1.347461in, left, base]{\color{textcolor}{\fontfamily{\familydefault}\fontsize{9.500000}{11.400000}\selectfont\catcode`\^=\active\def^{\ifmmode\sp\else\^{}\fi}\catcode`\%=\active\def
\end{pgfscope}%
\begin{pgfscope}%
\definecolor{textcolor}{rgb}{0.000000,0.000000,0.000000}%
\pgfsetstrokecolor{textcolor}%
\pgfsetfillcolor{textcolor}%
\pgftext[x=-0.003726in, y=0.975491in, left, base]{\color{textcolor}{\fontfamily{\familydefault}\fontsize{9.500000}{11.400000}\selectfont\catcode`\^=\active\def^{\ifmmode\sp\else\^{}\fi}\catcode`\%=\active\def
\end{pgfscope}%
\begin{pgfscope}%
\definecolor{textcolor}{rgb}{0.000000,0.000000,0.000000}%
\pgfsetstrokecolor{textcolor}%
\pgfsetfillcolor{textcolor}%
\pgftext[x=0.348770in, y=0.840794in, left, base]{\color{textcolor}{\fontfamily{\familydefault}\fontsize{9.500000}{11.400000}\selectfont\catcode`\^=\active\def^{\ifmmode\sp\else\^{}\fi}\catcode`\%=\active\def
\end{pgfscope}%
\begin{pgfscope}%
\pgfpathrectangle{\pgfqpoint{1.241000in}{0.722000in}}{\pgfqpoint{2.067483in}{2.508000in}}%
\pgfusepath{clip}%
\pgfsetbuttcap%
\pgfsetmiterjoin%
\definecolor{currentfill}{rgb}{0.094118,0.490196,0.513725}%
\pgfsetfillcolor{currentfill}%
\pgfsetlinewidth{0.000000pt}%
\definecolor{currentstroke}{rgb}{0.000000,0.000000,0.000000}%
\pgfsetstrokecolor{currentstroke}%
\pgfsetstrokeopacity{0.000000}%
\pgfsetdash{}{0pt}%
\pgfpathmoveto{\pgfqpoint{1.916908in}{3.116000in}}%
\pgfpathlineto{\pgfqpoint{1.946727in}{3.116000in}}%
\pgfpathlineto{\pgfqpoint{1.946727in}{2.862667in}}%
\pgfpathlineto{\pgfqpoint{1.916908in}{2.862667in}}%
\pgfpathlineto{\pgfqpoint{1.916908in}{3.116000in}}%
\pgfpathclose%
\pgfusepath{fill}%
\end{pgfscope}%
\begin{pgfscope}%
\pgfpathrectangle{\pgfqpoint{1.241000in}{0.722000in}}{\pgfqpoint{2.067483in}{2.508000in}}%
\pgfusepath{clip}%
\pgfsetbuttcap%
\pgfsetmiterjoin%
\definecolor{currentfill}{rgb}{0.094118,0.490196,0.513725}%
\pgfsetfillcolor{currentfill}%
\pgfsetlinewidth{0.000000pt}%
\definecolor{currentstroke}{rgb}{0.000000,0.000000,0.000000}%
\pgfsetstrokecolor{currentstroke}%
\pgfsetstrokeopacity{0.000000}%
\pgfsetdash{}{0pt}%
\pgfpathmoveto{\pgfqpoint{1.916908in}{2.609333in}}%
\pgfpathlineto{\pgfqpoint{2.182102in}{2.609333in}}%
\pgfpathlineto{\pgfqpoint{2.182102in}{2.356000in}}%
\pgfpathlineto{\pgfqpoint{1.916908in}{2.356000in}}%
\pgfpathlineto{\pgfqpoint{1.916908in}{2.609333in}}%
\pgfpathclose%
\pgfusepath{fill}%
\end{pgfscope}%
\begin{pgfscope}%
\pgfpathrectangle{\pgfqpoint{1.241000in}{0.722000in}}{\pgfqpoint{2.067483in}{2.508000in}}%
\pgfusepath{clip}%
\pgfsetbuttcap%
\pgfsetmiterjoin%
\definecolor{currentfill}{rgb}{0.094118,0.490196,0.513725}%
\pgfsetfillcolor{currentfill}%
\pgfsetlinewidth{0.000000pt}%
\definecolor{currentstroke}{rgb}{0.000000,0.000000,0.000000}%
\pgfsetstrokecolor{currentstroke}%
\pgfsetstrokeopacity{0.000000}%
\pgfsetdash{}{0pt}%
\pgfpathmoveto{\pgfqpoint{1.916908in}{2.102667in}}%
\pgfpathlineto{\pgfqpoint{2.428610in}{2.102667in}}%
\pgfpathlineto{\pgfqpoint{2.428610in}{1.849333in}}%
\pgfpathlineto{\pgfqpoint{1.916908in}{1.849333in}}%
\pgfpathlineto{\pgfqpoint{1.916908in}{2.102667in}}%
\pgfpathclose%
\pgfusepath{fill}%
\end{pgfscope}%
\begin{pgfscope}%
\pgfpathrectangle{\pgfqpoint{1.241000in}{0.722000in}}{\pgfqpoint{2.067483in}{2.508000in}}%
\pgfusepath{clip}%
\pgfsetbuttcap%
\pgfsetmiterjoin%
\definecolor{currentfill}{rgb}{0.094118,0.490196,0.513725}%
\pgfsetfillcolor{currentfill}%
\pgfsetlinewidth{0.000000pt}%
\definecolor{currentstroke}{rgb}{0.000000,0.000000,0.000000}%
\pgfsetstrokecolor{currentstroke}%
\pgfsetstrokeopacity{0.000000}%
\pgfsetdash{}{0pt}%
\pgfpathmoveto{\pgfqpoint{1.916908in}{1.596000in}}%
\pgfpathlineto{\pgfqpoint{2.986432in}{1.596000in}}%
\pgfpathlineto{\pgfqpoint{2.986432in}{1.342667in}}%
\pgfpathlineto{\pgfqpoint{1.916908in}{1.342667in}}%
\pgfpathlineto{\pgfqpoint{1.916908in}{1.596000in}}%
\pgfpathclose%
\pgfusepath{fill}%
\end{pgfscope}%
\begin{pgfscope}%
\pgfpathrectangle{\pgfqpoint{1.241000in}{0.722000in}}{\pgfqpoint{2.067483in}{2.508000in}}%
\pgfusepath{clip}%
\pgfsetbuttcap%
\pgfsetmiterjoin%
\definecolor{currentfill}{rgb}{0.713725,0.392157,0.239216}%
\pgfsetfillcolor{currentfill}%
\pgfsetlinewidth{0.000000pt}%
\definecolor{currentstroke}{rgb}{0.000000,0.000000,0.000000}%
\pgfsetstrokecolor{currentstroke}%
\pgfsetstrokeopacity{0.000000}%
\pgfsetdash{}{0pt}%
\pgfpathmoveto{\pgfqpoint{1.916908in}{1.089333in}}%
\pgfpathlineto{\pgfqpoint{1.495459in}{1.089333in}}%
\pgfpathlineto{\pgfqpoint{1.495459in}{0.836000in}}%
\pgfpathlineto{\pgfqpoint{1.916908in}{0.836000in}}%
\pgfpathlineto{\pgfqpoint{1.916908in}{1.089333in}}%
\pgfpathclose%
\pgfusepath{fill}%
\end{pgfscope}%
\begin{pgfscope}%
\pgfpathrectangle{\pgfqpoint{1.241000in}{0.722000in}}{\pgfqpoint{2.067483in}{2.508000in}}%
\pgfusepath{clip}%
\pgfsetrectcap%
\pgfsetroundjoin%
\pgfsetlinewidth{0.702625pt}%
\definecolor{currentstroke}{rgb}{0.384314,0.423529,0.478431}%
\pgfsetstrokecolor{currentstroke}%
\pgfsetdash{}{0pt}%
\pgfpathmoveto{\pgfqpoint{1.916908in}{0.722000in}}%
\pgfpathlineto{\pgfqpoint{1.916908in}{3.230000in}}%
\pgfusepath{stroke}%
\end{pgfscope}%
\begin{pgfscope}%
\pgfsetrectcap%
\pgfsetmiterjoin%
\pgfsetlinewidth{0.803000pt}%
\definecolor{currentstroke}{rgb}{0.713725,0.741176,0.768627}%
\pgfsetstrokecolor{currentstroke}%
\pgfsetdash{}{0pt}%
\pgfpathmoveto{\pgfqpoint{1.241000in}{0.722000in}}%
\pgfpathlineto{\pgfqpoint{3.308483in}{0.722000in}}%
\pgfusepath{stroke}%
\end{pgfscope}%
\begin{pgfscope}%
\definecolor{textcolor}{rgb}{0.000000,0.000000,0.000000}%
\pgfsetstrokecolor{textcolor}%
\pgfsetfillcolor{textcolor}%
\pgftext[x=1.241000in,y=3.280160in,left,base]{\color{textcolor}{\fontfamily{\familydefault}\fontsize{8.500000}{10.200000}\selectfont\catcode`\^=\active\def^{\ifmmode\sp\else\^{}\fi}\catcode`\%=\active\def
\end{pgfscope}%
\begin{pgfscope}%
\definecolor{textcolor}{rgb}{0.000000,0.000000,0.000000}%
\pgfsetstrokecolor{textcolor}%
\pgfsetfillcolor{textcolor}%
\pgftext[x=1.978535in,y=2.989333in,left,]{\color{textcolor}{\fontfamily{\familydefault}\fontsize{9.500000}{11.400000}\selectfont\catcode`\^=\active\def^{\ifmmode\sp\else\^{}\fi}\catcode`\%=\active\def
\end{pgfscope}%
\begin{pgfscope}%
\definecolor{textcolor}{rgb}{0.000000,0.000000,0.000000}%
\pgfsetstrokecolor{textcolor}%
\pgfsetfillcolor{textcolor}%
\pgftext[x=2.213910in,y=2.482667in,left,]{\color{textcolor}{\fontfamily{\familydefault}\fontsize{9.500000}{11.400000}\selectfont\catcode`\^=\active\def^{\ifmmode\sp\else\^{}\fi}\catcode`\%=\active\def
\end{pgfscope}%
\begin{pgfscope}%
\definecolor{textcolor}{rgb}{0.000000,0.000000,0.000000}%
\pgfsetstrokecolor{textcolor}%
\pgfsetfillcolor{textcolor}%
\pgftext[x=2.460417in,y=1.976000in,left,]{\color{textcolor}{\fontfamily{\familydefault}\fontsize{9.500000}{11.400000}\selectfont\catcode`\^=\active\def^{\ifmmode\sp\else\^{}\fi}\catcode`\%=\active\def
\end{pgfscope}%
\begin{pgfscope}%
\definecolor{textcolor}{rgb}{0.000000,0.000000,0.000000}%
\pgfsetstrokecolor{textcolor}%
\pgfsetfillcolor{textcolor}%
\pgftext[x=3.018240in,y=1.469333in,left,]{\color{textcolor}{\fontfamily{\familydefault}\fontsize{9.500000}{11.400000}\selectfont\catcode`\^=\active\def^{\ifmmode\sp\else\^{}\fi}\catcode`\%=\active\def
\end{pgfscope}%
\begin{pgfscope}%
\definecolor{textcolor}{rgb}{0.713725,0.392157,0.239216}%
\pgfsetstrokecolor{textcolor}%
\pgfsetfillcolor{textcolor}%
\pgftext[x=1.948715in,y=0.962667in,left,]{\color{textcolor}{\fontfamily{\familydefault}\fontsize{9.500000}{11.400000}\selectfont\catcode`\^=\active\def^{\ifmmode\sp\else\^{}\fi}\catcode`\%=\active\def
\end{pgfscope}%
\begin{pgfscope}%
\definecolor{textcolor}{rgb}{0.000000,0.000000,0.000000}%
\pgfsetstrokecolor{textcolor}%
\pgfsetfillcolor{textcolor}%
\pgftext[x=1.241000in,y=3.438333in,left,base]{\color{textcolor}{\fontfamily{\familydefault}\fontsize{11.500000}{13.800000}\selectfont\catcode`\^=\active\def^{\ifmmode\sp\else\^{}\fi}\catcode`\%=\active\def
\end{pgfscope}%
\begin{pgfscope}%
\pgfsetbuttcap%
\pgfsetmiterjoin%
\definecolor{currentfill}{rgb}{1.000000,1.000000,1.000000}%
\pgfsetfillcolor{currentfill}%
\pgfsetlinewidth{0.000000pt}%
\definecolor{currentstroke}{rgb}{0.000000,0.000000,0.000000}%
\pgfsetstrokecolor{currentstroke}%
\pgfsetstrokeopacity{0.000000}%
\pgfsetdash{}{0pt}%
\pgfpathmoveto{\pgfqpoint{5.086517in}{0.722000in}}%
\pgfpathlineto{\pgfqpoint{7.154000in}{0.722000in}}%
\pgfpathlineto{\pgfqpoint{7.154000in}{3.230000in}}%
\pgfpathlineto{\pgfqpoint{5.086517in}{3.230000in}}%
\pgfpathlineto{\pgfqpoint{5.086517in}{0.722000in}}%
\pgfpathclose%
\pgfusepath{fill}%
\end{pgfscope}%
\begin{pgfscope}%
\pgfpathrectangle{\pgfqpoint{5.086517in}{0.722000in}}{\pgfqpoint{2.067483in}{2.508000in}}%
\pgfusepath{clip}%
\pgfsetrectcap%
\pgfsetroundjoin%
\pgfsetlinewidth{0.501875pt}%
\definecolor{currentstroke}{rgb}{0.886275,0.898039,0.909804}%
\pgfsetstrokecolor{currentstroke}%
\pgfsetdash{}{0pt}%
\pgfpathmoveto{\pgfqpoint{5.086517in}{0.722000in}}%
\pgfpathlineto{\pgfqpoint{5.086517in}{3.230000in}}%
\pgfusepath{stroke}%
\end{pgfscope}%
\begin{pgfscope}%
\definecolor{textcolor}{rgb}{0.000000,0.000000,0.000000}%
\pgfsetstrokecolor{textcolor}%
\pgfsetfillcolor{textcolor}%
\pgftext[x=5.086517in,y=0.673389in,,top]{\color{textcolor}{\fontfamily{\familydefault}\fontsize{9.500000}{11.400000}\selectfont\catcode`\^=\active\def^{\ifmmode\sp\else\^{}\fi}\catcode`\%=\active\def
\end{pgfscope}%
\begin{pgfscope}%
\pgfpathrectangle{\pgfqpoint{5.086517in}{0.722000in}}{\pgfqpoint{2.067483in}{2.508000in}}%
\pgfusepath{clip}%
\pgfsetrectcap%
\pgfsetroundjoin%
\pgfsetlinewidth{0.501875pt}%
\definecolor{currentstroke}{rgb}{0.886275,0.898039,0.909804}%
\pgfsetstrokecolor{currentstroke}%
\pgfsetdash{}{0pt}%
\pgfpathmoveto{\pgfqpoint{6.017816in}{0.722000in}}%
\pgfpathlineto{\pgfqpoint{6.017816in}{3.230000in}}%
\pgfusepath{stroke}%
\end{pgfscope}%
\begin{pgfscope}%
\definecolor{textcolor}{rgb}{0.000000,0.000000,0.000000}%
\pgfsetstrokecolor{textcolor}%
\pgfsetfillcolor{textcolor}%
\pgftext[x=6.017816in,y=0.673389in,,top]{\color{textcolor}{\fontfamily{\familydefault}\fontsize{9.500000}{11.400000}\selectfont\catcode`\^=\active\def^{\ifmmode\sp\else\^{}\fi}\catcode`\%=\active\def
\end{pgfscope}%
\begin{pgfscope}%
\pgfpathrectangle{\pgfqpoint{5.086517in}{0.722000in}}{\pgfqpoint{2.067483in}{2.508000in}}%
\pgfusepath{clip}%
\pgfsetrectcap%
\pgfsetroundjoin%
\pgfsetlinewidth{0.501875pt}%
\definecolor{currentstroke}{rgb}{0.886275,0.898039,0.909804}%
\pgfsetstrokecolor{currentstroke}%
\pgfsetdash{}{0pt}%
\pgfpathmoveto{\pgfqpoint{6.949114in}{0.722000in}}%
\pgfpathlineto{\pgfqpoint{6.949114in}{3.230000in}}%
\pgfusepath{stroke}%
\end{pgfscope}%
\begin{pgfscope}%
\definecolor{textcolor}{rgb}{0.000000,0.000000,0.000000}%
\pgfsetstrokecolor{textcolor}%
\pgfsetfillcolor{textcolor}%
\pgftext[x=6.949114in,y=0.673389in,,top]{\color{textcolor}{\fontfamily{\familydefault}\fontsize{9.500000}{11.400000}\selectfont\catcode`\^=\active\def^{\ifmmode\sp\else\^{}\fi}\catcode`\%=\active\def
\end{pgfscope}%
\begin{pgfscope}%
\definecolor{textcolor}{rgb}{0.000000,0.000000,0.000000}%
\pgfsetstrokecolor{textcolor}%
\pgfsetfillcolor{textcolor}%
\pgftext[x=6.120259in,y=0.501311in,,top]{\color{textcolor}{\fontfamily{\familydefault}\fontsize{10.500000}{12.600000}\selectfont\catcode`\^=\active\def^{\ifmmode\sp\else\^{}\fi}\catcode`\%=\active\def
\end{pgfscope}%
\begin{pgfscope}%
\definecolor{textcolor}{rgb}{0.000000,0.000000,0.000000}%
\pgfsetstrokecolor{textcolor}%
\pgfsetfillcolor{textcolor}%
\pgftext[x=4.521882in, y=2.970981in, left, base]{\color{textcolor}{\fontfamily{\familydefault}\fontsize{9.500000}{11.400000}\selectfont\catcode`\^=\active\def^{\ifmmode\sp\else\^{}\fi}\catcode`\%=\active\def
\end{pgfscope}%
\begin{pgfscope}%
\definecolor{textcolor}{rgb}{0.000000,0.000000,0.000000}%
\pgfsetstrokecolor{textcolor}%
\pgfsetfillcolor{textcolor}%
\pgftext[x=4.587523in, y=2.673719in, left, base]{\color{textcolor}{\fontfamily{\familydefault}\fontsize{9.500000}{11.400000}\selectfont\catcode`\^=\active\def^{\ifmmode\sp\else\^{}\fi}\catcode`\%=\active\def
\end{pgfscope}%
\begin{pgfscope}%
\definecolor{textcolor}{rgb}{0.000000,0.000000,0.000000}%
\pgfsetstrokecolor{textcolor}%
\pgfsetfillcolor{textcolor}%
\pgftext[x=4.056465in, y=2.376457in, left, base]{\color{textcolor}{\fontfamily{\familydefault}\fontsize{9.500000}{11.400000}\selectfont\catcode`\^=\active\def^{\ifmmode\sp\else\^{}\fi}\catcode`\%=\active\def
\end{pgfscope}%
\begin{pgfscope}%
\definecolor{textcolor}{rgb}{0.000000,0.000000,0.000000}%
\pgfsetstrokecolor{textcolor}%
\pgfsetfillcolor{textcolor}%
\pgftext[x=4.305021in, y=2.079195in, left, base]{\color{textcolor}{\fontfamily{\familydefault}\fontsize{9.500000}{11.400000}\selectfont\catcode`\^=\active\def^{\ifmmode\sp\else\^{}\fi}\catcode`\%=\active\def
\end{pgfscope}%
\begin{pgfscope}%
\definecolor{textcolor}{rgb}{0.000000,0.000000,0.000000}%
\pgfsetstrokecolor{textcolor}%
\pgfsetfillcolor{textcolor}%
\pgftext[x=3.773963in, y=1.781933in, left, base]{\color{textcolor}{\fontfamily{\familydefault}\fontsize{9.500000}{11.400000}\selectfont\catcode`\^=\active\def^{\ifmmode\sp\else\^{}\fi}\catcode`\%=\active\def
\end{pgfscope}%
\begin{pgfscope}%
\definecolor{textcolor}{rgb}{0.000000,0.000000,0.000000}%
\pgfsetstrokecolor{textcolor}%
\pgfsetfillcolor{textcolor}%
\pgftext[x=4.158503in, y=1.484670in, left, base]{\color{textcolor}{\fontfamily{\familydefault}\fontsize{9.500000}{11.400000}\selectfont\catcode`\^=\active\def^{\ifmmode\sp\else\^{}\fi}\catcode`\%=\active\def
\end{pgfscope}%
\begin{pgfscope}%
\definecolor{textcolor}{rgb}{0.000000,0.000000,0.000000}%
\pgfsetstrokecolor{textcolor}%
\pgfsetfillcolor{textcolor}%
\pgftext[x=4.536997in, y=1.187408in, left, base]{\color{textcolor}{\fontfamily{\familydefault}\fontsize{9.500000}{11.400000}\selectfont\catcode`\^=\active\def^{\ifmmode\sp\else\^{}\fi}\catcode`\%=\active\def
\end{pgfscope}%
\begin{pgfscope}%
\definecolor{textcolor}{rgb}{0.000000,0.000000,0.000000}%
\pgfsetstrokecolor{textcolor}%
\pgfsetfillcolor{textcolor}%
\pgftext[x=4.106614in, y=0.890146in, left, base]{\color{textcolor}{\fontfamily{\familydefault}\fontsize{9.500000}{11.400000}\selectfont\catcode`\^=\active\def^{\ifmmode\sp\else\^{}\fi}\catcode`\%=\active\def
\end{pgfscope}%
\begin{pgfscope}%
\pgfpathrectangle{\pgfqpoint{5.086517in}{0.722000in}}{\pgfqpoint{2.067483in}{2.508000in}}%
\pgfusepath{clip}%
\pgfsetbuttcap%
\pgfsetmiterjoin%
\definecolor{currentfill}{rgb}{0.384314,0.423529,0.478431}%
\pgfsetfillcolor{currentfill}%
\pgfsetlinewidth{0.000000pt}%
\definecolor{currentstroke}{rgb}{0.000000,0.000000,0.000000}%
\pgfsetstrokecolor{currentstroke}%
\pgfsetstrokeopacity{0.000000}%
\pgfsetdash{}{0pt}%
\pgfpathmoveto{\pgfqpoint{5.086517in}{3.116000in}}%
\pgfpathlineto{\pgfqpoint{6.088781in}{3.116000in}}%
\pgfpathlineto{\pgfqpoint{6.088781in}{3.029794in}}%
\pgfpathlineto{\pgfqpoint{5.086517in}{3.029794in}}%
\pgfpathlineto{\pgfqpoint{5.086517in}{3.116000in}}%
\pgfpathclose%
\pgfusepath{fill}%
\end{pgfscope}%
\begin{pgfscope}%
\pgfpathrectangle{\pgfqpoint{5.086517in}{0.722000in}}{\pgfqpoint{2.067483in}{2.508000in}}%
\pgfusepath{clip}%
\pgfsetbuttcap%
\pgfsetmiterjoin%
\definecolor{currentfill}{rgb}{0.384314,0.423529,0.478431}%
\pgfsetfillcolor{currentfill}%
\pgfsetlinewidth{0.000000pt}%
\definecolor{currentstroke}{rgb}{0.000000,0.000000,0.000000}%
\pgfsetstrokecolor{currentstroke}%
\pgfsetstrokeopacity{0.000000}%
\pgfsetdash{}{0pt}%
\pgfpathmoveto{\pgfqpoint{5.086517in}{2.818738in}}%
\pgfpathlineto{\pgfqpoint{6.381022in}{2.818738in}}%
\pgfpathlineto{\pgfqpoint{6.381022in}{2.732532in}}%
\pgfpathlineto{\pgfqpoint{5.086517in}{2.732532in}}%
\pgfpathlineto{\pgfqpoint{5.086517in}{2.818738in}}%
\pgfpathclose%
\pgfusepath{fill}%
\end{pgfscope}%
\begin{pgfscope}%
\pgfpathrectangle{\pgfqpoint{5.086517in}{0.722000in}}{\pgfqpoint{2.067483in}{2.508000in}}%
\pgfusepath{clip}%
\pgfsetbuttcap%
\pgfsetmiterjoin%
\definecolor{currentfill}{rgb}{0.384314,0.423529,0.478431}%
\pgfsetfillcolor{currentfill}%
\pgfsetlinewidth{0.000000pt}%
\definecolor{currentstroke}{rgb}{0.000000,0.000000,0.000000}%
\pgfsetstrokecolor{currentstroke}%
\pgfsetstrokeopacity{0.000000}%
\pgfsetdash{}{0pt}%
\pgfpathmoveto{\pgfqpoint{5.086517in}{2.521476in}}%
\pgfpathlineto{\pgfqpoint{6.576595in}{2.521476in}}%
\pgfpathlineto{\pgfqpoint{6.576595in}{2.435270in}}%
\pgfpathlineto{\pgfqpoint{5.086517in}{2.435270in}}%
\pgfpathlineto{\pgfqpoint{5.086517in}{2.521476in}}%
\pgfpathclose%
\pgfusepath{fill}%
\end{pgfscope}%
\begin{pgfscope}%
\pgfpathrectangle{\pgfqpoint{5.086517in}{0.722000in}}{\pgfqpoint{2.067483in}{2.508000in}}%
\pgfusepath{clip}%
\pgfsetbuttcap%
\pgfsetmiterjoin%
\definecolor{currentfill}{rgb}{0.384314,0.423529,0.478431}%
\pgfsetfillcolor{currentfill}%
\pgfsetlinewidth{0.000000pt}%
\definecolor{currentstroke}{rgb}{0.000000,0.000000,0.000000}%
\pgfsetstrokecolor{currentstroke}%
\pgfsetstrokeopacity{0.000000}%
\pgfsetdash{}{0pt}%
\pgfpathmoveto{\pgfqpoint{5.086517in}{2.224214in}}%
\pgfpathlineto{\pgfqpoint{6.483465in}{2.224214in}}%
\pgfpathlineto{\pgfqpoint{6.483465in}{2.138008in}}%
\pgfpathlineto{\pgfqpoint{5.086517in}{2.138008in}}%
\pgfpathlineto{\pgfqpoint{5.086517in}{2.224214in}}%
\pgfpathclose%
\pgfusepath{fill}%
\end{pgfscope}%
\begin{pgfscope}%
\pgfpathrectangle{\pgfqpoint{5.086517in}{0.722000in}}{\pgfqpoint{2.067483in}{2.508000in}}%
\pgfusepath{clip}%
\pgfsetbuttcap%
\pgfsetmiterjoin%
\definecolor{currentfill}{rgb}{0.384314,0.423529,0.478431}%
\pgfsetfillcolor{currentfill}%
\pgfsetlinewidth{0.000000pt}%
\definecolor{currentstroke}{rgb}{0.000000,0.000000,0.000000}%
\pgfsetstrokecolor{currentstroke}%
\pgfsetstrokeopacity{0.000000}%
\pgfsetdash{}{0pt}%
\pgfpathmoveto{\pgfqpoint{5.086517in}{1.926952in}}%
\pgfpathlineto{\pgfqpoint{6.483465in}{1.926952in}}%
\pgfpathlineto{\pgfqpoint{6.483465in}{1.840746in}}%
\pgfpathlineto{\pgfqpoint{5.086517in}{1.840746in}}%
\pgfpathlineto{\pgfqpoint{5.086517in}{1.926952in}}%
\pgfpathclose%
\pgfusepath{fill}%
\end{pgfscope}%
\begin{pgfscope}%
\pgfpathrectangle{\pgfqpoint{5.086517in}{0.722000in}}{\pgfqpoint{2.067483in}{2.508000in}}%
\pgfusepath{clip}%
\pgfsetbuttcap%
\pgfsetmiterjoin%
\definecolor{currentfill}{rgb}{0.384314,0.423529,0.478431}%
\pgfsetfillcolor{currentfill}%
\pgfsetlinewidth{0.000000pt}%
\definecolor{currentstroke}{rgb}{0.000000,0.000000,0.000000}%
\pgfsetstrokecolor{currentstroke}%
\pgfsetstrokeopacity{0.000000}%
\pgfsetdash{}{0pt}%
\pgfpathmoveto{\pgfqpoint{5.086517in}{1.629690in}}%
\pgfpathlineto{\pgfqpoint{6.703252in}{1.629690in}}%
\pgfpathlineto{\pgfqpoint{6.703252in}{1.543484in}}%
\pgfpathlineto{\pgfqpoint{5.086517in}{1.543484in}}%
\pgfpathlineto{\pgfqpoint{5.086517in}{1.629690in}}%
\pgfpathclose%
\pgfusepath{fill}%
\end{pgfscope}%
\begin{pgfscope}%
\pgfpathrectangle{\pgfqpoint{5.086517in}{0.722000in}}{\pgfqpoint{2.067483in}{2.508000in}}%
\pgfusepath{clip}%
\pgfsetbuttcap%
\pgfsetmiterjoin%
\definecolor{currentfill}{rgb}{0.384314,0.423529,0.478431}%
\pgfsetfillcolor{currentfill}%
\pgfsetlinewidth{0.000000pt}%
\definecolor{currentstroke}{rgb}{0.000000,0.000000,0.000000}%
\pgfsetstrokecolor{currentstroke}%
\pgfsetstrokeopacity{0.000000}%
\pgfsetdash{}{0pt}%
\pgfpathmoveto{\pgfqpoint{5.086517in}{1.332428in}}%
\pgfpathlineto{\pgfqpoint{6.595221in}{1.332428in}}%
\pgfpathlineto{\pgfqpoint{6.595221in}{1.246222in}}%
\pgfpathlineto{\pgfqpoint{5.086517in}{1.246222in}}%
\pgfpathlineto{\pgfqpoint{5.086517in}{1.332428in}}%
\pgfpathclose%
\pgfusepath{fill}%
\end{pgfscope}%
\begin{pgfscope}%
\pgfpathrectangle{\pgfqpoint{5.086517in}{0.722000in}}{\pgfqpoint{2.067483in}{2.508000in}}%
\pgfusepath{clip}%
\pgfsetbuttcap%
\pgfsetmiterjoin%
\definecolor{currentfill}{rgb}{0.384314,0.423529,0.478431}%
\pgfsetfillcolor{currentfill}%
\pgfsetlinewidth{0.000000pt}%
\definecolor{currentstroke}{rgb}{0.000000,0.000000,0.000000}%
\pgfsetstrokecolor{currentstroke}%
\pgfsetstrokeopacity{0.000000}%
\pgfsetdash{}{0pt}%
\pgfpathmoveto{\pgfqpoint{5.086517in}{1.035166in}}%
\pgfpathlineto{\pgfqpoint{5.719800in}{1.035166in}}%
\pgfpathlineto{\pgfqpoint{5.719800in}{0.948960in}}%
\pgfpathlineto{\pgfqpoint{5.086517in}{0.948960in}}%
\pgfpathlineto{\pgfqpoint{5.086517in}{1.035166in}}%
\pgfpathclose%
\pgfusepath{fill}%
\end{pgfscope}%
\begin{pgfscope}%
\pgfpathrectangle{\pgfqpoint{5.086517in}{0.722000in}}{\pgfqpoint{2.067483in}{2.508000in}}%
\pgfusepath{clip}%
\pgfsetbuttcap%
\pgfsetmiterjoin%
\definecolor{currentfill}{rgb}{0.094118,0.490196,0.513725}%
\pgfsetfillcolor{currentfill}%
\pgfsetlinewidth{0.000000pt}%
\definecolor{currentstroke}{rgb}{0.000000,0.000000,0.000000}%
\pgfsetstrokecolor{currentstroke}%
\pgfsetstrokeopacity{0.000000}%
\pgfsetdash{}{0pt}%
\pgfpathmoveto{\pgfqpoint{5.086517in}{3.003040in}}%
\pgfpathlineto{\pgfqpoint{6.039422in}{3.003040in}}%
\pgfpathlineto{\pgfqpoint{6.039422in}{2.916834in}}%
\pgfpathlineto{\pgfqpoint{5.086517in}{2.916834in}}%
\pgfpathlineto{\pgfqpoint{5.086517in}{3.003040in}}%
\pgfpathclose%
\pgfusepath{fill}%
\end{pgfscope}%
\begin{pgfscope}%
\pgfpathrectangle{\pgfqpoint{5.086517in}{0.722000in}}{\pgfqpoint{2.067483in}{2.508000in}}%
\pgfusepath{clip}%
\pgfsetbuttcap%
\pgfsetmiterjoin%
\definecolor{currentfill}{rgb}{0.094118,0.490196,0.513725}%
\pgfsetfillcolor{currentfill}%
\pgfsetlinewidth{0.000000pt}%
\definecolor{currentstroke}{rgb}{0.000000,0.000000,0.000000}%
\pgfsetstrokecolor{currentstroke}%
\pgfsetstrokeopacity{0.000000}%
\pgfsetdash{}{0pt}%
\pgfpathmoveto{\pgfqpoint{5.086517in}{2.705778in}}%
\pgfpathlineto{\pgfqpoint{5.095830in}{2.705778in}}%
\pgfpathlineto{\pgfqpoint{5.095830in}{2.619572in}}%
\pgfpathlineto{\pgfqpoint{5.086517in}{2.619572in}}%
\pgfpathlineto{\pgfqpoint{5.086517in}{2.705778in}}%
\pgfpathclose%
\pgfusepath{fill}%
\end{pgfscope}%
\begin{pgfscope}%
\pgfpathrectangle{\pgfqpoint{5.086517in}{0.722000in}}{\pgfqpoint{2.067483in}{2.508000in}}%
\pgfusepath{clip}%
\pgfsetbuttcap%
\pgfsetmiterjoin%
\definecolor{currentfill}{rgb}{0.094118,0.490196,0.513725}%
\pgfsetfillcolor{currentfill}%
\pgfsetlinewidth{0.000000pt}%
\definecolor{currentstroke}{rgb}{0.000000,0.000000,0.000000}%
\pgfsetstrokecolor{currentstroke}%
\pgfsetstrokeopacity{0.000000}%
\pgfsetdash{}{0pt}%
\pgfpathmoveto{\pgfqpoint{5.086517in}{2.408516in}}%
\pgfpathlineto{\pgfqpoint{5.095830in}{2.408516in}}%
\pgfpathlineto{\pgfqpoint{5.095830in}{2.322310in}}%
\pgfpathlineto{\pgfqpoint{5.086517in}{2.322310in}}%
\pgfpathlineto{\pgfqpoint{5.086517in}{2.408516in}}%
\pgfpathclose%
\pgfusepath{fill}%
\end{pgfscope}%
\begin{pgfscope}%
\pgfpathrectangle{\pgfqpoint{5.086517in}{0.722000in}}{\pgfqpoint{2.067483in}{2.508000in}}%
\pgfusepath{clip}%
\pgfsetbuttcap%
\pgfsetmiterjoin%
\definecolor{currentfill}{rgb}{0.094118,0.490196,0.513725}%
\pgfsetfillcolor{currentfill}%
\pgfsetlinewidth{0.000000pt}%
\definecolor{currentstroke}{rgb}{0.000000,0.000000,0.000000}%
\pgfsetstrokecolor{currentstroke}%
\pgfsetstrokeopacity{0.000000}%
\pgfsetdash{}{0pt}%
\pgfpathmoveto{\pgfqpoint{5.086517in}{2.111254in}}%
\pgfpathlineto{\pgfqpoint{5.086517in}{2.111254in}}%
\pgfpathlineto{\pgfqpoint{5.086517in}{2.025048in}}%
\pgfpathlineto{\pgfqpoint{5.086517in}{2.025048in}}%
\pgfpathlineto{\pgfqpoint{5.086517in}{2.111254in}}%
\pgfpathclose%
\pgfusepath{fill}%
\end{pgfscope}%
\begin{pgfscope}%
\pgfpathrectangle{\pgfqpoint{5.086517in}{0.722000in}}{\pgfqpoint{2.067483in}{2.508000in}}%
\pgfusepath{clip}%
\pgfsetbuttcap%
\pgfsetmiterjoin%
\definecolor{currentfill}{rgb}{0.094118,0.490196,0.513725}%
\pgfsetfillcolor{currentfill}%
\pgfsetlinewidth{0.000000pt}%
\definecolor{currentstroke}{rgb}{0.000000,0.000000,0.000000}%
\pgfsetstrokecolor{currentstroke}%
\pgfsetstrokeopacity{0.000000}%
\pgfsetdash{}{0pt}%
\pgfpathmoveto{\pgfqpoint{5.086517in}{1.813992in}}%
\pgfpathlineto{\pgfqpoint{5.086517in}{1.813992in}}%
\pgfpathlineto{\pgfqpoint{5.086517in}{1.727786in}}%
\pgfpathlineto{\pgfqpoint{5.086517in}{1.727786in}}%
\pgfpathlineto{\pgfqpoint{5.086517in}{1.813992in}}%
\pgfpathclose%
\pgfusepath{fill}%
\end{pgfscope}%
\begin{pgfscope}%
\pgfpathrectangle{\pgfqpoint{5.086517in}{0.722000in}}{\pgfqpoint{2.067483in}{2.508000in}}%
\pgfusepath{clip}%
\pgfsetbuttcap%
\pgfsetmiterjoin%
\definecolor{currentfill}{rgb}{0.094118,0.490196,0.513725}%
\pgfsetfillcolor{currentfill}%
\pgfsetlinewidth{0.000000pt}%
\definecolor{currentstroke}{rgb}{0.000000,0.000000,0.000000}%
\pgfsetstrokecolor{currentstroke}%
\pgfsetstrokeopacity{0.000000}%
\pgfsetdash{}{0pt}%
\pgfpathmoveto{\pgfqpoint{5.086517in}{1.516730in}}%
\pgfpathlineto{\pgfqpoint{6.846672in}{1.516730in}}%
\pgfpathlineto{\pgfqpoint{6.846672in}{1.430524in}}%
\pgfpathlineto{\pgfqpoint{5.086517in}{1.430524in}}%
\pgfpathlineto{\pgfqpoint{5.086517in}{1.516730in}}%
\pgfpathclose%
\pgfusepath{fill}%
\end{pgfscope}%
\begin{pgfscope}%
\pgfpathrectangle{\pgfqpoint{5.086517in}{0.722000in}}{\pgfqpoint{2.067483in}{2.508000in}}%
\pgfusepath{clip}%
\pgfsetbuttcap%
\pgfsetmiterjoin%
\definecolor{currentfill}{rgb}{0.094118,0.490196,0.513725}%
\pgfsetfillcolor{currentfill}%
\pgfsetlinewidth{0.000000pt}%
\definecolor{currentstroke}{rgb}{0.000000,0.000000,0.000000}%
\pgfsetstrokecolor{currentstroke}%
\pgfsetstrokeopacity{0.000000}%
\pgfsetdash{}{0pt}%
\pgfpathmoveto{\pgfqpoint{5.086517in}{1.219468in}}%
\pgfpathlineto{\pgfqpoint{6.818733in}{1.219468in}}%
\pgfpathlineto{\pgfqpoint{6.818733in}{1.133262in}}%
\pgfpathlineto{\pgfqpoint{5.086517in}{1.133262in}}%
\pgfpathlineto{\pgfqpoint{5.086517in}{1.219468in}}%
\pgfpathclose%
\pgfusepath{fill}%
\end{pgfscope}%
\begin{pgfscope}%
\pgfpathrectangle{\pgfqpoint{5.086517in}{0.722000in}}{\pgfqpoint{2.067483in}{2.508000in}}%
\pgfusepath{clip}%
\pgfsetbuttcap%
\pgfsetmiterjoin%
\definecolor{currentfill}{rgb}{0.094118,0.490196,0.513725}%
\pgfsetfillcolor{currentfill}%
\pgfsetlinewidth{0.000000pt}%
\definecolor{currentstroke}{rgb}{0.000000,0.000000,0.000000}%
\pgfsetstrokecolor{currentstroke}%
\pgfsetstrokeopacity{0.000000}%
\pgfsetdash{}{0pt}%
\pgfpathmoveto{\pgfqpoint{5.086517in}{0.922206in}}%
\pgfpathlineto{\pgfqpoint{6.166824in}{0.922206in}}%
\pgfpathlineto{\pgfqpoint{6.166824in}{0.836000in}}%
\pgfpathlineto{\pgfqpoint{5.086517in}{0.836000in}}%
\pgfpathlineto{\pgfqpoint{5.086517in}{0.922206in}}%
\pgfpathclose%
\pgfusepath{fill}%
\end{pgfscope}%
\begin{pgfscope}%
\pgfsetrectcap%
\pgfsetmiterjoin%
\pgfsetlinewidth{0.803000pt}%
\definecolor{currentstroke}{rgb}{0.713725,0.741176,0.768627}%
\pgfsetstrokecolor{currentstroke}%
\pgfsetdash{}{0pt}%
\pgfpathmoveto{\pgfqpoint{5.086517in}{0.722000in}}%
\pgfpathlineto{\pgfqpoint{7.154000in}{0.722000in}}%
\pgfusepath{stroke}%
\end{pgfscope}%
\begin{pgfscope}%
\definecolor{textcolor}{rgb}{0.000000,0.000000,0.000000}%
\pgfsetstrokecolor{textcolor}%
\pgfsetfillcolor{textcolor}%
\pgftext[x=6.112995in,y=3.072897in,left,]{\color{textcolor}{\fontfamily{\familydefault}\fontsize{8.200000}{9.840000}\selectfont\catcode`\^=\active\def^{\ifmmode\sp\else\^{}\fi}\catcode`\%=\active\def
\end{pgfscope}%
\begin{pgfscope}%
\definecolor{textcolor}{rgb}{0.000000,0.000000,0.000000}%
\pgfsetstrokecolor{textcolor}%
\pgfsetfillcolor{textcolor}%
\pgftext[x=6.405236in,y=2.775635in,left,]{\color{textcolor}{\fontfamily{\familydefault}\fontsize{8.200000}{9.840000}\selectfont\catcode`\^=\active\def^{\ifmmode\sp\else\^{}\fi}\catcode`\%=\active\def
\end{pgfscope}%
\begin{pgfscope}%
\definecolor{textcolor}{rgb}{0.000000,0.000000,0.000000}%
\pgfsetstrokecolor{textcolor}%
\pgfsetfillcolor{textcolor}%
\pgftext[x=6.600809in,y=2.478373in,left,]{\color{textcolor}{\fontfamily{\familydefault}\fontsize{8.200000}{9.840000}\selectfont\catcode`\^=\active\def^{\ifmmode\sp\else\^{}\fi}\catcode`\%=\active\def
\end{pgfscope}%
\begin{pgfscope}%
\definecolor{textcolor}{rgb}{0.000000,0.000000,0.000000}%
\pgfsetstrokecolor{textcolor}%
\pgfsetfillcolor{textcolor}%
\pgftext[x=6.507679in,y=2.181111in,left,]{\color{textcolor}{\fontfamily{\familydefault}\fontsize{8.200000}{9.840000}\selectfont\catcode`\^=\active\def^{\ifmmode\sp\else\^{}\fi}\catcode`\%=\active\def
\end{pgfscope}%
\begin{pgfscope}%
\definecolor{textcolor}{rgb}{0.000000,0.000000,0.000000}%
\pgfsetstrokecolor{textcolor}%
\pgfsetfillcolor{textcolor}%
\pgftext[x=6.507679in,y=1.883849in,left,]{\color{textcolor}{\fontfamily{\familydefault}\fontsize{8.200000}{9.840000}\selectfont\catcode`\^=\active\def^{\ifmmode\sp\else\^{}\fi}\catcode`\%=\active\def
\end{pgfscope}%
\begin{pgfscope}%
\definecolor{textcolor}{rgb}{0.000000,0.000000,0.000000}%
\pgfsetstrokecolor{textcolor}%
\pgfsetfillcolor{textcolor}%
\pgftext[x=6.727465in,y=1.586587in,left,]{\color{textcolor}{\fontfamily{\familydefault}\fontsize{8.200000}{9.840000}\selectfont\catcode`\^=\active\def^{\ifmmode\sp\else\^{}\fi}\catcode`\%=\active\def
\end{pgfscope}%
\begin{pgfscope}%
\definecolor{textcolor}{rgb}{0.000000,0.000000,0.000000}%
\pgfsetstrokecolor{textcolor}%
\pgfsetfillcolor{textcolor}%
\pgftext[x=6.619435in,y=1.289325in,left,]{\color{textcolor}{\fontfamily{\familydefault}\fontsize{8.200000}{9.840000}\selectfont\catcode`\^=\active\def^{\ifmmode\sp\else\^{}\fi}\catcode`\%=\active\def
\end{pgfscope}%
\begin{pgfscope}%
\definecolor{textcolor}{rgb}{0.000000,0.000000,0.000000}%
\pgfsetstrokecolor{textcolor}%
\pgfsetfillcolor{textcolor}%
\pgftext[x=5.744014in,y=0.992063in,left,]{\color{textcolor}{\fontfamily{\familydefault}\fontsize{8.200000}{9.840000}\selectfont\catcode`\^=\active\def^{\ifmmode\sp\else\^{}\fi}\catcode`\%=\active\def
\end{pgfscope}%
\begin{pgfscope}%
\definecolor{textcolor}{rgb}{0.000000,0.000000,0.000000}%
\pgfsetstrokecolor{textcolor}%
\pgfsetfillcolor{textcolor}%
\pgftext[x=6.063636in,y=2.959937in,left,]{\color{textcolor}{\fontfamily{\familydefault}\fontsize{8.200000}{9.840000}\selectfont\catcode`\^=\active\def^{\ifmmode\sp\else\^{}\fi}\catcode`\%=\active\def
\end{pgfscope}%
\begin{pgfscope}%
\definecolor{textcolor}{rgb}{0.000000,0.000000,0.000000}%
\pgfsetstrokecolor{textcolor}%
\pgfsetfillcolor{textcolor}%
\pgftext[x=5.120044in,y=2.662675in,left,]{\color{textcolor}{\fontfamily{\familydefault}\fontsize{8.200000}{9.840000}\selectfont\catcode`\^=\active\def^{\ifmmode\sp\else\^{}\fi}\catcode`\%=\active\def
\end{pgfscope}%
\begin{pgfscope}%
\definecolor{textcolor}{rgb}{0.000000,0.000000,0.000000}%
\pgfsetstrokecolor{textcolor}%
\pgfsetfillcolor{textcolor}%
\pgftext[x=5.120044in,y=2.365413in,left,]{\color{textcolor}{\fontfamily{\familydefault}\fontsize{8.200000}{9.840000}\selectfont\catcode`\^=\active\def^{\ifmmode\sp\else\^{}\fi}\catcode`\%=\active\def
\end{pgfscope}%
\begin{pgfscope}%
\definecolor{textcolor}{rgb}{0.000000,0.000000,0.000000}%
\pgfsetstrokecolor{textcolor}%
\pgfsetfillcolor{textcolor}%
\pgftext[x=5.110731in,y=2.068151in,left,]{\color{textcolor}{\fontfamily{\familydefault}\fontsize{8.200000}{9.840000}\selectfont\catcode`\^=\active\def^{\ifmmode\sp\else\^{}\fi}\catcode`\%=\active\def
\end{pgfscope}%
\begin{pgfscope}%
\definecolor{textcolor}{rgb}{0.000000,0.000000,0.000000}%
\pgfsetstrokecolor{textcolor}%
\pgfsetfillcolor{textcolor}%
\pgftext[x=5.110731in,y=1.770889in,left,]{\color{textcolor}{\fontfamily{\familydefault}\fontsize{8.200000}{9.840000}\selectfont\catcode`\^=\active\def^{\ifmmode\sp\else\^{}\fi}\catcode`\%=\active\def
\end{pgfscope}%
\begin{pgfscope}%
\definecolor{textcolor}{rgb}{0.000000,0.000000,0.000000}%
\pgfsetstrokecolor{textcolor}%
\pgfsetfillcolor{textcolor}%
\pgftext[x=6.870885in,y=1.473627in,left,]{\color{textcolor}{\fontfamily{\familydefault}\fontsize{8.200000}{9.840000}\selectfont\catcode`\^=\active\def^{\ifmmode\sp\else\^{}\fi}\catcode`\%=\active\def
\end{pgfscope}%
\begin{pgfscope}%
\definecolor{textcolor}{rgb}{0.000000,0.000000,0.000000}%
\pgfsetstrokecolor{textcolor}%
\pgfsetfillcolor{textcolor}%
\pgftext[x=6.842946in,y=1.176365in,left,]{\color{textcolor}{\fontfamily{\familydefault}\fontsize{8.200000}{9.840000}\selectfont\catcode`\^=\active\def^{\ifmmode\sp\else\^{}\fi}\catcode`\%=\active\def
\end{pgfscope}%
\begin{pgfscope}%
\definecolor{textcolor}{rgb}{0.000000,0.000000,0.000000}%
\pgfsetstrokecolor{textcolor}%
\pgfsetfillcolor{textcolor}%
\pgftext[x=6.191037in,y=0.879103in,left,]{\color{textcolor}{\fontfamily{\familydefault}\fontsize{8.200000}{9.840000}\selectfont\catcode`\^=\active\def^{\ifmmode\sp\else\^{}\fi}\catcode`\%=\active\def
\end{pgfscope}%
\begin{pgfscope}%
\definecolor{textcolor}{rgb}{0.000000,0.000000,0.000000}%
\pgfsetstrokecolor{textcolor}%
\pgfsetfillcolor{textcolor}%
\pgftext[x=5.086517in,y=3.438333in,left,base]{\color{textcolor}{\fontfamily{\familydefault}\fontsize{11.500000}{13.800000}\selectfont\catcode`\^=\active\def^{\ifmmode\sp\else\^{}\fi}\catcode`\%=\active\def
\end{pgfscope}%
\begin{pgfscope}%
\pgfsetbuttcap%
\pgfsetmiterjoin%
\definecolor{currentfill}{rgb}{0.384314,0.423529,0.478431}%
\pgfsetfillcolor{currentfill}%
\pgfsetlinewidth{0.000000pt}%
\definecolor{currentstroke}{rgb}{0.000000,0.000000,0.000000}%
\pgfsetstrokecolor{currentstroke}%
\pgfsetstrokeopacity{0.000000}%
\pgfsetdash{}{0pt}%
\pgfpathmoveto{\pgfqpoint{5.122568in}{3.219529in}}%
\pgfpathlineto{\pgfqpoint{5.386457in}{3.219529in}}%
\pgfpathlineto{\pgfqpoint{5.386457in}{3.311890in}}%
\pgfpathlineto{\pgfqpoint{5.122568in}{3.311890in}}%
\pgfpathlineto{\pgfqpoint{5.122568in}{3.219529in}}%
\pgfpathclose%
\pgfusepath{fill}%
\end{pgfscope}%
\begin{pgfscope}%
\definecolor{textcolor}{rgb}{0.000000,0.000000,0.000000}%
\pgfsetstrokecolor{textcolor}%
\pgfsetfillcolor{textcolor}%
\pgftext[x=5.492013in,y=3.219529in,left,base]{\color{textcolor}{\fontfamily{\familydefault}\fontsize{9.500000}{11.400000}\selectfont\catcode`\^=\active\def^{\ifmmode\sp\else\^{}\fi}\catcode`\%=\active\def
\end{pgfscope}%
\begin{pgfscope}%
\pgfsetbuttcap%
\pgfsetmiterjoin%
\definecolor{currentfill}{rgb}{0.094118,0.490196,0.513725}%
\pgfsetfillcolor{currentfill}%
\pgfsetlinewidth{0.000000pt}%
\definecolor{currentstroke}{rgb}{0.000000,0.000000,0.000000}%
\pgfsetstrokecolor{currentstroke}%
\pgfsetstrokeopacity{0.000000}%
\pgfsetdash{}{0pt}%
\pgfpathmoveto{\pgfqpoint{6.277463in}{3.219529in}}%
\pgfpathlineto{\pgfqpoint{6.541352in}{3.219529in}}%
\pgfpathlineto{\pgfqpoint{6.541352in}{3.311890in}}%
\pgfpathlineto{\pgfqpoint{6.277463in}{3.311890in}}%
\pgfpathlineto{\pgfqpoint{6.277463in}{3.219529in}}%
\pgfpathclose%
\pgfusepath{fill}%
\end{pgfscope}%
\begin{pgfscope}%
\definecolor{textcolor}{rgb}{0.000000,0.000000,0.000000}%
\pgfsetstrokecolor{textcolor}%
\pgfsetfillcolor{textcolor}%
\pgftext[x=6.646907in,y=3.219529in,left,base]{\color{textcolor}{\fontfamily{\familydefault}\fontsize{9.500000}{11.400000}\selectfont\catcode`\^=\active\def^{\ifmmode\sp\else\^{}\fi}\catcode`\%=\active\def
\end{pgfscope}%
\end{pgfpicture}%
\makeatother%
\endgroup%

%% file: main_extended.tex
\section{Scope, naming, and what is not claimed}
\label{app:scope}

Both RL branches start from \textsc{SFT-Init}. \textsc{RL-Base} is the released checkpoint, and
\textsc{RL-Indic} is an independently trained, unreleased sibling. The evaluated RL checkpoints
are both from step 40.

This is not a state-of-the-art claim, a factorial reward ablation, or a hardware-normalized
efficiency comparison. The benchmark pattern and bundled intervention support a reward-coverage
account, but do not establish a causal effect of any single reward component.

An earlier comparison used the previous epoch's SFT endpoint rather than the export that initializes
RL. Matching nearly all baseline scores did not detect the mistake. The campaign's recorded model
path did. Table~\ref{tab:baseline-corrections} preserves the resulting corrections, and the current
comparison uses the corrected initialization. The episode illustrates why agreement between result
files cannot replace checkpoint provenance.

\begin{table}[h]
\centering\small
\caption{Changes to reported campaign-A deltas after correcting the SFT baseline.}
\label{tab:baseline-corrections}
\begin{tabular}{@{}lrr@{}}\toprule
Benchmark & Earlier delta & Corrected delta \\\midrule
MMLU & $-0.78$ & $+9.47$ \\
CrossSum-Indic (chrF) & $-10.87$ & $+5.81$ \\
HumanEval+ & $+2.26$ & $-0.12$ \\
AIME 2025 & $+40.00$ & $+1.46$ \\\bottomrule
\end{tabular}
\end{table}

\section{Supervised initialization}
\label{app:sft}

\begin{table}[h]
\centering\small
\caption{Training record for \textsc{SFT-Init}, the export used to initialize both RL branches.}
\label{tab:sft-config}
\begin{tabular}{ll}\toprule
Setting & Value \\\midrule
Export point & partway through the final pass, verified from export metadata \\
Scheduled sequence positions & 80.53B ($2{,}400\times256\times131{,}072$), including padding \\
Approximate loss-bearing positions & $\sim$15B (18.2\%, assistant-only mask) \\
Trainable-rate measurement & 95,515 / 524,288 positions over four packed rows \\
Intra-pack packing efficiency & 99.98\% \\
Prepared packed capacity & 92.12B positions across 702,824 sequences \\
Packed rows / scheduled documents & 702,824 / 15,160,245 \\
Configured iters / endpoint / RL export & 2,752 / 2,665 / 2,400 \\
Epoch traversal & sequential bucket blocks, no repeat or wrap-around \\
Sequence length / precision & 131,072 / BF16 \\
GPUs / parallelism & 256 H200, TP=1, PP=1, CP=4, EP=8, DP=64 \\
Global / micro batch & 256 / 1 \\
Learning-rate schedule & cosine, $5\times10^{-7}$ peak to $1\times10^{-7}$ minimum \\
Optimizer & Adam, $\beta_2=0.95$, $\epsilon=10^{-8}$, weight decay 0.1, clip 1.0 \\
Data organization & 15 broadly homogeneous curriculum buckets \\
Materialized rows / language balance & 15,214,121, with 64.0\% English and 36.0\% Indic \\
Loss at RL-init export & 0.613583 (per-step), median 0.606070 over iters 2,350--2,450 \\
Retained SFT curriculum compute & $\sim$17,205 GPU-hours (entire curriculum, not only to iter 2,400) \\
\bottomrule
\end{tabular}
\end{table}

\paragraph{Iteration accounting.} Each curriculum bucket ends on its own step boundary, so the
scheduled iteration count is $\sum_b \lceil N_b/256 \rceil$, not a single ceiling over all packed
sequences. This gives 2,752 iterations rather than 2,746. Final partial batches are padded with
EOS sequences whose loss mask is zero. No examples are repeated or dropped at these boundaries.

Boundary padding occupies about 0.2\% of scheduled batch slots. It does not change the reported
18.2\% loss-bearing-position rate at the shown precision. That rate was measured before runtime
padding on a small sample of packed rows, so it is illustrative rather than a corpus-wide constant
(Table~\ref{tab:sft-config}).

\paragraph{Corpus.} The mixture spans English and eight Indic languages (Bengali, Gujarati, Hindi,
Kannada, Malayalam, Marathi, Tamil, and Telugu) in native and romanized script, across question
answering, reasoning, mathematics, summarization, instruction following, tool interaction, and
multi-turn dialogue. Hindi has the largest Indic row share at 11.1\%. Tamil, Telugu, Kannada, and
Malayalam each contribute 4.8\%. These are row shares, not token or loss weights.

Translation preserves structure. User-facing prose is translated, while code, mathematical spans,
tool schemas, function names, and machine-readable arguments are reinserted byte-identically.
Incomplete translations are quarantined rather than replaced with English. Loss applies only to
the assistant spans selected for training. System and user history remain visible but masked, and
tool actions and observations stay together as coherent examples.

\paragraph{Pre-RL conventions.} In-script output rates range from 93.5\% to 99.1\% in the multilingual
harness. A later matched IndicIFEval-ground evaluation records still higher retention
(Figure~\ref{fig:script-diagnostics}B). These measurements establish the output conventions present
before reinforcement learning.

\section{Complete matched evaluation}
\label{app:full}

\input{full_table}

\section{Reinforcement-learning configurations}
\label{app:rl}

\begin{table}[h]
\centering\small
\caption{Released \textsc{RL-Base} configuration. No custom reward shaping is applied. Each
environment returns its own binary or task-specific scalar.}
\label{tab:rl-base-config}
\begin{tabular}{@{}lp{8.85cm}@{}}\toprule
Setting & Value \\\midrule
Initialization & \textsc{SFT-Init} \\
Algorithm & GRPO with a leave-one-out baseline and per-group Bessel-corrected standard-deviation normalization, divisor $\sigma+10^{-6}$, \emph{no floor} \\
Resolved estimator flags & \texttt{use\_leave\_one\_out\_baseline=true}, \texttt{normalize\_rewards=true} \\
Generations per prompt & 16 \\
Dynamic sampling & enabled, zero-variance groups resampled \\
Sampling temperature & 1.0 \\
Environments & judged mathematics, structured outputs, workplace assistant, and MCQA \\
Language / length / repetition reward & none \\
Inference placement & colocated generation and training workers \\
Recorded compute & 763 GPU-hours \\
Validation reward & 0.508 at step 10 $\rightarrow$ 0.719 at step 40 (not accuracies) \\
\bottomrule
\end{tabular}
\end{table}

\begin{table}[h]
\centering\small
\caption{\textsc{RL-Indic} configuration and input mixture after difficulty filtering and
per-environment caps. Rounded shares sum to 100.1\%, while row counts total 15,300.
These are surviving row shares, not target weights.}
\label{tab:rl-indic-config}
\begin{tabular}{@{}lp{4.35cm}@{\ }lrr@{}}\toprule
Setting & Value & Pool & Rows & Share \\\midrule
Added environment & \texttt{indic\_if} & Mathematics & 6,192 & 40.5\% \\
Reward & $R_{\text{task}}\!+\!0.3R_{\text{lang}}\!-\!0.0R_{\text{len}}\!-\!0.3R_{\text{rep}}$ & Indic instruction following & 3,179 & 20.8\% \\
Malformed / abandoned & $-0.25$ & Structured outputs & 1,932 & 12.6\% \\
Advantage std.\ floor & $\max(\sigma,0.05)$, with 44\% of \texttt{indic\_if} groups below it & Workplace assistant & 1,804 & 11.8\% \\
Steps / rows / validation & 40 / 15,300 / 256 & English guardrail & 1,509 & 9.9\% \\
Recorded rollouts & 251,906 & MCQA & 684 & 4.5\% \\
Hardware / wall time & 16$\times$8 H200 / 5h34m & \textbf{Total} & \textbf{15,300} & \textbf{100\%} \\
Max generated tokens & 16,384 & & & \\
Max RL sequence length & 49,152 (rollout cap only) & & & \\
Checkpoint cadence & every 5 steps & & & \\
Validation reward & 0.320 $\rightarrow$ 0.649 at step 40 & & & \\
\bottomrule
\end{tabular}
\end{table}

\paragraph{Reward scope.} Section~\ref{sec:indic-reward} gives the reward construction and the
scope of each term. The configuration here retains the zero length coefficient explicitly to
distinguish an inactive term from an unreported one.

\paragraph{Context limit.} The 49,152-token rollout cap does not change the trained or exported
context capacity of 131,072 tokens. It limits rollout collection, not the model's context window.

\paragraph{Checkpoint selection.} The rule preferred step 30 unless step 40 retained MATH.
The later candidate did so and also improved GSM8K, IFEval, and both IndicIFEval suites.
Table~\ref{tab:rl-indic-config} records the selected checkpoint and cadence.

The mixture table describes input rows, not gradient shares. Dynamic sampling removes constant-reward
groups, so the realized distribution can differ as described in \S\ref{sec:setup}.

\subsection{Advantage estimation}
\label{app:advantage}

Both branches center rewards with a leave-one-out baseline. For a group of size $G$,
\begin{equation}
r_i-\mu_{-i}=\frac{G}{G-1}(r_i-\bar r).
\end{equation}
The centering factors are $16/15$ for \textsc{RL-Base} and $8/7$ for \textsc{RL-Indic}.
Both use the group's Bessel-corrected standard deviation in Equation~\ref{eq:advantages}.
Since the divisor is common within a group, it preserves reward ordering and the sign of centered
rewards. The floor changes low-spread advantage magnitudes rather than their ordering.

\begin{table}[h]
\centering\small
\caption{Language-directive diagnostic. Explicit Hindi directives improve in-script output over
20 training steps, while implicit and referential phrasings remain near zero.}
\label{tab:language-cues}
\begin{tabular}{@{}lrr@{}}\toprule
Directive & Initial in-script rate & Rate after 20 steps \\\midrule
Explicit Hindi name & 38.4\% & 90.5\% \\
Implicit or referential wording & \multicolumn{2}{c}{near zero} \\\bottomrule
\end{tabular}
\end{table}

\input{diff_spec}

\section{The generative judge, and the distillation round}
\label{app:judge}
This appendix records the construction of the round-two reward, the round-three objective, and the benchmark profile the two later rounds produce.

\paragraph{Preference data.} The judge is trained on 262{,}834 judging examples, each pair presented in both presentation orders, of which 72.4\% come from our Indic preference data and 27.6\% from HelpSteer3. By domain the mixture is 61.2\% general, 20.3\% code, 13.9\% STEM, and 4.6\% multilingual. The base model is a 26B-parameter mixture of experts with 4B active parameters and a frozen vision tower. Training uses 128 prompts per step with eight sampled verdicts each, in a 16{,}384-token context.

\paragraph{Output contract and objective.} For a conversation and two candidate responses the judge emits an absolute score $\hat h_k$ for each response and a relative rank $\hat r$ on a six-point scale. Writing $d_{h_k}=\hat h_k-h_k$ and $d_r=\hat r-r$ against the ground truth, and $H_\delta$ for the Huber loss, the per-sample terms are
\begin{equation}
R_{\mathrm{acc}}=4.0\,\mathbf{1}\!\left[\operatorname{sgn}(\hat r-3.5)=\operatorname{sgn}(r-3.5)\right],
\qquad
R_{\mathrm{cal}}=-1.2H_{1.5}(d_r)-0.6\left(H_1(d_{h_1})+H_1(d_{h_2})\right),
\end{equation}
together with a consistency penalty of $-0.5$ when the sign of $\hat h_1-\hat h_2$ contradicts $\hat r$. Their sum is clipped to $[-6,4]$ and an unparseable verdict scores $-8$, which places it strictly below the clip floor, so every parseable verdict dominates every unparseable one. A batch-level term adds up to $3$ in proportion to the agreement between the predicted and true rank histograms, measured by total variation distance; without it the predicted ranking collapses onto a few modes. Advantages are group-centred but not standardized, because most groups agree on the preferred side and dividing by the group standard deviation would rescale pure calibration noise.

\paragraph{Validation.} Pairwise accuracy saturates by step 20 while calibration continues to move, so the exported checkpoint is chosen on total variation distance rather than on accuracy.

\begin{table}[h]
\centering\small
\caption{Judge validation. Step 100 is the exported checkpoint; later steps lose on both axes.}
\label{tab:judge-val}
\begin{tabular}{@{}lrrrrrr@{}}\toprule
Step & 0 & 20 & 60 & 100 & 140 & 160 \\\midrule
Pairwise accuracy & 0.661 & 0.810 & 0.813 & 0.820 & 0.831 & 0.812 \\
Format validity & 0.996 & 0.999 & 0.998 & 1.000 & -- & -- \\
Total variation distance $\downarrow$ & 0.229 & 0.305 & 0.108 & 0.083 & 0.173 & 0.156 \\\bottomrule
\end{tabular}
\end{table}

\paragraph{Judge benchmarks.} Under greedy decoding with both presentation orders averaged and unparseable verdicts scored as wrong, the judge reaches 88.3 on the JudgeBench GPT-4o subset and 90.8 overall on RM-Bench, against 87.4 and 87.3 for a Qwen3-Nemotron-235B-A22B reference judge. The RM-Bench margin concentrates in the hard, style-adversarial slice, 86.4 against 77.4. Position bias is 0.042 over 11{,}736 comparisons. Because a pairwise judge sees both responses, these figures are not comparable to those of scalar reward models.

\paragraph{Round two.} The policy is optimized against the frozen judge with the GRPO objective of Section~\ref{sec:later-rounds} and a KL penalty toward the initializing policy. Rather than score all pairs in a group, the judge is called on a circular ring in which rollout $i$ is compared with rollout $i+1$. Each rollout is judged twice, once in each position, and the base reward is the mean of its two scores. Ties in helpfulness are broken by the relative rank, which is never exactly neutral, so no comparison leaves a pair undifferentiated.

\paragraph{Length and auxiliary rewards.} Both configurations use group-relative, zero-sum length terms for reasoning and answer length. The shortest answer receives a bonus when its base reward clears the group threshold. Rollouts with no answer score $-2$, and truncations are penalized in proportion to how far they run past the cap. In the \texttt{+} configuration, a think-length penalty applies to a pseudo-random fifth of groups whose reasoning exceeds a 5,000-token budget. This configuration also adds a repetition penalty and a script-consistency bonus of $+0.25$ when the answer's script matches the prompt's and $-0.15$ on a confident mismatch. An earlier ramp-style reasoning penalty removed 60\% of validation reasoning tokens.

\paragraph{Prompts.} Prompts are the input side of the judge's own training examples, deduplicated and rebalanced to 40\% general, 25\% code, 25\% STEM, and 10\% multilingual, giving 46{,}920 prompts in 23 languages.

\paragraph{Round three.} \textsc{RL-OPD} minimizes an equally weighted sum of the forward and reverse divergences between teacher and student next-token distributions, restricted to the teacher's top-$64$ set and not renormalized over that support, so the objective is not a proper divergence. The forward term is mass-covering and penalizes the student for abandoning mass the teacher still places, which is the behavior the targeted regressions destroy; the reverse term is mode-seeking. No reward model, verifier, or frozen reference policy appears in the objective. Training runs for 100 steps at a constant learning rate of $10^{-5}$ with 128 student rollouts per step, and we evaluate at step 20.

\paragraph{Round-two profile.} Table~\ref{tab:round2} gives the measured benchmarks for both round-two configurations started from \textsc{RL-Indic}. Grounded Indic instruction following improves while grounded Indic extraction is destroyed, in the same runs and the same language.

\begin{table}[h]
\centering\small
\caption{Round two from the \textsc{RL-Indic} branch. These campaigns scored a six-benchmark subset; IndicQA is absent from all of them, and six of the eight branch-by-step combinations were never scored. Exact match here is the more permissive post-August grader, so it is not comparable with the \textsc{RL-Indic} column.}
\label{tab:round2}
\begin{tabular}{@{}lrrrr@{}}\toprule
 & \textsc{RL-Indic} & \textsc{Judge-I} 20 & \textsc{Judge-I+} 20 & \textsc{Judge-I+} 50 \\\midrule
IndicIFEval-ground, strict & 60.00 & 62.11 & 63.96 & 65.06 \\
IndicIFEval-translated, strict & 63.27 & 59.31 & 59.14 & 58.91 \\
IFBench, strict & 50.33 & 44.00 & 45.33 & 40.67 \\
AIME 2025, pass@1 & 65.42 & 67.29 & 67.08 & 66.04 \\
XQuAD-Indic, F1 & 50.44 & 19.34 & 17.67 & 19.08 \\
XQuAD-Indic, EM & 32.00 & 3.79 & 1.91 & 2.85 \\\bottomrule
\end{tabular}
\end{table}

\paragraph{A note on exact match.} The exact-match implementation for the extractive question-answering benchmarks changed in late August 2026, from requiring the gold string to requiring it as a contiguous run inside a longer answer. Each campaign froze whichever version was live when it launched, so the \textsc{SFT-Init}, round-two, and round-three campaigns use the permissive rule while \textsc{RL-Base} and \textsc{RL-Indic} use the strict one. Table~\ref{tab:full} reports these four benchmarks rescored strictly on both columns. Elsewhere we read comparisons that cross the boundary on F1, which the change did not affect, and where we quote exact match across it the permissive grader sits on the side that would flatter the later checkpoint, so the stated losses are conservative.

\section{Head-to-head and tool-use decomposition}
\label{app:head}

Table~\ref{tab:threeway} gives the complete benchmark comparison. The pooled means are nearly equal,
even though individual benchmark differences reach roughly $+33$ and $-14$ points. \textsc{RL-Indic}
is stronger on instruction following, translation, and several Indic generation and QA tasks.
\textsc{RL-Base} remains stronger on MMLU-Pro, LiveCodeBench, HMMT, AIME, GPQA-Diamond, MMLU,
and MBPP+.

Table~\ref{tab:aggregate} shows how the aggregate depends on metric choice. Using QA F1 instead of
exact match favors the sibling slightly. Removing the targeted instruction-following rows favors
the released branch. Regrading XQuAD-Indic under a common extraction rule leaves its exact-match
component unchanged. The similar full-set means therefore conceal a trade-off rather than a
uniformly similar outcome.

\begin{table}[h]
\centering\small
\caption{Sensitivity of the pooled comparison. Differences are \textsc{RL-Indic} minus
\textsc{RL-Base} on the 0--1 scale. Unreported branch means are shown as dashes.}
\label{tab:aggregate}
\begin{tabular}{@{}lrrr@{}}\toprule
Aggregation & \textsc{RL-Base} & \textsc{RL-Indic} & Difference \\\midrule
Shared metrics, QA exact match (33) & 0.661188 & 0.661233 & $+0.000045$ \\
QA rows scored by F1 instead & --- & --- & $+0.009$ \\
Non-instruction-following subset (27) & --- & --- & $-0.026$ \\\bottomrule
\end{tabular}
\end{table}

\paragraph{Measured and reconstructed deltas.} Each checkpoint was scored in its own campaign.
Comparison~A pairs two of them to give $\textsc{RL-Base}-\textsc{SFT-Init}$, and comparison~B gives
$\textsc{RL-Indic}-\textsc{RL-Base}$; only $\textsc{RL-Indic}-\textsc{SFT-Init}$ is reconstructed, by
chaining the two. The \textsc{RL-Base} column is one scorecard read in both comparisons, so its
agreement across them is an identity and lends no independent support to that chaining, which
missing harness versions and sample identifiers leave uncertain. Prompt templates, reasoning modes,
and extraction rules were recorded per campaign. Without the comparability assumption, the
comparison-B finding still holds: the sibling improves every regressing Indic benchmark relative to
the released branch.

\input{three_way_full}

Figure~\ref{fig:tool-policy} combines the agentic comparison with the BFCL decomposition.
The BFCL headline is renormalized over 17 categories. Its parallel and multi-turn movements
are more informative than the aggregate decline.

The sibling's tool-call syntax parses correctly, but it usually emits only one call per message.
The second-turn diagnostic suggests that at least some parallel failures reflect turn-taking
rather than an inability to produce the missing call. The operational cost is an extra round trip
on parallelizable requests. Tasks requiring independent calls in a single response would provide
a useful test of this interpretation.

\section{Script-compliance measurements across campaigns}
\label{app:script}

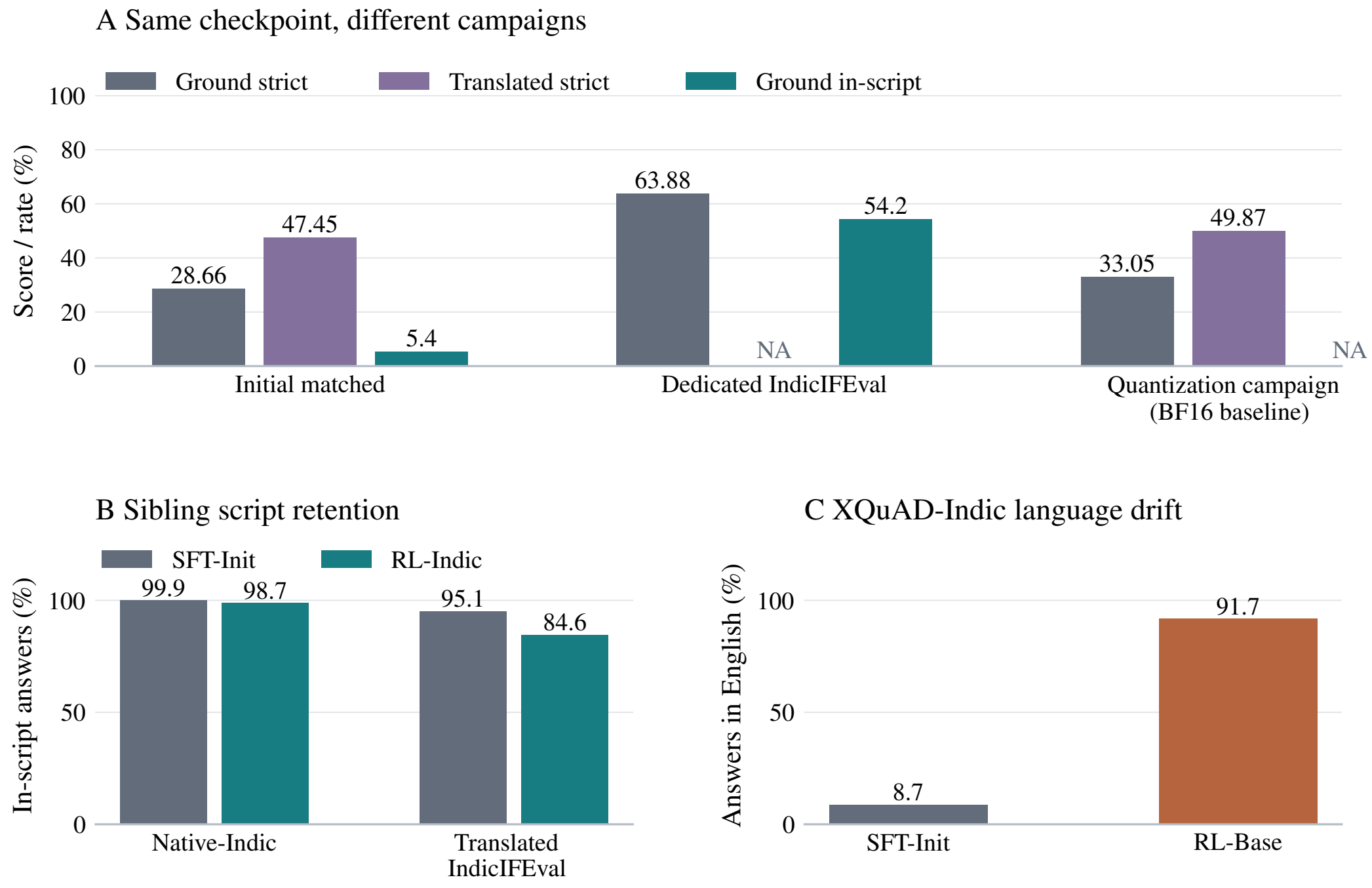
\begin{figure}[tbp]
\centering\small
\resizebox{\linewidth}{!}{\input{figures/script_diagnostics.tex}}
\caption{Task scores and answer language are distinct measurements. (A) IndicIFEval measurements
of the same \textsc{RL-Base} checkpoint across campaigns. NA denotes an unretained metric, not zero.
The dedicated campaign also records a loose score of 65.62. (B) In-script output before and after
sibling training. (C) English answers on XQuAD-Indic before and after the released run.
A prose-tolerant regrade changes macro F1 by at most 0.24 points.}
\label{fig:script-diagnostics}
\end{figure}

Figure~\ref{fig:script-diagnostics} combines the cross-campaign measurements, sibling script
retention, and XQuAD output-language diagnostic. The cross-campaign spread is a sensitivity
check, not an uncertainty interval. Unretained measurements are marked as unavailable.

Translated IndicIFEval prompt-strict accuracy improves by 25.76 points against \textsc{SFT-Init}
under the reconstructed comparison. Its directly measured gain against \textsc{RL-Base} is 13.40.
These task gains coexist with the script losses in Figure~\ref{fig:script-diagnostics}B.

%% file: full_table.tex
{\footnotesize\setlength{\tabcolsep}{4pt}
\begin{longtable}{llrrr}
\caption{Matched evaluation of the final SFT checkpoint (\textsc{SFT-Init}) and the
released \textsc{RL-Base} checkpoint. Both columns come from a single campaign on one
harness, model mode, and decoding configuration. All 55 recorded metrics completed, displayed as 53 rows because the
two IndicIFEval-ground rows each combine a prompt-level and an instruction-level score. Every row was
scored within this campaign. Deltas are
RL minus SFT. Accuracy-family metrics are percentages. chrF++, chrF, and BLEU are native
sacreBLEU scores on the same 0--100 range and are not percentages. Rows marked $\downarrow$ are
better when lower, so their delta sign is inverted relative to the rest of the table.}\label{tab:full}\\
\toprule
Benchmark & Metric & Final SFT & RL ckpt & $\Delta$ (RL $-$ SFT) \\
          &        & ckpt      & step 40 & \\
\midrule
\endfirsthead
\toprule Benchmark & Metric & Final SFT & RL ckpt & $\Delta$ (RL $-$ SFT) \\\midrule
\endhead
\addlinespace[2pt]
\multicolumn{5}{l}{\textit{Math \& Reasoning}} \\
MATH & exact match & 48.69 & 84.76 & $+36.07$ \\
MATH-500 & exact match & 48.80 & 84.00 & $+35.20$ \\
GSM8K & exact match & 85.75 & 94.31 & $+8.56$ \\
MGSM-English & exact match & 91.60 & 98.40 & $+6.80$ \\
MGSM-Indic & exact match & 79.60 & 84.60 & $+5.00$ \\
AIME 2025 & pass@1 & 72.92 & 74.38 & $+1.46$ \\
AIME 2026 & pass@1 & 72.08 & 74.58 & $+2.50$ \\
Beyond AIME & pass@1 & 48.50 & 53.56 & $+5.06$ \\
HMMT February 2025 & pass@1 & 59.58 & 60.83 & $+1.25$ \\
HMMT November 2025 & pass@1 & 58.54 & 63.33 & $+4.79$ \\
GPQA Diamond & pass@1 & 38.19 & 65.03 & $+26.84$ \\
\addlinespace[2pt]
\addlinespace[2pt]
\multicolumn{5}{l}{\textit{Code}} \\
LiveCodeBench & pass@1 & 51.35 & 61.27 & $+9.92$ \\
MBPP+ & pass@1 & 75.24 & 78.78 & $+3.54$ \\
HumanEval+ & pass@1 & 88.66 & 88.54 & $-0.12$ \\
\addlinespace[2pt]
\addlinespace[2pt]
\multicolumn{5}{l}{\textit{Knowledge and MCQ}} \\
MMLU$^{\dagger}$ & exact match & 77.21 & 86.68 & $+9.47$ \\
MMLU-Pro & accuracy & 66.67 & 77.26 & $+10.59$ \\
IndicMMLU-Pro & accuracy & 47.07 & 68.88 & $+21.81$ \\
MILU & accuracy & 73.16 & 78.33 & $+5.17$ \\
ARC-Challenge & exact match & 93.26 & 95.90 & $+2.64$ \\
AGIEval English & exact match & 60.02 & 73.47 & $+13.45$ \\
BhashaBench-Multi & accuracy & 42.06 & 49.22 & $+7.16$ \\
\addlinespace[2pt]
\addlinespace[2pt]
\multicolumn{5}{l}{\textit{Instruction Following}} \\
IFEval & prompt strict & 66.36 & 82.81 & $+16.45$ \\
IFEval & prompt loose & 67.65 & 83.73 & $+16.08$ \\
IFEval & inst.\ strict & 73.26 & 87.17 & $+13.91$ \\
IFEval & inst.\ loose & 74.22 & 87.77 & $+13.55$ \\
IFEval-Hindi & prompt strict & 52.71 & 50.59 & $-2.12$ \\
IFEval-Hindi & prompt loose & 53.42 & 51.30 & $-2.12$ \\
IFEval-Hindi & inst.\ strict & 60.51 & 59.64 & $-0.87$ \\
IFEval-Hindi & inst.\ loose & 61.62 & 60.27 & $-1.35$ \\
IndicIFEval-translated & prompt strict & 37.51 & 49.87 & $+12.36$ \\
IndicIFEval-translated & prompt loose & 38.88 & 50.87 & $+11.99$ \\
IndicIFEval-translated & inst.\ strict & 47.68 & 61.65 & $+13.97$ \\
IndicIFEval-translated & inst.\ loose & 49.04 & 62.43 & $+13.39$ \\
IndicIFEval-ground & prompt/inst.\ strict & 40.52 & 33.05 & $-7.47$ \\
IndicIFEval-ground & prompt/inst.\ loose & 43.59 & 33.43 & $-10.16$ \\
IFBench & prompt strict & 21.67 & 44.33 & $+22.66$ \\
IndicParam & accuracy & 32.94 & 39.28 & $+6.34$ \\
\addlinespace[2pt]
\addlinespace[2pt]
\multicolumn{5}{l}{\textit{QA and Translation}} \\
XQuAD-English & exact match & 46.97 & 63.70 & $+16.73$ \\
XQuAD-English & F1 & 67.15 & 78.86 & $+11.71$ \\
XQuAD-Indic & exact match & 18.84 & 12.86 & $-5.98$ \\
XQuAD-Indic & F1 & 36.51 & 17.19 & $-19.32$ \\
XorQA-Indic & exact match & 6.67 & 35.24 & $+28.57$ \\
XorQA-Indic & F1 & 16.16 & 44.09 & $+27.93$ \\
IndicQA & exact match & 12.77 & 12.22 & $-0.55$ \\
IndicQA & F1 & 26.36 & 17.72 & $-8.64$ \\
FLORES-IN & chrF++ & 14.55 & 36.33 & $+21.78$ \\
FLORES-IN & BLEU & 3.64 & 14.94 & $+11.30$ \\
CrossSum-IN & chrF & 7.00 & 12.81 & $+5.81$ \\
\addlinespace[2pt]
\addlinespace[2pt]
\multicolumn{5}{l}{\textit{Bias diagnostic}} \\
BBQ & accuracy (disamb.) & 94.13 & 97.70 & $+3.57$ \\
BBQ & accuracy (amb.) & 92.29 & 88.69 & $-3.60$ \\
BBQ & bias score (disamb.)$\downarrow$ & 0.99 & 0.60 & $-0.39$ \\
BBQ & bias score (amb.)$\downarrow$ & 4.33 & 7.39 & $+3.06$ \\
\addlinespace[2pt]
\bottomrule
\end{longtable}}

\footnotesize $^{\dagger}$MMLU here is generative exact match and is not comparable to loglikelihood accuracy
reported under other conventions.

%% file: diff_spec.tex
\section{Measuring the English--Indic reward differential}
\label{app:differential}

This diagnostic compares rewards for Latin- and Indic-script answers to the same prompt. It measures
an association within retained mixed-script groups, not the causal effect of script choice.

\subsection{Estimator}

We compare completions within prompt groups to hold task difficulty and wording fixed. Groups with
only one script are excluded. We also report differences by environment, because rewards have
different scales and the environment mix can differ across scripts. Dynamic sampling means the
retained groups need not represent the input distribution.

For each mixed-script group, the statistic is
\begin{equation}
d_g \;=\; \frac{1}{|E_g|}\sum_{i \in E_g} r_i \;-\; \frac{1}{|I_g|}\sum_{i \in I_g} r_i ,
\qquad
\bar{d} \;=\; \frac{1}{|G^{*}|}\sum_{g \in G^{*}} d_g ,
\end{equation}
where $E_g$ and $I_g$ are the Latin- and Indic-script completions in group $g$, and $G^{*}$ contains
groups with both. Script is assigned after the final reasoning block, matching the sibling branch's
answer-scoped language reward. English reasoning followed by an Indic answer is therefore not
classified as a Latin-script completion.

Completions within a pair can still differ in correctness, length, formatting, and content.
Rescoring semantically matched translations or transliterations would better isolate script,
with code and schemas held byte-identical. We have not run that test. The environment decomposition
is $\bar d = \sum_e w_e\bar d_e$, where $w_e$ is environment $e$'s share of retained paired groups.

\subsection{Audit result}

\begin{table}[h]
\centering\small
\caption{Reward differential and environment decomposition in the retained rollout sample.
The final column separately reports script-free answers as a share of Indic-prompt rollouts,
not of paired groups. Dashes indicate unreported values.}
\label{tab:reward-differential}
\begin{tabular}{@{}lrrrr@{}}\toprule
Environment & $w_e$ & $\bar d_e$ & $w_e\bar d_e$ & Script-free share \\\midrule
Structured outputs & 43.4\% & $+0.2588$ & $+0.1123$ & 0.3\% \\
Workplace assistant & 19.3\% & $+0.2483$ & $+0.0480$ & 7.5\% \\
MCQA & --- & $+0.1030$ & --- & --- \\
Mathematics with judge & --- & $+0.0006$ & --- & 10.1\% \\\midrule
\textbf{Aggregate $\bar d$} & \multicolumn{4}{r}{$\mathbf{+0.1681}$ \; (SE 0.0095)} \\
95\% bootstrap CI & \multicolumn{4}{r}{$[+0.1489, +0.1871]$} \\
Paired groups $|G^{*}|$ & \multicolumn{4}{r}{1,148} \\
Unpaired contrast & \multicolumn{4}{r}{$+0.1373$} \\\bottomrule
\end{tabular}
\end{table}

Structured outputs and workplace assistance account for 95.4\% of the aggregate. The bootstrap
interval excludes zero. However, the aggregate and attribution shares depend on the environments'
reward scales, so the per-environment differences are more informative than a scale-free reading
of the pooled number.

The diagnostic uses raw rewards, while GRPO updates on normalized advantages. It therefore does not
quantify update pressure directly. The unpaired contrast is included for completeness, but its
different inclusion rules prevent attributing the paired--unpaired gap to task difficulty alone.

\subsection{Sampling and length-filter sensitivity}

\paragraph{Incomplete rollout coverage.} The dump retains 80,000 records from approximately
125,800 generated completions. Worker files have no step index, so their location in training is
inferred from mean rewards rather than known exactly. Table~\ref{tab:rollout-location} records this
comparison. Most paired groups come from the files whose rewards match early training.

\begin{table}[h]
\centering\small
\caption{Locating retained files by reward level. Four files survive, each capped at 20,000
records, from a run spanning 49 steps. The available reward-level comparisons locate three files.}
\label{tab:rollout-location}
\begin{tabular}{@{}lrr@{}}\toprule
Retained file & Mean reward & Logged reference \\\midrule
Early-level file A & 0.324 & 0.333 (steps 1--8) \\
Early-level file B & 0.312 & 0.333 (steps 1--8) \\
Late-level file & 0.717 & 0.698 (steps 24--49) \\\bottomrule
\end{tabular}
\end{table}

Within an early file, sequential chunks track the logged reward trajectory with correlation
$r=+0.58$. The differential rises across those chunks rather than disappearing
(Table~\ref{tab:mixed-script-coverage}). The late-level file has too few paired groups for a useful
estimate. Fresh checkpoint samples likewise show mixed-script groups becoming scarce and then absent.
Once a sample has no such groups, the differential is undefined, not zero.

\begin{table}[h]
\centering\small
\caption{Within-file differential and fresh-checkpoint mixed-script coverage. The final column
is a count for retained-file rows and a percentage of prompt groups for checkpoint samples.}
\label{tab:mixed-script-coverage}
\begin{tabular}{@{}lrr@{}}\toprule
Sample & $\bar d$ & Mixed-script groups \\\midrule
Early file, first chunk & $+0.062$ & 232 \\
Early file, middle chunk & $+0.150$ & 224 \\
Early file, final chunk & $+0.165$ & 175 \\
Late-level retained file & --- & 2 \\\midrule
Fresh sample, step 20 & --- & 1.7\% \\
Fresh sample, step 30 & --- & 0.9\% \\
Fresh sample, step 40 & undefined & 0\% \\\bottomrule
\end{tabular}
\end{table}

The dump does not identify groups removed by dynamic sampling, which discarded 22\% of completions
run-wide. Selection bias cannot therefore be reconstructed. A second seed gives a differential of
$+0.115$, but shares the truncation problem. Its paired-group count is recorded in
Table~\ref{tab:filter-sensitivity} rather than treated as independent confirmation.

\paragraph{Why answer length matters.} A bare numeral or option letter provides little evidence
about the language of an answer. The primary estimate requires a sufficiently long answer span.
Changing the filter to count only script-bearing characters gives a similar result, while removing
the length filter reduces the aggregate to $+0.066$ (Table~\ref{tab:filter-sensitivity}). The
unfiltered mathematics difference becomes negative, and structured environments no longer dominate.

\begin{table}[h]
\centering\small
\caption{Sensitivity to filtering and seed. The primary filter requires a 20-character answer
span. Counting 20 script-bearing characters is a distinct variant.}
\label{tab:filter-sensitivity}
\begin{tabular}{@{}lrr@{}}\toprule
Variant & $\bar d$ & Paired groups \\\midrule
Primary answer-span filter & $+0.1681$ & 1,148 \\
Script-bearing-character filter & $+0.1745$ & 1,116 \\
No length filter & $+0.066$ & --- \\
Second seed, truncated sample & $+0.115$ & 554 \\\midrule
\multicolumn{2}{@{}l}{Change from dropping versus absorbing ambiguous spans} & 0.0009 \\
\multicolumn{2}{@{}l}{Unfiltered mathematics differential} & $-0.082$ \\\bottomrule
\end{tabular}
\end{table}

Filtering makes script classification more meaningful, but restricts the analysis to a subpopulation
of rewarded answers. Structured outputs has the largest per-environment differential and the smallest
script-free share (Table~\ref{tab:reward-differential}). This makes that environment less exposed
to short-answer ambiguity. It does not remove the selection problem or establish script as the cause
of the reward difference.

%% file: three_way_full.tex
\begin{table}[p]
\centering\footnotesize
\caption{Three-checkpoint comparison across campaigns A and B. The shared \textsc{RL-Base} scores
match Table~\ref{tab:full}. Both deltas use \textsc{SFT-Init} as baseline, so
$\Delta_{\text{Indic}}$ is reconstructed under the campaign-comparability assumption
(\S\ref{sec:setup}). One benchmark-native metric is shown per benchmark.
Table~\ref{tab:full} gives the complete campaign-A metric set.}
\label{tab:threeway}
\setlength{\tabcolsep}{4pt}
\begin{tabular}{@{}lrrrrr@{}}\toprule
Benchmark & SFT & RL-Base & $\Delta_{\text{Base}}$ & RL-Indic & $\Delta_{\text{Indic}}$ \\\midrule
\addlinespace[2pt]\multicolumn{6}{@{}l}{\textit{Math \& Reasoning}} \\
MATH & 48.69 & 84.76 & $+36.07$ & 86.76 & $+38.07$ \\
MATH-500 & 48.80 & 84.00 & $+35.20$ & 87.60 & $+38.80$ \\
GSM8K & 85.75 & 94.31 & $+8.56$ & 91.58 & $+5.83$ \\
MGSM-English & 91.60 & 98.40 & $+6.80$ & 97.60 & $+6.00$ \\
MGSM-Indic & 79.60 & 84.60 & $+5.00$ & 86.40 & $+6.80$ \\
AIME 2025 & 72.92 & 74.38 & $+1.46$ & 65.42 & $-7.50$ \\
AIME 2026 & 72.08 & 74.58 & $+2.50$ & 67.50 & $-4.58$ \\
Beyond AIME & 48.50 & 53.56 & $+5.06$ & 46.37 & $-2.13$ \\
HMMT February 2025 & 59.58 & 60.83 & $+1.25$ & 50.42 & $-9.16$ \\
HMMT November 2025 & 58.54 & 63.33 & $+4.79$ & 58.13 & $-0.41$ \\
GPQA Diamond & 38.19 & 65.03 & $+26.84$ & 60.04 & $+21.85$ \\
\addlinespace[2pt]\multicolumn{6}{@{}l}{\textit{Code}} \\
LiveCodeBench & 51.35 & 61.27 & $+9.92$ & 48.21 & $-3.14$ \\
MBPP+ & 75.24 & 78.78 & $+3.54$ & 74.29 & $-0.95$ \\
HumanEval+ & 88.66 & 88.54 & $-0.12$ & 88.29 & $-0.37$ \\
\addlinespace[2pt]\multicolumn{6}{@{}l}{\textit{Knowledge and MCQ}} \\
MMLU$^{\dagger}$ & 77.21 & 86.68 & $+9.47$ & 81.67 & $+4.46$ \\
MMLU-Pro & 66.67 & 77.26 & $+10.59$ & 63.34 & $-3.33$ \\
IndicMMLU-Pro & 47.07 & 68.88 & $+21.81$ & 61.91 & $+14.84$ \\
MILU & 73.16 & 78.33 & $+5.17$ & 76.90 & $+3.74$ \\
ARC-Challenge & 93.26 & 95.90 & $+2.64$ & 94.97 & $+1.71$ \\
AGIEval English & 60.02 & 73.47 & $+13.45$ & 70.52 & $+10.50$ \\
BhashaBench-Multi & 42.06 & 49.22 & $+7.16$ & 48.33 & $+6.27$ \\
\addlinespace[2pt]\multicolumn{6}{@{}l}{\textit{Instruction following}} \\
IFEval (prompt strict) & 66.36 & 82.81 & $+16.45$ & 86.88 & $+20.52$ \\
IFEval-Hindi (prompt strict) & 52.71 & 50.59 & $-2.12$ & 72.29 & $+19.58$ \\
IndicIFEval-translated (p.s.) & 37.51 & 49.87 & $+12.36$ & 63.27 & $+25.76$ \\
IndicIFEval-ground (p.s.) & 40.52 & 33.05 & $-7.47$ & 60.00 & $+19.48$ \\
IFBench (prompt strict) & 21.67 & 44.33 & $+22.66$ & 50.33 & $+28.66$ \\
IndicParam & 32.94 & 39.28 & $+6.34$ & 38.09 & $+5.15$ \\
\addlinespace[2pt]\multicolumn{6}{@{}l}{\textit{QA and translation}} \\
XQuAD-English (EM) & 46.97 & 63.70 & $+16.73$ & 61.85 & $+14.88$ \\
XQuAD-Indic (EM) & 18.84 & 12.86 & $-5.98$ & 32.00 & $+13.16$ \\
XorQA-Indic (EM) & 6.67 & 35.24 & $+28.57$ & 27.86 & $+21.19$ \\
IndicQA (EM) & 12.77 & 12.22 & $-0.55$ & 23.46 & $+10.69$ \\
FLORES-IN (chrF++) & 14.55 & 36.33 & $+21.78$ & 39.37 & $+24.82$ \\
CrossSum-IN (chrF) & 7.00 & 12.81 & $+5.81$ & 15.06 & $+8.06$ \\
\addlinespace[2pt]\multicolumn{6}{@{}l}{\textit{Bias diagnostic}} \\
BBQ (accuracy, amb.) & 92.29 & 88.69 & $-3.60$ & 88.59 & $-3.70$ \\
\bottomrule
\end{tabular}

\footnotesize $^{\dagger}$Generative exact match.
\end{table}

%% file: figures/script_diagnostics.tex
\begingroup%
\makeatletter%
\begin{pgfpicture}%
\pgfpathrectangle{\pgfpointorigin}{\pgfqpoint{7.300000in}{4.800000in}}%
\pgfusepath{use as bounding box, clip}%
\begin{pgfscope}%
\pgfsetbuttcap%
\pgfsetmiterjoin%
\definecolor{currentfill}{rgb}{1.000000,1.000000,1.000000}%
\pgfsetfillcolor{currentfill}%
\pgfsetlinewidth{0.000000pt}%
\definecolor{currentstroke}{rgb}{1.000000,1.000000,1.000000}%
\pgfsetstrokecolor{currentstroke}%
\pgfsetdash{}{0pt}%
\pgfpathmoveto{\pgfqpoint{0.000000in}{0.000000in}}%
\pgfpathlineto{\pgfqpoint{7.300000in}{0.000000in}}%
\pgfpathlineto{\pgfqpoint{7.300000in}{4.800000in}}%
\pgfpathlineto{\pgfqpoint{0.000000in}{4.800000in}}%
\pgfpathlineto{\pgfqpoint{0.000000in}{0.000000in}}%
\pgfpathclose%
\pgfusepath{fill}%
\end{pgfscope}%
\begin{pgfscope}%
\pgfsetbuttcap%
\pgfsetmiterjoin%
\definecolor{currentfill}{rgb}{1.000000,1.000000,1.000000}%
\pgfsetfillcolor{currentfill}%
\pgfsetlinewidth{0.000000pt}%
\definecolor{currentstroke}{rgb}{0.000000,0.000000,0.000000}%
\pgfsetstrokecolor{currentstroke}%
\pgfsetstrokeopacity{0.000000}%
\pgfsetdash{}{0pt}%
\pgfpathmoveto{\pgfqpoint{0.584000in}{2.894761in}}%
\pgfpathlineto{\pgfqpoint{7.154000in}{2.894761in}}%
\pgfpathlineto{\pgfqpoint{7.154000in}{4.320000in}}%
\pgfpathlineto{\pgfqpoint{0.584000in}{4.320000in}}%
\pgfpathlineto{\pgfqpoint{0.584000in}{2.894761in}}%
\pgfpathclose%
\pgfusepath{fill}%
\end{pgfscope}%
\begin{pgfscope}%
\definecolor{textcolor}{rgb}{0.000000,0.000000,0.000000}%
\pgfsetstrokecolor{textcolor}%
\pgfsetfillcolor{textcolor}%
\pgftext[x=1.714902in,y=2.846150in,,top]{\color{textcolor}{\fontfamily{\familydefault}\fontsize{9.500000}{11.400000}\selectfont\catcode`\^=\active\def^{\ifmmode\sp\else\^{}\fi}\catcode`\%=\active\def
\end{pgfscope}%
\begin{pgfscope}%
\definecolor{textcolor}{rgb}{0.000000,0.000000,0.000000}%
\pgfsetstrokecolor{textcolor}%
\pgfsetfillcolor{textcolor}%
\pgftext[x=4.162741in,y=2.846150in,,top]{\color{textcolor}{\fontfamily{\familydefault}\fontsize{9.500000}{11.400000}\selectfont\catcode`\^=\active\def^{\ifmmode\sp\else\^{}\fi}\catcode`\%=\active\def
\end{pgfscope}%
\begin{pgfscope}%
\definecolor{textcolor}{rgb}{0.000000,0.000000,0.000000}%
\pgfsetstrokecolor{textcolor}%
\pgfsetfillcolor{textcolor}%
\pgftext[x=5.919517in, y=2.755277in, left, base]{\color{textcolor}{\fontfamily{\familydefault}\fontsize{9.500000}{11.400000}\selectfont\catcode`\^=\active\def^{\ifmmode\sp\else\^{}\fi}\catcode`\%=\active\def
\end{pgfscope}%
\begin{pgfscope}%
\definecolor{textcolor}{rgb}{0.000000,0.000000,0.000000}%
\pgfsetstrokecolor{textcolor}%
\pgfsetfillcolor{textcolor}%
\pgftext[x=6.141294in, y=2.610907in, left, base]{\color{textcolor}{\fontfamily{\familydefault}\fontsize{9.500000}{11.400000}\selectfont\catcode`\^=\active\def^{\ifmmode\sp\else\^{}\fi}\catcode`\%=\active\def
\end{pgfscope}%
\begin{pgfscope}%
\pgfpathrectangle{\pgfqpoint{0.584000in}{2.894761in}}{\pgfqpoint{6.570000in}{1.425239in}}%
\pgfusepath{clip}%
\pgfsetrectcap%
\pgfsetroundjoin%
\pgfsetlinewidth{0.501875pt}%
\definecolor{currentstroke}{rgb}{0.886275,0.898039,0.909804}%
\pgfsetstrokecolor{currentstroke}%
\pgfsetdash{}{0pt}%
\pgfpathmoveto{\pgfqpoint{0.584000in}{2.894761in}}%
\pgfpathlineto{\pgfqpoint{7.154000in}{2.894761in}}%
\pgfusepath{stroke}%
\end{pgfscope}%
\begin{pgfscope}%
\definecolor{textcolor}{rgb}{0.000000,0.000000,0.000000}%
\pgfsetstrokecolor{textcolor}%
\pgfsetfillcolor{textcolor}%
\pgftext[x=0.467601in, y=2.849325in, left, base]{\color{textcolor}{\fontfamily{\familydefault}\fontsize{9.500000}{11.400000}\selectfont\catcode`\^=\active\def^{\ifmmode\sp\else\^{}\fi}\catcode`\%=\active\def
\end{pgfscope}%
\begin{pgfscope}%
\pgfpathrectangle{\pgfqpoint{0.584000in}{2.894761in}}{\pgfqpoint{6.570000in}{1.425239in}}%
\pgfusepath{clip}%
\pgfsetrectcap%
\pgfsetroundjoin%
\pgfsetlinewidth{0.501875pt}%
\definecolor{currentstroke}{rgb}{0.886275,0.898039,0.909804}%
\pgfsetstrokecolor{currentstroke}%
\pgfsetdash{}{0pt}%
\pgfpathmoveto{\pgfqpoint{0.584000in}{3.179809in}}%
\pgfpathlineto{\pgfqpoint{7.154000in}{3.179809in}}%
\pgfusepath{stroke}%
\end{pgfscope}%
\begin{pgfscope}%
\definecolor{textcolor}{rgb}{0.000000,0.000000,0.000000}%
\pgfsetstrokecolor{textcolor}%
\pgfsetfillcolor{textcolor}%
\pgftext[x=0.399812in, y=3.134373in, left, base]{\color{textcolor}{\fontfamily{\familydefault}\fontsize{9.500000}{11.400000}\selectfont\catcode`\^=\active\def^{\ifmmode\sp\else\^{}\fi}\catcode`\%=\active\def
\end{pgfscope}%
\begin{pgfscope}%
\pgfpathrectangle{\pgfqpoint{0.584000in}{2.894761in}}{\pgfqpoint{6.570000in}{1.425239in}}%
\pgfusepath{clip}%
\pgfsetrectcap%
\pgfsetroundjoin%
\pgfsetlinewidth{0.501875pt}%
\definecolor{currentstroke}{rgb}{0.886275,0.898039,0.909804}%
\pgfsetstrokecolor{currentstroke}%
\pgfsetdash{}{0pt}%
\pgfpathmoveto{\pgfqpoint{0.584000in}{3.464857in}}%
\pgfpathlineto{\pgfqpoint{7.154000in}{3.464857in}}%
\pgfusepath{stroke}%
\end{pgfscope}%
\begin{pgfscope}%
\definecolor{textcolor}{rgb}{0.000000,0.000000,0.000000}%
\pgfsetstrokecolor{textcolor}%
\pgfsetfillcolor{textcolor}%
\pgftext[x=0.399812in, y=3.419420in, left, base]{\color{textcolor}{\fontfamily{\familydefault}\fontsize{9.500000}{11.400000}\selectfont\catcode`\^=\active\def^{\ifmmode\sp\else\^{}\fi}\catcode`\%=\active\def
\end{pgfscope}%
\begin{pgfscope}%
\pgfpathrectangle{\pgfqpoint{0.584000in}{2.894761in}}{\pgfqpoint{6.570000in}{1.425239in}}%
\pgfusepath{clip}%
\pgfsetrectcap%
\pgfsetroundjoin%
\pgfsetlinewidth{0.501875pt}%
\definecolor{currentstroke}{rgb}{0.886275,0.898039,0.909804}%
\pgfsetstrokecolor{currentstroke}%
\pgfsetdash{}{0pt}%
\pgfpathmoveto{\pgfqpoint{0.584000in}{3.749905in}}%
\pgfpathlineto{\pgfqpoint{7.154000in}{3.749905in}}%
\pgfusepath{stroke}%
\end{pgfscope}%
\begin{pgfscope}%
\definecolor{textcolor}{rgb}{0.000000,0.000000,0.000000}%
\pgfsetstrokecolor{textcolor}%
\pgfsetfillcolor{textcolor}%
\pgftext[x=0.399812in, y=3.704468in, left, base]{\color{textcolor}{\fontfamily{\familydefault}\fontsize{9.500000}{11.400000}\selectfont\catcode`\^=\active\def^{\ifmmode\sp\else\^{}\fi}\catcode`\%=\active\def
\end{pgfscope}%
\begin{pgfscope}%
\pgfpathrectangle{\pgfqpoint{0.584000in}{2.894761in}}{\pgfqpoint{6.570000in}{1.425239in}}%
\pgfusepath{clip}%
\pgfsetrectcap%
\pgfsetroundjoin%
\pgfsetlinewidth{0.501875pt}%
\definecolor{currentstroke}{rgb}{0.886275,0.898039,0.909804}%
\pgfsetstrokecolor{currentstroke}%
\pgfsetdash{}{0pt}%
\pgfpathmoveto{\pgfqpoint{0.584000in}{4.034952in}}%
\pgfpathlineto{\pgfqpoint{7.154000in}{4.034952in}}%
\pgfusepath{stroke}%
\end{pgfscope}%
\begin{pgfscope}%
\definecolor{textcolor}{rgb}{0.000000,0.000000,0.000000}%
\pgfsetstrokecolor{textcolor}%
\pgfsetfillcolor{textcolor}%
\pgftext[x=0.399812in, y=3.989516in, left, base]{\color{textcolor}{\fontfamily{\familydefault}\fontsize{9.500000}{11.400000}\selectfont\catcode`\^=\active\def^{\ifmmode\sp\else\^{}\fi}\catcode`\%=\active\def
\end{pgfscope}%
\begin{pgfscope}%
\pgfpathrectangle{\pgfqpoint{0.584000in}{2.894761in}}{\pgfqpoint{6.570000in}{1.425239in}}%
\pgfusepath{clip}%
\pgfsetrectcap%
\pgfsetroundjoin%
\pgfsetlinewidth{0.501875pt}%
\definecolor{currentstroke}{rgb}{0.886275,0.898039,0.909804}%
\pgfsetstrokecolor{currentstroke}%
\pgfsetdash{}{0pt}%
\pgfpathmoveto{\pgfqpoint{0.584000in}{4.320000in}}%
\pgfpathlineto{\pgfqpoint{7.154000in}{4.320000in}}%
\pgfusepath{stroke}%
\end{pgfscope}%
\begin{pgfscope}%
\definecolor{textcolor}{rgb}{0.000000,0.000000,0.000000}%
\pgfsetstrokecolor{textcolor}%
\pgfsetfillcolor{textcolor}%
\pgftext[x=0.332024in, y=4.274564in, left, base]{\color{textcolor}{\fontfamily{\familydefault}\fontsize{9.500000}{11.400000}\selectfont\catcode`\^=\active\def^{\ifmmode\sp\else\^{}\fi}\catcode`\%=\active\def
\end{pgfscope}%
\begin{pgfscope}%
\definecolor{textcolor}{rgb}{0.000000,0.000000,0.000000}%
\pgfsetstrokecolor{textcolor}%
\pgfsetfillcolor{textcolor}%
\pgftext[x=0.276469in,y=3.607381in,,bottom,rotate=90.000000]{\color{textcolor}{\fontfamily{\familydefault}\fontsize{10.500000}{12.600000}\selectfont\catcode`\^=\active\def^{\ifmmode\sp\else\^{}\fi}\catcode`\%=\active\def
\end{pgfscope}%
\begin{pgfscope}%
\pgfpathrectangle{\pgfqpoint{0.584000in}{2.894761in}}{\pgfqpoint{6.570000in}{1.425239in}}%
\pgfusepath{clip}%
\pgfsetbuttcap%
\pgfsetmiterjoin%
\definecolor{currentfill}{rgb}{0.384314,0.423529,0.478431}%
\pgfsetfillcolor{currentfill}%
\pgfsetlinewidth{0.000000pt}%
\definecolor{currentstroke}{rgb}{0.000000,0.000000,0.000000}%
\pgfsetstrokecolor{currentstroke}%
\pgfsetstrokeopacity{0.000000}%
\pgfsetdash{}{0pt}%
\pgfpathmoveto{\pgfqpoint{0.882636in}{2.894761in}}%
\pgfpathlineto{\pgfqpoint{1.372204in}{2.894761in}}%
\pgfpathlineto{\pgfqpoint{1.372204in}{3.303235in}}%
\pgfpathlineto{\pgfqpoint{0.882636in}{3.303235in}}%
\pgfpathlineto{\pgfqpoint{0.882636in}{2.894761in}}%
\pgfpathclose%
\pgfusepath{fill}%
\end{pgfscope}%
\begin{pgfscope}%
\pgfpathrectangle{\pgfqpoint{0.584000in}{2.894761in}}{\pgfqpoint{6.570000in}{1.425239in}}%
\pgfusepath{clip}%
\pgfsetbuttcap%
\pgfsetmiterjoin%
\definecolor{currentfill}{rgb}{0.384314,0.423529,0.478431}%
\pgfsetfillcolor{currentfill}%
\pgfsetlinewidth{0.000000pt}%
\definecolor{currentstroke}{rgb}{0.000000,0.000000,0.000000}%
\pgfsetstrokecolor{currentstroke}%
\pgfsetstrokeopacity{0.000000}%
\pgfsetdash{}{0pt}%
\pgfpathmoveto{\pgfqpoint{3.330475in}{2.894761in}}%
\pgfpathlineto{\pgfqpoint{3.820043in}{2.894761in}}%
\pgfpathlineto{\pgfqpoint{3.820043in}{3.805204in}}%
\pgfpathlineto{\pgfqpoint{3.330475in}{3.805204in}}%
\pgfpathlineto{\pgfqpoint{3.330475in}{2.894761in}}%
\pgfpathclose%
\pgfusepath{fill}%
\end{pgfscope}%
\begin{pgfscope}%
\pgfpathrectangle{\pgfqpoint{0.584000in}{2.894761in}}{\pgfqpoint{6.570000in}{1.425239in}}%
\pgfusepath{clip}%
\pgfsetbuttcap%
\pgfsetmiterjoin%
\definecolor{currentfill}{rgb}{0.384314,0.423529,0.478431}%
\pgfsetfillcolor{currentfill}%
\pgfsetlinewidth{0.000000pt}%
\definecolor{currentstroke}{rgb}{0.000000,0.000000,0.000000}%
\pgfsetstrokecolor{currentstroke}%
\pgfsetstrokeopacity{0.000000}%
\pgfsetdash{}{0pt}%
\pgfpathmoveto{\pgfqpoint{5.778314in}{2.894761in}}%
\pgfpathlineto{\pgfqpoint{6.267882in}{2.894761in}}%
\pgfpathlineto{\pgfqpoint{6.267882in}{3.365803in}}%
\pgfpathlineto{\pgfqpoint{5.778314in}{3.365803in}}%
\pgfpathlineto{\pgfqpoint{5.778314in}{2.894761in}}%
\pgfpathclose%
\pgfusepath{fill}%
\end{pgfscope}%
\begin{pgfscope}%
\pgfpathrectangle{\pgfqpoint{0.584000in}{2.894761in}}{\pgfqpoint{6.570000in}{1.425239in}}%
\pgfusepath{clip}%
\pgfsetbuttcap%
\pgfsetmiterjoin%
\definecolor{currentfill}{rgb}{0.517647,0.439216,0.615686}%
\pgfsetfillcolor{currentfill}%
\pgfsetlinewidth{0.000000pt}%
\definecolor{currentstroke}{rgb}{0.000000,0.000000,0.000000}%
\pgfsetstrokecolor{currentstroke}%
\pgfsetstrokeopacity{0.000000}%
\pgfsetdash{}{0pt}%
\pgfpathmoveto{\pgfqpoint{1.470118in}{2.894761in}}%
\pgfpathlineto{\pgfqpoint{1.959686in}{2.894761in}}%
\pgfpathlineto{\pgfqpoint{1.959686in}{3.571037in}}%
\pgfpathlineto{\pgfqpoint{1.470118in}{3.571037in}}%
\pgfpathlineto{\pgfqpoint{1.470118in}{2.894761in}}%
\pgfpathclose%
\pgfusepath{fill}%
\end{pgfscope}%
\begin{pgfscope}%
\pgfpathrectangle{\pgfqpoint{0.584000in}{2.894761in}}{\pgfqpoint{6.570000in}{1.425239in}}%
\pgfusepath{clip}%
\pgfsetbuttcap%
\pgfsetmiterjoin%
\definecolor{currentfill}{rgb}{0.517647,0.439216,0.615686}%
\pgfsetfillcolor{currentfill}%
\pgfsetlinewidth{0.000000pt}%
\definecolor{currentstroke}{rgb}{0.000000,0.000000,0.000000}%
\pgfsetstrokecolor{currentstroke}%
\pgfsetstrokeopacity{0.000000}%
\pgfsetdash{}{0pt}%
\pgfpathmoveto{\pgfqpoint{0.000000in}{0.000000in}}%
\pgfpathlineto{\pgfqpoint{0.000000in}{0.000000in}}%
\pgfpathclose%
\pgfusepath{fill}%
\end{pgfscope}%
\begin{pgfscope}%
\pgfpathrectangle{\pgfqpoint{0.584000in}{2.894761in}}{\pgfqpoint{6.570000in}{1.425239in}}%
\pgfusepath{clip}%
\pgfsetbuttcap%
\pgfsetmiterjoin%
\definecolor{currentfill}{rgb}{0.517647,0.439216,0.615686}%
\pgfsetfillcolor{currentfill}%
\pgfsetlinewidth{0.000000pt}%
\definecolor{currentstroke}{rgb}{0.000000,0.000000,0.000000}%
\pgfsetstrokecolor{currentstroke}%
\pgfsetstrokeopacity{0.000000}%
\pgfsetdash{}{0pt}%
\pgfpathmoveto{\pgfqpoint{6.365796in}{2.894761in}}%
\pgfpathlineto{\pgfqpoint{6.855364in}{2.894761in}}%
\pgfpathlineto{\pgfqpoint{6.855364in}{3.605528in}}%
\pgfpathlineto{\pgfqpoint{6.365796in}{3.605528in}}%
\pgfpathlineto{\pgfqpoint{6.365796in}{2.894761in}}%
\pgfpathclose%
\pgfusepath{fill}%
\end{pgfscope}%
\begin{pgfscope}%
\pgfpathrectangle{\pgfqpoint{0.584000in}{2.894761in}}{\pgfqpoint{6.570000in}{1.425239in}}%
\pgfusepath{clip}%
\pgfsetbuttcap%
\pgfsetmiterjoin%
\definecolor{currentfill}{rgb}{0.094118,0.490196,0.513725}%
\pgfsetfillcolor{currentfill}%
\pgfsetlinewidth{0.000000pt}%
\definecolor{currentstroke}{rgb}{0.000000,0.000000,0.000000}%
\pgfsetstrokecolor{currentstroke}%
\pgfsetstrokeopacity{0.000000}%
\pgfsetdash{}{0pt}%
\pgfpathmoveto{\pgfqpoint{2.057599in}{2.894761in}}%
\pgfpathlineto{\pgfqpoint{2.547167in}{2.894761in}}%
\pgfpathlineto{\pgfqpoint{2.547167in}{2.971724in}}%
\pgfpathlineto{\pgfqpoint{2.057599in}{2.971724in}}%
\pgfpathlineto{\pgfqpoint{2.057599in}{2.894761in}}%
\pgfpathclose%
\pgfusepath{fill}%
\end{pgfscope}%
\begin{pgfscope}%
\pgfpathrectangle{\pgfqpoint{0.584000in}{2.894761in}}{\pgfqpoint{6.570000in}{1.425239in}}%
\pgfusepath{clip}%
\pgfsetbuttcap%
\pgfsetmiterjoin%
\definecolor{currentfill}{rgb}{0.094118,0.490196,0.513725}%
\pgfsetfillcolor{currentfill}%
\pgfsetlinewidth{0.000000pt}%
\definecolor{currentstroke}{rgb}{0.000000,0.000000,0.000000}%
\pgfsetstrokecolor{currentstroke}%
\pgfsetstrokeopacity{0.000000}%
\pgfsetdash{}{0pt}%
\pgfpathmoveto{\pgfqpoint{4.505438in}{2.894761in}}%
\pgfpathlineto{\pgfqpoint{4.995006in}{2.894761in}}%
\pgfpathlineto{\pgfqpoint{4.995006in}{3.667241in}}%
\pgfpathlineto{\pgfqpoint{4.505438in}{3.667241in}}%
\pgfpathlineto{\pgfqpoint{4.505438in}{2.894761in}}%
\pgfpathclose%
\pgfusepath{fill}%
\end{pgfscope}%
\begin{pgfscope}%
\pgfpathrectangle{\pgfqpoint{0.584000in}{2.894761in}}{\pgfqpoint{6.570000in}{1.425239in}}%
\pgfusepath{clip}%
\pgfsetbuttcap%
\pgfsetmiterjoin%
\definecolor{currentfill}{rgb}{0.094118,0.490196,0.513725}%
\pgfsetfillcolor{currentfill}%
\pgfsetlinewidth{0.000000pt}%
\definecolor{currentstroke}{rgb}{0.000000,0.000000,0.000000}%
\pgfsetstrokecolor{currentstroke}%
\pgfsetstrokeopacity{0.000000}%
\pgfsetdash{}{0pt}%
\pgfpathmoveto{\pgfqpoint{0.000000in}{0.000000in}}%
\pgfpathlineto{\pgfqpoint{0.000000in}{0.000000in}}%
\pgfpathclose%
\pgfusepath{fill}%
\end{pgfscope}%
\begin{pgfscope}%
\pgfsetrectcap%
\pgfsetmiterjoin%
\pgfsetlinewidth{0.803000pt}%
\definecolor{currentstroke}{rgb}{0.713725,0.741176,0.768627}%
\pgfsetstrokecolor{currentstroke}%
\pgfsetdash{}{0pt}%
\pgfpathmoveto{\pgfqpoint{0.584000in}{2.894761in}}%
\pgfpathlineto{\pgfqpoint{7.154000in}{2.894761in}}%
\pgfusepath{stroke}%
\end{pgfscope}%
\begin{pgfscope}%
\definecolor{textcolor}{rgb}{0.000000,0.000000,0.000000}%
\pgfsetstrokecolor{textcolor}%
\pgfsetfillcolor{textcolor}%
\pgftext[x=1.127420in,y=3.331740in,,base]{\color{textcolor}{\fontfamily{\familydefault}\fontsize{9.500000}{11.400000}\selectfont\catcode`\^=\active\def^{\ifmmode\sp\else\^{}\fi}\catcode`\%=\active\def
\end{pgfscope}%
\begin{pgfscope}%
\definecolor{textcolor}{rgb}{0.000000,0.000000,0.000000}%
\pgfsetstrokecolor{textcolor}%
\pgfsetfillcolor{textcolor}%
\pgftext[x=3.575259in,y=3.833709in,,base]{\color{textcolor}{\fontfamily{\familydefault}\fontsize{9.500000}{11.400000}\selectfont\catcode`\^=\active\def^{\ifmmode\sp\else\^{}\fi}\catcode`\%=\active\def
\end{pgfscope}%
\begin{pgfscope}%
\definecolor{textcolor}{rgb}{0.000000,0.000000,0.000000}%
\pgfsetstrokecolor{textcolor}%
\pgfsetfillcolor{textcolor}%
\pgftext[x=6.023098in,y=3.394308in,,base]{\color{textcolor}{\fontfamily{\familydefault}\fontsize{9.500000}{11.400000}\selectfont\catcode`\^=\active\def^{\ifmmode\sp\else\^{}\fi}\catcode`\%=\active\def
\end{pgfscope}%
\begin{pgfscope}%
\definecolor{textcolor}{rgb}{0.000000,0.000000,0.000000}%
\pgfsetstrokecolor{textcolor}%
\pgfsetfillcolor{textcolor}%
\pgftext[x=1.714902in,y=3.599542in,,base]{\color{textcolor}{\fontfamily{\familydefault}\fontsize{9.500000}{11.400000}\selectfont\catcode`\^=\active\def^{\ifmmode\sp\else\^{}\fi}\catcode`\%=\active\def
\end{pgfscope}%
\begin{pgfscope}%
\definecolor{textcolor}{rgb}{0.384314,0.423529,0.478431}%
\pgfsetstrokecolor{textcolor}%
\pgfsetfillcolor{textcolor}%
\pgftext[x=4.162741in,y=2.937519in,,base]{\color{textcolor}{\fontfamily{\familydefault}\fontsize{9.500000}{11.400000}\selectfont\catcode`\^=\active\def^{\ifmmode\sp\else\^{}\fi}\catcode`\%=\active\def
\end{pgfscope}%
\begin{pgfscope}%
\definecolor{textcolor}{rgb}{0.000000,0.000000,0.000000}%
\pgfsetstrokecolor{textcolor}%
\pgfsetfillcolor{textcolor}%
\pgftext[x=6.610580in,y=3.634033in,,base]{\color{textcolor}{\fontfamily{\familydefault}\fontsize{9.500000}{11.400000}\selectfont\catcode`\^=\active\def^{\ifmmode\sp\else\^{}\fi}\catcode`\%=\active\def
\end{pgfscope}%
\begin{pgfscope}%
\definecolor{textcolor}{rgb}{0.000000,0.000000,0.000000}%
\pgfsetstrokecolor{textcolor}%
\pgfsetfillcolor{textcolor}%
\pgftext[x=2.302383in,y=3.000229in,,base]{\color{textcolor}{\fontfamily{\familydefault}\fontsize{9.500000}{11.400000}\selectfont\catcode`\^=\active\def^{\ifmmode\sp\else\^{}\fi}\catcode`\%=\active\def
\end{pgfscope}%
\begin{pgfscope}%
\definecolor{textcolor}{rgb}{0.000000,0.000000,0.000000}%
\pgfsetstrokecolor{textcolor}%
\pgfsetfillcolor{textcolor}%
\pgftext[x=4.750222in,y=3.695745in,,base]{\color{textcolor}{\fontfamily{\familydefault}\fontsize{9.500000}{11.400000}\selectfont\catcode`\^=\active\def^{\ifmmode\sp\else\^{}\fi}\catcode`\%=\active\def
\end{pgfscope}%
\begin{pgfscope}%
\definecolor{textcolor}{rgb}{0.384314,0.423529,0.478431}%
\pgfsetstrokecolor{textcolor}%
\pgfsetfillcolor{textcolor}%
\pgftext[x=7.198061in,y=2.937519in,,base]{\color{textcolor}{\fontfamily{\familydefault}\fontsize{9.500000}{11.400000}\selectfont\catcode`\^=\active\def^{\ifmmode\sp\else\^{}\fi}\catcode`\%=\active\def
\end{pgfscope}%
\begin{pgfscope}%
\definecolor{textcolor}{rgb}{0.000000,0.000000,0.000000}%
\pgfsetstrokecolor{textcolor}%
\pgfsetfillcolor{textcolor}%
\pgftext[x=0.584000in,y=4.667222in,left,base]{\color{textcolor}{\fontfamily{\familydefault}\fontsize{11.500000}{13.800000}\selectfont\catcode`\^=\active\def^{\ifmmode\sp\else\^{}\fi}\catcode`\%=\active\def
\end{pgfscope}%
\begin{pgfscope}%
\pgfsetbuttcap%
\pgfsetmiterjoin%
\definecolor{currentfill}{rgb}{0.384314,0.423529,0.478431}%
\pgfsetfillcolor{currentfill}%
\pgfsetlinewidth{0.000000pt}%
\definecolor{currentstroke}{rgb}{0.000000,0.000000,0.000000}%
\pgfsetstrokecolor{currentstroke}%
\pgfsetstrokeopacity{0.000000}%
\pgfsetdash{}{0pt}%
\pgfpathmoveto{\pgfqpoint{0.637050in}{4.351179in}}%
\pgfpathlineto{\pgfqpoint{0.900939in}{4.351179in}}%
\pgfpathlineto{\pgfqpoint{0.900939in}{4.443541in}}%
\pgfpathlineto{\pgfqpoint{0.637050in}{4.443541in}}%
\pgfpathlineto{\pgfqpoint{0.637050in}{4.351179in}}%
\pgfpathclose%
\pgfusepath{fill}%
\end{pgfscope}%
\begin{pgfscope}%
\definecolor{textcolor}{rgb}{0.000000,0.000000,0.000000}%
\pgfsetstrokecolor{textcolor}%
\pgfsetfillcolor{textcolor}%
\pgftext[x=1.006494in,y=4.351179in,left,base]{\color{textcolor}{\fontfamily{\familydefault}\fontsize{9.500000}{11.400000}\selectfont\catcode`\^=\active\def^{\ifmmode\sp\else\^{}\fi}\catcode`\%=\active\def
\end{pgfscope}%
\begin{pgfscope}%
\pgfsetbuttcap%
\pgfsetmiterjoin%
\definecolor{currentfill}{rgb}{0.517647,0.439216,0.615686}%
\pgfsetfillcolor{currentfill}%
\pgfsetlinewidth{0.000000pt}%
\definecolor{currentstroke}{rgb}{0.000000,0.000000,0.000000}%
\pgfsetstrokecolor{currentstroke}%
\pgfsetstrokeopacity{0.000000}%
\pgfsetdash{}{0pt}%
\pgfpathmoveto{\pgfqpoint{2.078609in}{4.351179in}}%
\pgfpathlineto{\pgfqpoint{2.342498in}{4.351179in}}%
\pgfpathlineto{\pgfqpoint{2.342498in}{4.443541in}}%
\pgfpathlineto{\pgfqpoint{2.078609in}{4.443541in}}%
\pgfpathlineto{\pgfqpoint{2.078609in}{4.351179in}}%
\pgfpathclose%
\pgfusepath{fill}%
\end{pgfscope}%
\begin{pgfscope}%
\definecolor{textcolor}{rgb}{0.000000,0.000000,0.000000}%
\pgfsetstrokecolor{textcolor}%
\pgfsetfillcolor{textcolor}%
\pgftext[x=2.448053in,y=4.351179in,left,base]{\color{textcolor}{\fontfamily{\familydefault}\fontsize{9.500000}{11.400000}\selectfont\catcode`\^=\active\def^{\ifmmode\sp\else\^{}\fi}\catcode`\%=\active\def
\end{pgfscope}%
\begin{pgfscope}%
\pgfsetbuttcap%
\pgfsetmiterjoin%
\definecolor{currentfill}{rgb}{0.094118,0.490196,0.513725}%
\pgfsetfillcolor{currentfill}%
\pgfsetlinewidth{0.000000pt}%
\definecolor{currentstroke}{rgb}{0.000000,0.000000,0.000000}%
\pgfsetstrokecolor{currentstroke}%
\pgfsetstrokeopacity{0.000000}%
\pgfsetdash{}{0pt}%
\pgfpathmoveto{\pgfqpoint{3.697037in}{4.351179in}}%
\pgfpathlineto{\pgfqpoint{3.960926in}{4.351179in}}%
\pgfpathlineto{\pgfqpoint{3.960926in}{4.443541in}}%
\pgfpathlineto{\pgfqpoint{3.697037in}{4.443541in}}%
\pgfpathlineto{\pgfqpoint{3.697037in}{4.351179in}}%
\pgfpathclose%
\pgfusepath{fill}%
\end{pgfscope}%
\begin{pgfscope}%
\definecolor{textcolor}{rgb}{0.000000,0.000000,0.000000}%
\pgfsetstrokecolor{textcolor}%
\pgfsetfillcolor{textcolor}%
\pgftext[x=4.066482in,y=4.351179in,left,base]{\color{textcolor}{\fontfamily{\familydefault}\fontsize{9.500000}{11.400000}\selectfont\catcode`\^=\active\def^{\ifmmode\sp\else\^{}\fi}\catcode`\%=\active\def
\end{pgfscope}%
\begin{pgfscope}%
\pgfsetbuttcap%
\pgfsetmiterjoin%
\definecolor{currentfill}{rgb}{1.000000,1.000000,1.000000}%
\pgfsetfillcolor{currentfill}%
\pgfsetlinewidth{0.000000pt}%
\definecolor{currentstroke}{rgb}{0.000000,0.000000,0.000000}%
\pgfsetstrokecolor{currentstroke}%
\pgfsetstrokeopacity{0.000000}%
\pgfsetdash{}{0pt}%
\pgfpathmoveto{\pgfqpoint{0.584000in}{0.480000in}}%
\pgfpathlineto{\pgfqpoint{3.415897in}{0.480000in}}%
\pgfpathlineto{\pgfqpoint{3.415897in}{1.837370in}}%
\pgfpathlineto{\pgfqpoint{0.584000in}{1.837370in}}%
\pgfpathlineto{\pgfqpoint{0.584000in}{0.480000in}}%
\pgfpathclose%
\pgfusepath{fill}%
\end{pgfscope}%
\begin{pgfscope}%
\definecolor{textcolor}{rgb}{0.000000,0.000000,0.000000}%
\pgfsetstrokecolor{textcolor}%
\pgfsetfillcolor{textcolor}%
\pgftext[x=1.210239in,y=0.431389in,,top]{\color{textcolor}{\fontfamily{\familydefault}\fontsize{9.500000}{11.400000}\selectfont\catcode`\^=\active\def^{\ifmmode\sp\else\^{}\fi}\catcode`\%=\active\def
\end{pgfscope}%
\begin{pgfscope}%
\definecolor{textcolor}{rgb}{0.000000,0.000000,0.000000}%
\pgfsetstrokecolor{textcolor}%
\pgfsetfillcolor{textcolor}%
\pgftext[x=2.474641in, y=0.340516in, left, base]{\color{textcolor}{\fontfamily{\familydefault}\fontsize{9.500000}{11.400000}\selectfont\catcode`\^=\active\def^{\ifmmode\sp\else\^{}\fi}\catcode`\%=\active\def
\end{pgfscope}%
\begin{pgfscope}%
\definecolor{textcolor}{rgb}{0.000000,0.000000,0.000000}%
\pgfsetstrokecolor{textcolor}%
\pgfsetfillcolor{textcolor}%
\pgftext[x=2.441332in, y=0.205819in, left, base]{\color{textcolor}{\fontfamily{\familydefault}\fontsize{9.500000}{11.400000}\selectfont\catcode`\^=\active\def^{\ifmmode\sp\else\^{}\fi}\catcode`\%=\active\def
\end{pgfscope}%
\begin{pgfscope}%
\pgfpathrectangle{\pgfqpoint{0.584000in}{0.480000in}}{\pgfqpoint{2.831897in}{1.357370in}}%
\pgfusepath{clip}%
\pgfsetrectcap%
\pgfsetroundjoin%
\pgfsetlinewidth{0.501875pt}%
\definecolor{currentstroke}{rgb}{0.886275,0.898039,0.909804}%
\pgfsetstrokecolor{currentstroke}%
\pgfsetdash{}{0pt}%
\pgfpathmoveto{\pgfqpoint{0.584000in}{0.480000in}}%
\pgfpathlineto{\pgfqpoint{3.415897in}{0.480000in}}%
\pgfusepath{stroke}%
\end{pgfscope}%
\begin{pgfscope}%
\definecolor{textcolor}{rgb}{0.000000,0.000000,0.000000}%
\pgfsetstrokecolor{textcolor}%
\pgfsetfillcolor{textcolor}%
\pgftext[x=0.467601in, y=0.434564in, left, base]{\color{textcolor}{\fontfamily{\familydefault}\fontsize{9.500000}{11.400000}\selectfont\catcode`\^=\active\def^{\ifmmode\sp\else\^{}\fi}\catcode`\%=\active\def
\end{pgfscope}%
\begin{pgfscope}%
\pgfpathrectangle{\pgfqpoint{0.584000in}{0.480000in}}{\pgfqpoint{2.831897in}{1.357370in}}%
\pgfusepath{clip}%
\pgfsetrectcap%
\pgfsetroundjoin%
\pgfsetlinewidth{0.501875pt}%
\definecolor{currentstroke}{rgb}{0.886275,0.898039,0.909804}%
\pgfsetstrokecolor{currentstroke}%
\pgfsetdash{}{0pt}%
\pgfpathmoveto{\pgfqpoint{0.584000in}{1.070161in}}%
\pgfpathlineto{\pgfqpoint{3.415897in}{1.070161in}}%
\pgfusepath{stroke}%
\end{pgfscope}%
\begin{pgfscope}%
\definecolor{textcolor}{rgb}{0.000000,0.000000,0.000000}%
\pgfsetstrokecolor{textcolor}%
\pgfsetfillcolor{textcolor}%
\pgftext[x=0.399812in, y=1.024724in, left, base]{\color{textcolor}{\fontfamily{\familydefault}\fontsize{9.500000}{11.400000}\selectfont\catcode`\^=\active\def^{\ifmmode\sp\else\^{}\fi}\catcode`\%=\active\def
\end{pgfscope}%
\begin{pgfscope}%
\pgfpathrectangle{\pgfqpoint{0.584000in}{0.480000in}}{\pgfqpoint{2.831897in}{1.357370in}}%
\pgfusepath{clip}%
\pgfsetrectcap%
\pgfsetroundjoin%
\pgfsetlinewidth{0.501875pt}%
\definecolor{currentstroke}{rgb}{0.886275,0.898039,0.909804}%
\pgfsetstrokecolor{currentstroke}%
\pgfsetdash{}{0pt}%
\pgfpathmoveto{\pgfqpoint{0.584000in}{1.660322in}}%
\pgfpathlineto{\pgfqpoint{3.415897in}{1.660322in}}%
\pgfusepath{stroke}%
\end{pgfscope}%
\begin{pgfscope}%
\definecolor{textcolor}{rgb}{0.000000,0.000000,0.000000}%
\pgfsetstrokecolor{textcolor}%
\pgfsetfillcolor{textcolor}%
\pgftext[x=0.332024in, y=1.614885in, left, base]{\color{textcolor}{\fontfamily{\familydefault}\fontsize{9.500000}{11.400000}\selectfont\catcode`\^=\active\def^{\ifmmode\sp\else\^{}\fi}\catcode`\%=\active\def
\end{pgfscope}%
\begin{pgfscope}%
\definecolor{textcolor}{rgb}{0.000000,0.000000,0.000000}%
\pgfsetstrokecolor{textcolor}%
\pgfsetfillcolor{textcolor}%
\pgftext[x=0.276469in,y=1.158685in,,bottom,rotate=90.000000]{\color{textcolor}{\fontfamily{\familydefault}\fontsize{10.500000}{12.600000}\selectfont\catcode`\^=\active\def^{\ifmmode\sp\else\^{}\fi}\catcode`\%=\active\def
\end{pgfscope}%
\begin{pgfscope}%
\pgfpathrectangle{\pgfqpoint{0.584000in}{0.480000in}}{\pgfqpoint{2.831897in}{1.357370in}}%
\pgfusepath{clip}%
\pgfsetbuttcap%
\pgfsetmiterjoin%
\definecolor{currentfill}{rgb}{0.384314,0.423529,0.478431}%
\pgfsetfillcolor{currentfill}%
\pgfsetlinewidth{0.000000pt}%
\definecolor{currentstroke}{rgb}{0.000000,0.000000,0.000000}%
\pgfsetstrokecolor{currentstroke}%
\pgfsetstrokeopacity{0.000000}%
\pgfsetdash{}{0pt}%
\pgfpathmoveto{\pgfqpoint{0.712723in}{0.480000in}}%
\pgfpathlineto{\pgfqpoint{1.170754in}{0.480000in}}%
\pgfpathlineto{\pgfqpoint{1.170754in}{1.659142in}}%
\pgfpathlineto{\pgfqpoint{0.712723in}{1.659142in}}%
\pgfpathlineto{\pgfqpoint{0.712723in}{0.480000in}}%
\pgfpathclose%
\pgfusepath{fill}%
\end{pgfscope}%
\begin{pgfscope}%
\pgfpathrectangle{\pgfqpoint{0.584000in}{0.480000in}}{\pgfqpoint{2.831897in}{1.357370in}}%
\pgfusepath{clip}%
\pgfsetbuttcap%
\pgfsetmiterjoin%
\definecolor{currentfill}{rgb}{0.384314,0.423529,0.478431}%
\pgfsetfillcolor{currentfill}%
\pgfsetlinewidth{0.000000pt}%
\definecolor{currentstroke}{rgb}{0.000000,0.000000,0.000000}%
\pgfsetstrokecolor{currentstroke}%
\pgfsetstrokeopacity{0.000000}%
\pgfsetdash{}{0pt}%
\pgfpathmoveto{\pgfqpoint{2.292141in}{0.480000in}}%
\pgfpathlineto{\pgfqpoint{2.750172in}{0.480000in}}%
\pgfpathlineto{\pgfqpoint{2.750172in}{1.602486in}}%
\pgfpathlineto{\pgfqpoint{2.292141in}{1.602486in}}%
\pgfpathlineto{\pgfqpoint{2.292141in}{0.480000in}}%
\pgfpathclose%
\pgfusepath{fill}%
\end{pgfscope}%
\begin{pgfscope}%
\pgfpathrectangle{\pgfqpoint{0.584000in}{0.480000in}}{\pgfqpoint{2.831897in}{1.357370in}}%
\pgfusepath{clip}%
\pgfsetbuttcap%
\pgfsetmiterjoin%
\definecolor{currentfill}{rgb}{0.094118,0.490196,0.513725}%
\pgfsetfillcolor{currentfill}%
\pgfsetlinewidth{0.000000pt}%
\definecolor{currentstroke}{rgb}{0.000000,0.000000,0.000000}%
\pgfsetstrokecolor{currentstroke}%
\pgfsetstrokeopacity{0.000000}%
\pgfsetdash{}{0pt}%
\pgfpathmoveto{\pgfqpoint{1.249725in}{0.480000in}}%
\pgfpathlineto{\pgfqpoint{1.707756in}{0.480000in}}%
\pgfpathlineto{\pgfqpoint{1.707756in}{1.644978in}}%
\pgfpathlineto{\pgfqpoint{1.249725in}{1.644978in}}%
\pgfpathlineto{\pgfqpoint{1.249725in}{0.480000in}}%
\pgfpathclose%
\pgfusepath{fill}%
\end{pgfscope}%
\begin{pgfscope}%
\pgfpathrectangle{\pgfqpoint{0.584000in}{0.480000in}}{\pgfqpoint{2.831897in}{1.357370in}}%
\pgfusepath{clip}%
\pgfsetbuttcap%
\pgfsetmiterjoin%
\definecolor{currentfill}{rgb}{0.094118,0.490196,0.513725}%
\pgfsetfillcolor{currentfill}%
\pgfsetlinewidth{0.000000pt}%
\definecolor{currentstroke}{rgb}{0.000000,0.000000,0.000000}%
\pgfsetstrokecolor{currentstroke}%
\pgfsetstrokeopacity{0.000000}%
\pgfsetdash{}{0pt}%
\pgfpathmoveto{\pgfqpoint{2.829143in}{0.480000in}}%
\pgfpathlineto{\pgfqpoint{3.287174in}{0.480000in}}%
\pgfpathlineto{\pgfqpoint{3.287174in}{1.478552in}}%
\pgfpathlineto{\pgfqpoint{2.829143in}{1.478552in}}%
\pgfpathlineto{\pgfqpoint{2.829143in}{0.480000in}}%
\pgfpathclose%
\pgfusepath{fill}%
\end{pgfscope}%
\begin{pgfscope}%
\pgfsetrectcap%
\pgfsetmiterjoin%
\pgfsetlinewidth{0.803000pt}%
\definecolor{currentstroke}{rgb}{0.713725,0.741176,0.768627}%
\pgfsetstrokecolor{currentstroke}%
\pgfsetdash{}{0pt}%
\pgfpathmoveto{\pgfqpoint{0.584000in}{0.480000in}}%
\pgfpathlineto{\pgfqpoint{3.415897in}{0.480000in}}%
\pgfusepath{stroke}%
\end{pgfscope}%
\begin{pgfscope}%
\definecolor{textcolor}{rgb}{0.000000,0.000000,0.000000}%
\pgfsetstrokecolor{textcolor}%
\pgfsetfillcolor{textcolor}%
\pgftext[x=0.941738in,y=1.682748in,,base]{\color{textcolor}{\fontfamily{\familydefault}\fontsize{9.500000}{11.400000}\selectfont\catcode`\^=\active\def^{\ifmmode\sp\else\^{}\fi}\catcode`\%=\active\def
\end{pgfscope}%
\begin{pgfscope}%
\definecolor{textcolor}{rgb}{0.000000,0.000000,0.000000}%
\pgfsetstrokecolor{textcolor}%
\pgfsetfillcolor{textcolor}%
\pgftext[x=2.521156in,y=1.626092in,,base]{\color{textcolor}{\fontfamily{\familydefault}\fontsize{9.500000}{11.400000}\selectfont\catcode`\^=\active\def^{\ifmmode\sp\else\^{}\fi}\catcode`\%=\active\def
\end{pgfscope}%
\begin{pgfscope}%
\definecolor{textcolor}{rgb}{0.000000,0.000000,0.000000}%
\pgfsetstrokecolor{textcolor}%
\pgfsetfillcolor{textcolor}%
\pgftext[x=1.478740in,y=1.668584in,,base]{\color{textcolor}{\fontfamily{\familydefault}\fontsize{9.500000}{11.400000}\selectfont\catcode`\^=\active\def^{\ifmmode\sp\else\^{}\fi}\catcode`\%=\active\def
\end{pgfscope}%
\begin{pgfscope}%
\definecolor{textcolor}{rgb}{0.000000,0.000000,0.000000}%
\pgfsetstrokecolor{textcolor}%
\pgfsetfillcolor{textcolor}%
\pgftext[x=3.058158in,y=1.502159in,,base]{\color{textcolor}{\fontfamily{\familydefault}\fontsize{9.500000}{11.400000}\selectfont\catcode`\^=\active\def^{\ifmmode\sp\else\^{}\fi}\catcode`\%=\active\def
\end{pgfscope}%
\begin{pgfscope}%
\definecolor{textcolor}{rgb}{0.000000,0.000000,0.000000}%
\pgfsetstrokecolor{textcolor}%
\pgfsetfillcolor{textcolor}%
\pgftext[x=0.584000in,y=2.087370in,left,base]{\color{textcolor}{\fontfamily{\familydefault}\fontsize{11.500000}{13.800000}\selectfont\catcode`\^=\active\def^{\ifmmode\sp\else\^{}\fi}\catcode`\%=\active\def
\end{pgfscope}%
\begin{pgfscope}%
\pgfsetbuttcap%
\pgfsetmiterjoin%
\definecolor{currentfill}{rgb}{0.384314,0.423529,0.478431}%
\pgfsetfillcolor{currentfill}%
\pgfsetlinewidth{0.000000pt}%
\definecolor{currentstroke}{rgb}{0.000000,0.000000,0.000000}%
\pgfsetstrokecolor{currentstroke}%
\pgfsetstrokeopacity{0.000000}%
\pgfsetdash{}{0pt}%
\pgfpathmoveto{\pgfqpoint{0.617793in}{1.829864in}}%
\pgfpathlineto{\pgfqpoint{0.881682in}{1.829864in}}%
\pgfpathlineto{\pgfqpoint{0.881682in}{1.922226in}}%
\pgfpathlineto{\pgfqpoint{0.617793in}{1.922226in}}%
\pgfpathlineto{\pgfqpoint{0.617793in}{1.829864in}}%
\pgfpathclose%
\pgfusepath{fill}%
\end{pgfscope}%
\begin{pgfscope}%
\definecolor{textcolor}{rgb}{0.000000,0.000000,0.000000}%
\pgfsetstrokecolor{textcolor}%
\pgfsetfillcolor{textcolor}%
\pgftext[x=0.987238in,y=1.829864in,left,base]{\color{textcolor}{\fontfamily{\familydefault}\fontsize{9.500000}{11.400000}\selectfont\catcode`\^=\active\def^{\ifmmode\sp\else\^{}\fi}\catcode`\%=\active\def
\end{pgfscope}%
\begin{pgfscope}%
\pgfsetbuttcap%
\pgfsetmiterjoin%
\definecolor{currentfill}{rgb}{0.094118,0.490196,0.513725}%
\pgfsetfillcolor{currentfill}%
\pgfsetlinewidth{0.000000pt}%
\definecolor{currentstroke}{rgb}{0.000000,0.000000,0.000000}%
\pgfsetstrokecolor{currentstroke}%
\pgfsetstrokeopacity{0.000000}%
\pgfsetdash{}{0pt}%
\pgfpathmoveto{\pgfqpoint{1.772688in}{1.829864in}}%
\pgfpathlineto{\pgfqpoint{2.036577in}{1.829864in}}%
\pgfpathlineto{\pgfqpoint{2.036577in}{1.922226in}}%
\pgfpathlineto{\pgfqpoint{1.772688in}{1.922226in}}%
\pgfpathlineto{\pgfqpoint{1.772688in}{1.829864in}}%
\pgfpathclose%
\pgfusepath{fill}%
\end{pgfscope}%
\begin{pgfscope}%
\definecolor{textcolor}{rgb}{0.000000,0.000000,0.000000}%
\pgfsetstrokecolor{textcolor}%
\pgfsetfillcolor{textcolor}%
\pgftext[x=2.142132in,y=1.829864in,left,base]{\color{textcolor}{\fontfamily{\familydefault}\fontsize{9.500000}{11.400000}\selectfont\catcode`\^=\active\def^{\ifmmode\sp\else\^{}\fi}\catcode`\%=\active\def
\end{pgfscope}%
\begin{pgfscope}%
\pgfsetbuttcap%
\pgfsetmiterjoin%
\definecolor{currentfill}{rgb}{1.000000,1.000000,1.000000}%
\pgfsetfillcolor{currentfill}%
\pgfsetlinewidth{0.000000pt}%
\definecolor{currentstroke}{rgb}{0.000000,0.000000,0.000000}%
\pgfsetstrokecolor{currentstroke}%
\pgfsetstrokeopacity{0.000000}%
\pgfsetdash{}{0pt}%
\pgfpathmoveto{\pgfqpoint{4.322103in}{0.480000in}}%
\pgfpathlineto{\pgfqpoint{7.154000in}{0.480000in}}%
\pgfpathlineto{\pgfqpoint{7.154000in}{1.837370in}}%
\pgfpathlineto{\pgfqpoint{4.322103in}{1.837370in}}%
\pgfpathlineto{\pgfqpoint{4.322103in}{0.480000in}}%
\pgfpathclose%
\pgfusepath{fill}%
\end{pgfscope}%
\begin{pgfscope}%
\definecolor{textcolor}{rgb}{0.000000,0.000000,0.000000}%
\pgfsetstrokecolor{textcolor}%
\pgfsetfillcolor{textcolor}%
\pgftext[x=4.868305in,y=0.431389in,,top]{\color{textcolor}{\fontfamily{\familydefault}\fontsize{9.500000}{11.400000}\selectfont\catcode`\^=\active\def^{\ifmmode\sp\else\^{}\fi}\catcode`\%=\active\def
\end{pgfscope}%
\begin{pgfscope}%
\definecolor{textcolor}{rgb}{0.000000,0.000000,0.000000}%
\pgfsetstrokecolor{textcolor}%
\pgfsetfillcolor{textcolor}%
\pgftext[x=6.607799in,y=0.431389in,,top]{\color{textcolor}{\fontfamily{\familydefault}\fontsize{9.500000}{11.400000}\selectfont\catcode`\^=\active\def^{\ifmmode\sp\else\^{}\fi}\catcode`\%=\active\def
\end{pgfscope}%
\begin{pgfscope}%
\pgfpathrectangle{\pgfqpoint{4.322103in}{0.480000in}}{\pgfqpoint{2.831897in}{1.357370in}}%
\pgfusepath{clip}%
\pgfsetrectcap%
\pgfsetroundjoin%
\pgfsetlinewidth{0.501875pt}%
\definecolor{currentstroke}{rgb}{0.886275,0.898039,0.909804}%
\pgfsetstrokecolor{currentstroke}%
\pgfsetdash{}{0pt}%
\pgfpathmoveto{\pgfqpoint{4.322103in}{0.480000in}}%
\pgfpathlineto{\pgfqpoint{7.154000in}{0.480000in}}%
\pgfusepath{stroke}%
\end{pgfscope}%
\begin{pgfscope}%
\definecolor{textcolor}{rgb}{0.000000,0.000000,0.000000}%
\pgfsetstrokecolor{textcolor}%
\pgfsetfillcolor{textcolor}%
\pgftext[x=4.205704in, y=0.434564in, left, base]{\color{textcolor}{\fontfamily{\familydefault}\fontsize{9.500000}{11.400000}\selectfont\catcode`\^=\active\def^{\ifmmode\sp\else\^{}\fi}\catcode`\%=\active\def
\end{pgfscope}%
\begin{pgfscope}%
\pgfpathrectangle{\pgfqpoint{4.322103in}{0.480000in}}{\pgfqpoint{2.831897in}{1.357370in}}%
\pgfusepath{clip}%
\pgfsetrectcap%
\pgfsetroundjoin%
\pgfsetlinewidth{0.501875pt}%
\definecolor{currentstroke}{rgb}{0.886275,0.898039,0.909804}%
\pgfsetstrokecolor{currentstroke}%
\pgfsetdash{}{0pt}%
\pgfpathmoveto{\pgfqpoint{4.322103in}{1.070161in}}%
\pgfpathlineto{\pgfqpoint{7.154000in}{1.070161in}}%
\pgfusepath{stroke}%
\end{pgfscope}%
\begin{pgfscope}%
\definecolor{textcolor}{rgb}{0.000000,0.000000,0.000000}%
\pgfsetstrokecolor{textcolor}%
\pgfsetfillcolor{textcolor}%
\pgftext[x=4.137916in, y=1.024724in, left, base]{\color{textcolor}{\fontfamily{\familydefault}\fontsize{9.500000}{11.400000}\selectfont\catcode`\^=\active\def^{\ifmmode\sp\else\^{}\fi}\catcode`\%=\active\def
\end{pgfscope}%
\begin{pgfscope}%
\pgfpathrectangle{\pgfqpoint{4.322103in}{0.480000in}}{\pgfqpoint{2.831897in}{1.357370in}}%
\pgfusepath{clip}%
\pgfsetrectcap%
\pgfsetroundjoin%
\pgfsetlinewidth{0.501875pt}%
\definecolor{currentstroke}{rgb}{0.886275,0.898039,0.909804}%
\pgfsetstrokecolor{currentstroke}%
\pgfsetdash{}{0pt}%
\pgfpathmoveto{\pgfqpoint{4.322103in}{1.660322in}}%
\pgfpathlineto{\pgfqpoint{7.154000in}{1.660322in}}%
\pgfusepath{stroke}%
\end{pgfscope}%
\begin{pgfscope}%
\definecolor{textcolor}{rgb}{0.000000,0.000000,0.000000}%
\pgfsetstrokecolor{textcolor}%
\pgfsetfillcolor{textcolor}%
\pgftext[x=4.070128in, y=1.614885in, left, base]{\color{textcolor}{\fontfamily{\familydefault}\fontsize{9.500000}{11.400000}\selectfont\catcode`\^=\active\def^{\ifmmode\sp\else\^{}\fi}\catcode`\%=\active\def
\end{pgfscope}%
\begin{pgfscope}%
\definecolor{textcolor}{rgb}{0.000000,0.000000,0.000000}%
\pgfsetstrokecolor{textcolor}%
\pgfsetfillcolor{textcolor}%
\pgftext[x=4.014572in,y=1.158685in,,bottom,rotate=90.000000]{\color{textcolor}{\fontfamily{\familydefault}\fontsize{10.500000}{12.600000}\selectfont\catcode`\^=\active\def^{\ifmmode\sp\else\^{}\fi}\catcode`\%=\active\def
\end{pgfscope}%
\begin{pgfscope}%
\pgfpathrectangle{\pgfqpoint{4.322103in}{0.480000in}}{\pgfqpoint{2.831897in}{1.357370in}}%
\pgfusepath{clip}%
\pgfsetbuttcap%
\pgfsetmiterjoin%
\definecolor{currentfill}{rgb}{0.384314,0.423529,0.478431}%
\pgfsetfillcolor{currentfill}%
\pgfsetlinewidth{0.000000pt}%
\definecolor{currentstroke}{rgb}{0.000000,0.000000,0.000000}%
\pgfsetstrokecolor{currentstroke}%
\pgfsetstrokeopacity{0.000000}%
\pgfsetdash{}{0pt}%
\pgfpathmoveto{\pgfqpoint{4.450826in}{0.480000in}}%
\pgfpathlineto{\pgfqpoint{5.285783in}{0.480000in}}%
\pgfpathlineto{\pgfqpoint{5.285783in}{0.582688in}}%
\pgfpathlineto{\pgfqpoint{4.450826in}{0.582688in}}%
\pgfpathlineto{\pgfqpoint{4.450826in}{0.480000in}}%
\pgfpathclose%
\pgfusepath{fill}%
\end{pgfscope}%
\begin{pgfscope}%
\pgfpathrectangle{\pgfqpoint{4.322103in}{0.480000in}}{\pgfqpoint{2.831897in}{1.357370in}}%
\pgfusepath{clip}%
\pgfsetbuttcap%
\pgfsetmiterjoin%
\definecolor{currentfill}{rgb}{0.713725,0.392157,0.239216}%
\pgfsetfillcolor{currentfill}%
\pgfsetlinewidth{0.000000pt}%
\definecolor{currentstroke}{rgb}{0.000000,0.000000,0.000000}%
\pgfsetstrokecolor{currentstroke}%
\pgfsetstrokeopacity{0.000000}%
\pgfsetdash{}{0pt}%
\pgfpathmoveto{\pgfqpoint{6.190320in}{0.480000in}}%
\pgfpathlineto{\pgfqpoint{7.025277in}{0.480000in}}%
\pgfpathlineto{\pgfqpoint{7.025277in}{1.562355in}}%
\pgfpathlineto{\pgfqpoint{6.190320in}{1.562355in}}%
\pgfpathlineto{\pgfqpoint{6.190320in}{0.480000in}}%
\pgfpathclose%
\pgfusepath{fill}%
\end{pgfscope}%
\begin{pgfscope}%
\pgfsetrectcap%
\pgfsetmiterjoin%
\pgfsetlinewidth{0.803000pt}%
\definecolor{currentstroke}{rgb}{0.713725,0.741176,0.768627}%
\pgfsetstrokecolor{currentstroke}%
\pgfsetdash{}{0pt}%
\pgfpathmoveto{\pgfqpoint{4.322103in}{0.480000in}}%
\pgfpathlineto{\pgfqpoint{7.154000in}{0.480000in}}%
\pgfusepath{stroke}%
\end{pgfscope}%
\begin{pgfscope}%
\definecolor{textcolor}{rgb}{0.000000,0.000000,0.000000}%
\pgfsetstrokecolor{textcolor}%
\pgfsetfillcolor{textcolor}%
\pgftext[x=4.868305in,y=0.606294in,,base]{\color{textcolor}{\fontfamily{\familydefault}\fontsize{9.500000}{11.400000}\selectfont\catcode`\^=\active\def^{\ifmmode\sp\else\^{}\fi}\catcode`\%=\active\def
\end{pgfscope}%
\begin{pgfscope}%
\definecolor{textcolor}{rgb}{0.000000,0.000000,0.000000}%
\pgfsetstrokecolor{textcolor}%
\pgfsetfillcolor{textcolor}%
\pgftext[x=6.607799in,y=1.585962in,,base]{\color{textcolor}{\fontfamily{\familydefault}\fontsize{9.500000}{11.400000}\selectfont\catcode`\^=\active\def^{\ifmmode\sp\else\^{}\fi}\catcode`\%=\active\def
\end{pgfscope}%
\begin{pgfscope}%
\definecolor{textcolor}{rgb}{0.000000,0.000000,0.000000}%
\pgfsetstrokecolor{textcolor}%
\pgfsetfillcolor{textcolor}%
\pgftext[x=4.322103in,y=2.087370in,left,base]{\color{textcolor}{\fontfamily{\familydefault}\fontsize{11.500000}{13.800000}\selectfont\catcode`\^=\active\def^{\ifmmode\sp\else\^{}\fi}\catcode`\%=\active\def
\end{pgfscope}%
\end{pgfpicture}%
\makeatother%
\endgroup%